\documentclass{article}

\PassOptionsToPackage{numbers}{natbib}
\usepackage[preprint]{neurips_2025}

\usepackage[utf8]{inputenc} 
\usepackage[T1]{fontenc}    
\usepackage{hyperref}       
\usepackage{url}            
\usepackage{booktabs}       
\usepackage{amsfonts}       
\usepackage{nicefrac}       
\usepackage{microtype}      
\usepackage{xcolor}         

\usepackage[pdftex]{graphicx}
\usepackage{placeins}
\usepackage{amsmath}
\usepackage{amssymb}

\usepackage{algorithm}
\usepackage{algorithmicx}
\usepackage{algpseudocode}
\usepackage{bm}

\newcommand{\suppressTOC}{%
  \let\realaddcontentsline\addcontentsline
  \renewcommand{\addcontentsline}[3]{}
}
\newcommand{\restoreTOC}{%
  \let\addcontentsline\realaddcontentsline
}

\suppressTOC

\title{PolyMicros: Bootstrapping a Foundation Model for Polycrystalline Material Structure}

\author{%
  Michael O.~Buzzy \\
  School of Computational Science and Engineering \\
  Georgia Institute of Technology \\
  Atlanta, GA 30332 \\
  \texttt{mbuzzy3@gatech.edu} \\
  \And
  Andreas E. ~Robertson \\
  Center for Integrated Nanotechnology \\
  Sandia National Laboratories \\
  Albuquerque, NM 87110 \\
  \texttt{aerober@sandia.gov} \\
  \And
  Peng ~Chen \\
  School of Computational Science and Engineering \\
  Georgia Institute of Technology \\
 Atlanta, GA 30332 \\
  \texttt{pchen402@gatech.edu} \\
  \And
  Surya R. ~Kalidindi \\
  School of Computational Science and Engineering \\
  George W. Woodruff School of Mechanical Engineering \\
  Georgia Institute of Technology \\
  Atlanta, GA 30332 \\
  \texttt{surya.kalidindi@me.gatech.edu} \\
}

\begin{document}

\maketitle

\begin{abstract}
  Recent advances in Foundation Models for Materials Science are poised to revolutionize the discovery, manufacture, and design of novel materials with tailored properties and responses. Although great strides have been made, successes have been restricted to materials classes where multi-million sample data repositories can be readily curated (e.g., atomistic structures). Unfortunately, for many structural and functional materials (e.g., mesoscale structured metal alloys), such datasets are too costly or prohibitive to construct; instead, datasets are limited to very few examples. To address this challenge, we introduce a novel machine learning approach for learning from hyper-sparse, complex spatial data in scientific domains. Our core contribution is a physics-driven data augmentation scheme that leverages an ensemble of local generative models, trained on as few as five experimental observations, and coordinates them through a novel diversity curation strategy to generate a large-scale, physically diverse dataset. We utilize this framework to construct PolyMicros, the first Foundation Model for polycrystalline materials (a structural material class important across a broad range of industrial and scientific applications). We demonstrate the utility of PolyMicros by zero-shot solving several long standing challenges related to accelerating 3D experimental microscopy. Finally, we make both our models and datasets openly available to the community.
  
\end{abstract}

\begin{figure}[!h]
\centering
\includegraphics[width=\linewidth]{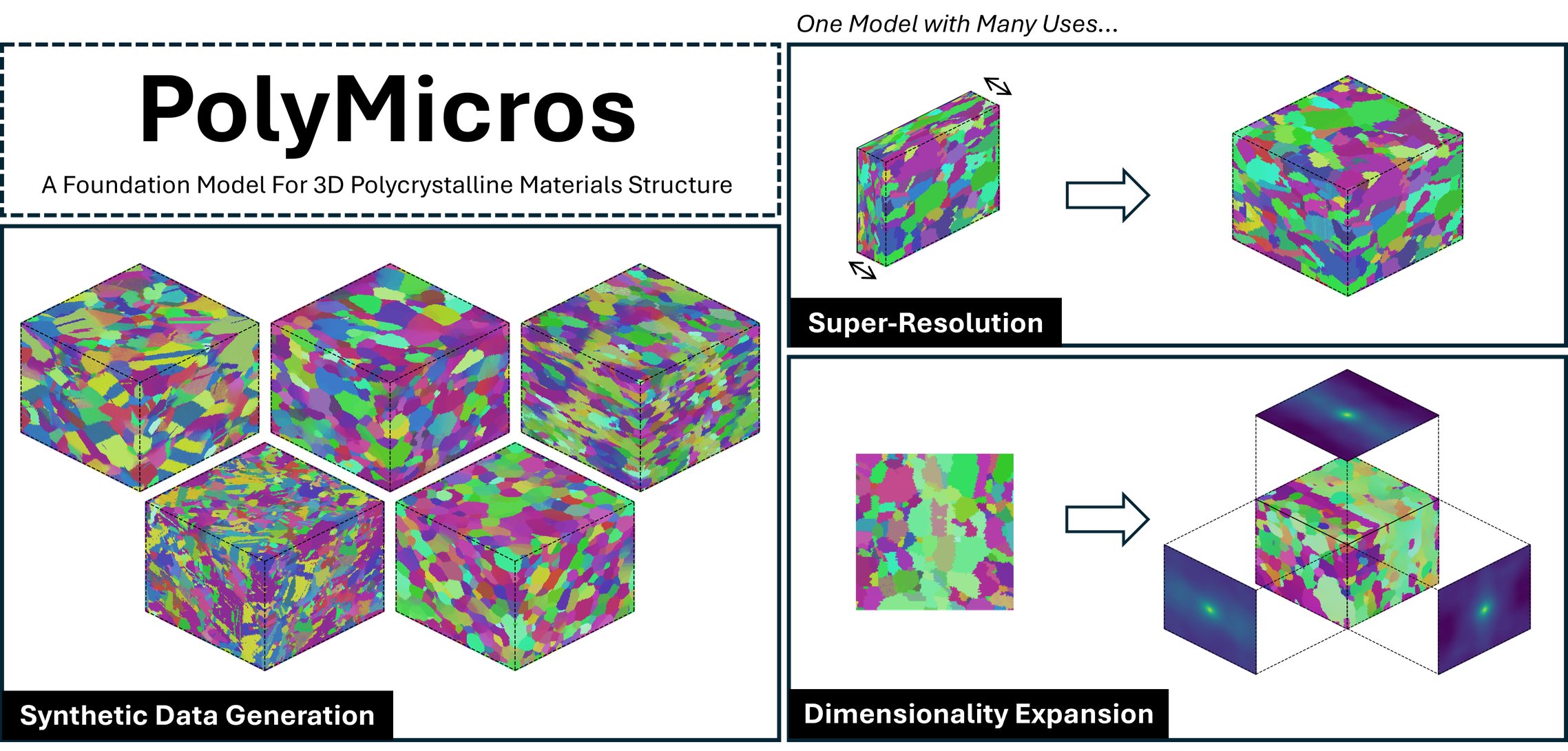}

\caption{ PolyMicros is a Foundation model for polycrystalline materials structure trained on a diverse synthetic dataset made by other smaller physics-driven generative models. By utilizing post-training conditioning, PolyMicros can be applied to address several problems in experimental microscopy without any specialized training. Shown above are two such applications.}
\label{fig:misc:headliner}
\end{figure}

\newpage

\section{Introduction}

Foundation Models are a class of generative machine learning models designed as generalist tools with the ability to solve diverse problems beyond their initial training scope while requiring no (zero-shot) or only minimal additional modification (fine-tuning) \cite{bommasani2022opportunitiesrisksfoundationmodels, jakubik2025terramind, brown2020language}. They are a sharp deviation from traditional solutions as they are no longer bespoke systems useful for a single task. Instead, they encode some level of fundamental knowledge about a domain area which can be readily exploited across many applications. In this way, foundation models have the potential to provide immense value. In scientific machine learning, available Foundation Models, such as AlphaFold \cite{jumper2021highly}, Aurora\cite{bodnar2024foundationmodelearth}, MatterGen \cite{zeni2025generative}, etc have already accelerated discovery and design efforts. The reusability of these generalist models significantly reduces the burden of fundamental tasks -- e.g., learning representations -- resulting in greater agility for the scientific ML community by allowing researchers to rapidly prototype new methods and application areas. 

Foundation Models achieve this generalist scope by training on tasks expected to extract generalized features (such as unconditional generation \cite{zeni2025generative, yenduri2023generativepretrainedtransformercomprehensive,devlin2019bertpretrainingdeepbidirectional} and multi-task learning \cite{bodnar2024foundationmodelearth, ye2025pdeformerfoundationmodelonedimensional, mizrahi20234m, jakubik2025terramind}), as well as training using vast and -- more importantly -- diverse corpuses of data. Unfortunately, this second requirement has practically limited their applicability to domains with access to large datasets. This requirement is incompatible with many scientific pursuits because they operate in spaces where information is limited \cite{dalla2024nucleotide}. The few exceptions (e.g., chemistry \cite{batatia2024foundationmodelatomisticmaterials, zeni2025generative}, drug design \cite{xia2024target-aware}, and astro-physics \cite{Leung_2023, parker2024astroclip}) have been achieved via intensive data collection campaigns of high quality data. For example, MatterGen -- a foundation model aimed at designing synthesizable inorganic compounds -- leverages a dataset of over a million carefully curated thermodynamically stable compounds \cite{zeni2025generative}. In many scientific domains, the necessary experimentation, analysis, and discarding of insufficient data demanded by such an effort remains unachievable.

For example, in mesoscale materials science -- a discipline that studies the important relationship between superstructures of atoms (called microstructures, e.g., Fig. \ref{fig:misc:headliner}) and engineering properties of interest, experimental observations may be in the double or even single digits -- far below the necessary quantity to train a generative model -- and cost thousands of dollars per sample \cite{brodniksurvey, neizgodasurvey, buzzy2025active, chapman2021afrl}. 
Within this domain, this is a significant roadblock, but achieving a Foundation Model for microstructures would be widely valuable because of the ubiquitous importance of materials across many industries and the central importance of the microstructure itself as the core quantity of interest in a variety of materials tasks (e.g., design \cite{torquato, kalidbook1, marshallmeanfield,  gao_stochasticdesign, titanium_generator, vlassis_diffusion_design, jong_design_latentspaces, adam_ISMD}, manufacturing \cite{generale2023a, sepidifferential, polycrysEvolution}, and monitoring \cite{rossin_nondestructive}). In this work, we challenge the perception that data-scare scientific pursuits and Foundation Models are incompatible by constructing a bootstrapping procedure for creating a Foundation Model for polycrystalline material microstructures -- a broadly important class of material systems \cite{bunge, mcdowellviscoplast, cpfem_raabe}. The term bootstrapping refers to a self-starting process by which a tool or system is iteratively improved through the use of itself. Specifically, we propose a physics directed augmentation scheme -- built on specialized generative models -- to extract large, suitable datasets from limited experimental field data. The bootstrapping procedure follows the recipe: train an ensemble of physics-based, statistics conditioned generative models on limited experimental data, coordinate these models using a suitable design of experiments scheme to carefully and systematically generate a new, more diverse physics-driven synthetic dataset, and finally train a foundation model on this synthetic dataset. Although this recipe appears simple, the critical challenge is designing and utilizing the smaller generative models to generate data which is both scientifically reasonable and \textit{quantitatively more diverse} than the limited initial dataset. 

\subsection{Contributions}

Our core motivation is the hypothesis that carefully designed synthetic datasets, while imperfect compared to high fidelity data sources, still support useful foundation models in data scarce scientific applications where preferable data sources are simply unachievable. We demonstrate this procedure by leveraging a small dataset of 5 experimentally gathered 3D polycrystalline material microstructures into a quantitatively diverse dataset and a practically useful Foundation Model. We provide analysis of the achieved expansion in the statistical diversity of the microstructure distribution. Furthermore, we showcase the zero-shot performance of the PolyMicros Foundation Model on two long standing challenges in experimental microscopy of 3D polycrystalline materials -- Microstructure Super-Resolution and Dimensionality Expansion (both shown in Fig. \ref{fig:misc:headliner}). We show in this work, that by acting in the role of a high-entropy prior, PolyMicros is able to be modularly conditioned to solve both of these tasks with no additional training. This indicates that, like other impactful scientific Foundation models, PolyMicros is capable of operating as a generalist tool, and has the potential to aid materials scientists across a wide variety of problems relating to polycrystalline material microstructure. Overall, we put forward several key contributions in this work:

\begin{enumerate}
    \item \textbf{Statistical Physics Directed Data Augmentation}: Propose a dataset augmentation scheme for stationary fields utilizing ensembles of statistics conditioned generative models to add physically meaningful diversity while retaining equal representation of the original data.
    
    \item \textbf{Versatile Foundation Model}: We curate the necessary data and train a foundation model -- specifically, a generative prior -- for polycrystalline microstructures.

    \item \textbf{Demonstrate Reuseability}: We demonstrate that modular conditioning techniques can be used to repeatedly reuse the PolyMicros Foundation Model as a generative prior to address several independent problems in polycrystalline microscopy without the need for retraining.
\end{enumerate}

\section{Background \& Related Work}

\subsection{Polycrystalline Materials}

Bulk engineering materials are composed of large structured arrays of periodically repeated base units -- often arrangements of atoms or molecules. As the unit is repeated, the system's entropy causes disruptions in the arrangement leading to a finite length to this regular ordering. We refer to each prestine region as a crystal; the entire system -- composed of a finite set of these regions -- as a polycrystalline material. The vast majority of important engineering materials -- from metal alloys in aerospace applications to photovoltaic thin films in solar cells -- are polycrystalline. Quantitatively, a polycrystalline material is defined as an orientation field. Here, we abstract the atoms or molecules that describes the material locally and instead describe the relative orientation of their crystal (a rotation in SO(3)). Polycrystalline materials are challenging to quantify due to their myriad important features spanning approximately three key length scales. 1) \textit{Pointwise}, polycrystalline materials display well understood symmetries due to the underlying symmetry in the base atomic lattice or molecule anisotropy \cite{bunge}. To account for these symmetries we utilize the Reduced Order Generalized Spherical Harmonics (ROGSH), a compact 3D vector description of a single orientation in SO(3) which adheres to specific material symmetries (i.e., cubic or hexagonal). Notably, learning on ROGSH fields outperforms alternatives \cite{paxti_polycrystal_diffusion, buzzy2024statistically}. 2) \textit{Locally}, individual crystals take on less deterministic characteristic forms -- i.e., boundary shapes, e.g., Fig. \ref{fig:misc:headliner}. 3) \textit{Globally}, ensembles of crystals arrange globally into specific statistical patterns. The co-dependent interplay between all three lengthscales dictates the polycrystal's engineering properties. All three must be carefully quantified and incorporated in the pursuit of a generalist dataset. Local and global descriptor choices are detailed below; see App. \ref{app:rogsh_appendix} for ROGSH specifics.

\subsubsection{Two-Point Spatial Correlations}
\label{sec:2ps}

To generate a diverse polycrystal dataset, we need a material representation where distance reflects functional differences (i.e., difference in material properties)—ensuring the generated diversity is meaningful for downstream materials science tasks. We use n-point spatial statistics, a common tool for describing local and global polycrystalline structure \cite{kalidbook1, polycrystal2ps}. These statistics address degeneracy in microstructure representation: visually and numerically distinct microstructures with similar spatial statistics exhibit similar properties and responses \cite{brownSCE}. The 1-point (e.g., the average) and the 2-point statistics (i.e., the auto- and cross-correlations of the spatial ROGSH fields, shown in Eq. \ref{eq:2ps}) are computationally manageable to utilize directly. For property prediction and simple process modeling, just the 1- and 2-point statistics are often sufficient to quantify the salient differences between polycrystalline microstructures and indicate differences in properties, and therefore will be our primary tool for microstructure quantification. For periodic spatial domains (i.e., representative volume elements (RVE)), the 2-point statistics can be computed efficiently using the Fast Fourier Transform \cite{polycrystal2ps} 

\begin{equation}
\label{eq:2ps}
    f_{r}^{\beta \gamma} = \frac{1}{S} \sum_{s=0}^{S-1} m_{s}^{\beta} m_{s+r}^{\gamma}.
\end{equation}

\noindent Here, $S$ is the number of voxels. $m_{s}^{\beta}$ is the $\beta$-component of the vector-valued microstructure descriptor in voxel $s$.

\subsection{Local-Global Decompositions Generative Framework}

This paper proposes to bridge the gap between limited experimental data and a large generalist dataset by using a design of experiments (DoX) scheme to target and populate the statistics space using an ensemble of specialized generative models. Local-Global Decomposition Generative Models (LGD) are a statistical physics-informed generative framework designed for precisely this task. LGD models generate microstructures conditioned on 1- and 2-point statistics as well as an abstraction of the higher order statistics called a `neighborhood distribution'. Critically, they are trained using a single example experimental microstructure -- defined by a single set of conditions -- and then can extrapolate with respect to the 1- and 2-point statistical conditioning \cite{andreas_LGD}. 

LGD models are a 2-stage model constructed by assuming a decomposition of the statistically conditioned microstructure generating process. Global, long range patterns are ascribed to 1- and 2-point statistics and the statistical process is approximated using a Multi-Output Gaussian Random Field model \cite{andreas_NGRF}. Higher order deviations from Gaussian structure -- e.g., deviations that describe complex local patterns such as grain boundaries -- are assumed to be localized. These deviations are approximated using a deep diffusion model. In practice, samples are drawn by first sampling the multi-output Gaussian Random Field and, subsequently, refining them using a post-training conditioned diffusion model. By compartmentalizing the impact of the 1- and 2-point statistics to an analytic model, the complete LGD model displays tremendous stability. We refer the interested reader to \cite{andreas_LGD, buzzy2024statistically} for more detail. We outline important equations and sampling schemes in App. \ref{app:lgd_appendix}.

\section{Synthetic Data Generation}
\label{sec:synthetic_data}

\begin{figure}[!h]
\centering
\includegraphics[width=\linewidth]{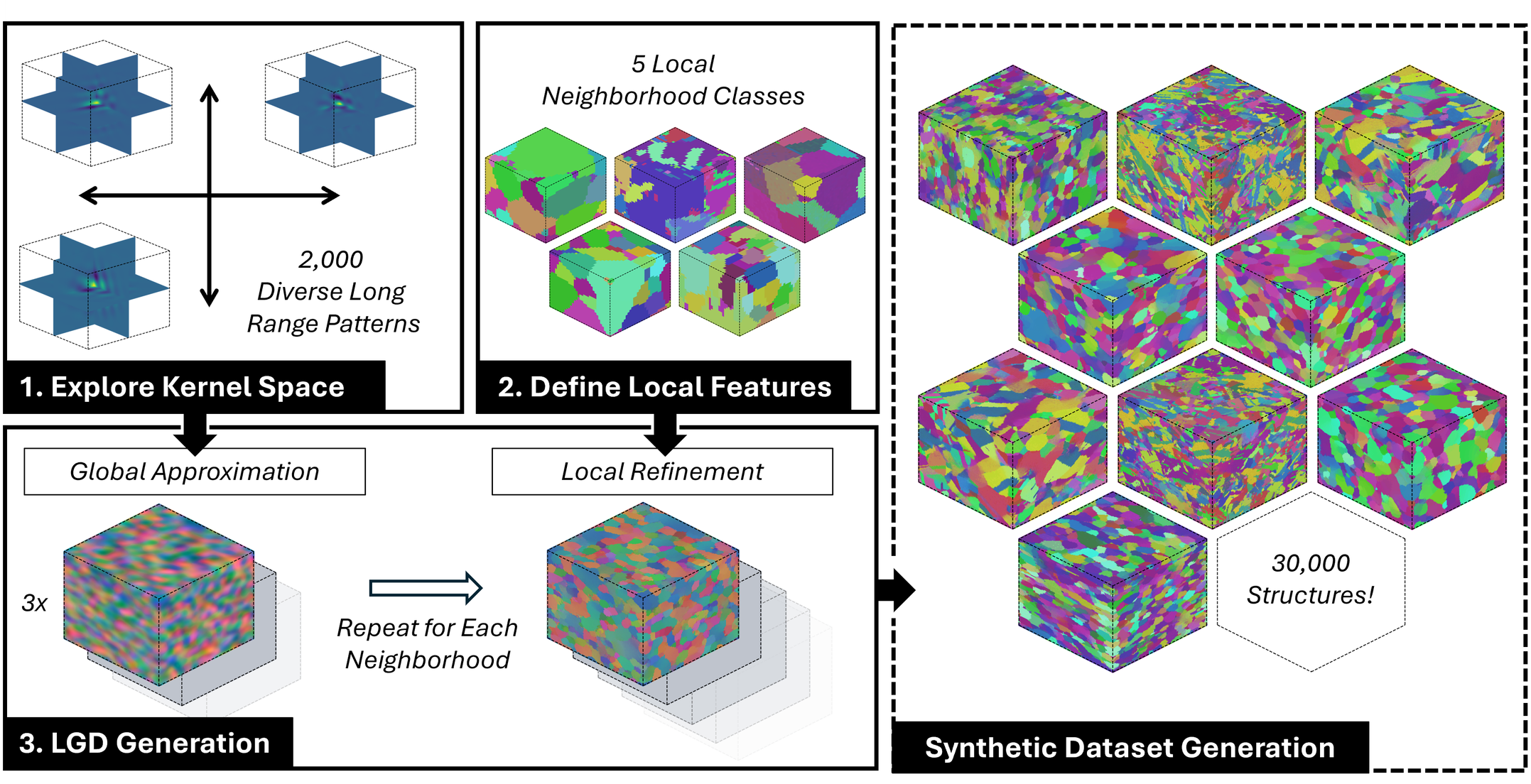}
\caption{Overview of the proposed Statistical Physics driven Data Augmentation algorithm.}
\label{fig:misc:framework}
\end{figure}

\FloatBarrier

In this section, we outline the proposed bootstrapping framework for constructing the dataset needed to train the PolyMicros Foundation Model (outlined in Fig. \ref{fig:misc:framework}). We begin with five data points -- experimentally collected 3D polycrystalline microstructures -- representing a wide range of material classes, manufacturing processes, and local microstructure morphology \cite{chapman2021afrl,rowenhorst2010three,vaughan2024mechanistic, stinville2022multi, jangid2023titanium, BELADI20131404}. Notably, these five data points constitute a majority of the publicly available 3D data, the remainder of which was held out for evaluating model performance during the case studies. Clearly, data augmentation is necessary. However, standard augmentation techniques (e.g., rotations, flips, color jittering, interpolation, etc.) are not applicable; they disrupt the physicality of the system (interpolation, scaling, jittering) or simply amount to changes in reference frame (rotation, flips). In addition, such augmentation schemes do not induce physically meaningful diversity. Instead, we propose an augmentation scheme guided by statistical physics which aims to induce the meaningful diversity which will make PolyMicros robust in downstream tasks. 

We utilize the 1- and 2-point spatial statistics introduced in Sec. \ref{sec:2ps} to assign a definition to the phrase `meaningful diversity'. It is well established theoretically \cite{brownSCE, torquatov1SCE, garmestaniSCE, fullwoodpolycrystalSCE} and empirically \cite{adam, polycrysEvolution, polycrystal2ps} that these statistical microstructure representations strongly correlate with many emergent properties of interest to materials scientists. As a result, wide coverage in the space of 2-point statistics as a definition for dataset diversity has the major benefit that it is a goal agnostic measure. Their strong correlation with arbitrary properties indicates that the dataset will include diversity with respect to a general class of materials problems without needing to specifically consider each problem individually \cite{robertson2024micro2d}. 

Our proposed augmentation framework combines the local neighborhood information present in the experimental data -- proxy definitions of the minimum characteristics necessary to be a realistic microstructure -- with a greatly expanded set of 2-point statistics -- i.e., long range spatial patterns -- in order to create a realistic, but diverse large microstructure dataset. To this end, we use the recently developed Local-Global Decomposition (LGD) framework \cite{buzzy2024statistically, andreas_LGD} for building data efficient deep generative microstructure models to synthetically generate this dataset. Like prior works utilizing the LGD generation framework \cite{buzzy2024statistically, andreas_LGD}, we choose to utilize Gaussian Random Fields for our global approximation. For polycrystals, we will need to utilize a Multi-Output Gaussian Random Field (MOGRF) \cite{andreas_NGRF} since the polycrystalline local state is an ROGSH vector rather than a scaler. The augmentation framework, outlined in Fig. \ref{fig:misc:framework}, involves two preparation steps corresponding to the two conditional inputs to an LGD model. Step 1 uses a Design of Experiments scheme to construct a large dataset of diverse (i.e., numerically dissimilar) 2-point statistics ($2000$ in total). Step 2 trains an ensemble of diffusion models; one model to represent each experimental microstructure's local neighborhood distribution ($5$ in total). Finally, these two inputs are combinatorially combined using a proposed adapted LGD sampling scheme to generate the augmented dataset, see App. \ref{app:lgd_sampling_adjustments}. Each 2-point statistic is repeated $3$ times leading to $30,000$ total microstructures sampled. 

\subsection{Global Statistics}
\label{sec:globalstatistics}

Prior work \cite{buzzy2024statistically} has sourced target 2-point statistics (i.e., auto-correlations and cross-correlations) for generating polycrystalline microstructures directly from experimental observations. For our purposes, this strategy -- or even an extension interpolating between our experimental statistics -- results in insufficient diversity. Instead, we propose to sample a generalized covariance kernel parameterization \cite{samo2015generalizedspectralkernels} to produce plausible, diverse 2-point statistics, extending initial ideas from \cite{robertson2024micro2d}. Specifically, we propose to use the Multi-Output Spectral Mixture Kernel (MOSM) parameterization \cite{parra2017spectralmixturekernelsmultioutput}
\begin{equation}
    k_{\beta \gamma}(r) = \sum_{q=1}^Q \alpha_{\beta \gamma}^{(q)} 
    \exp{\left( -\frac{1}{2} \left( r + \theta_{\beta \gamma}^{(q)} \right) ^{\intercal} { {A}_{\beta \gamma}^{(q)} } \left( r + \theta_{\beta \gamma}^{(q)} \right) \right)} \cos{\left( \left( r + \theta_{\beta \gamma}^{(q)} \right)^{\intercal} \varsigma_{\beta \gamma}^{(q)} + \phi_{\beta \gamma}^{(q)} \right)},
    \label{eq:MOSM}
\end{equation}
where the super index $(\cdot)^{(q)}$ denotes the $q^{th}$ component for the spectral mixture. A list of kernel parameters and their impacts on the covariance structure can be found in App. \ref{app:MOSM}. Unlike standard multi-output kernel methods, such as the Linear Method of Co-regionalization, MOSM is able to generate cross correlations which are "out-of-phase" of the autocorrelation (i.e., the ROGSH terms correlate with one another with respect to some spatial offset). Such structure is extremely common in real polycrystalline systems, see App. \ref{app:MOSM}.
 
With the parametric kernel form selected, we must finally define permissible ranges over each parameter as well as select the number of terms, $Q$, before we can use the kernel to generate a large dataset of diverse multi-output covariances. We aim to make such selections such that the $\mathrm{MOSM}+\mathrm{MOGRF}$ samples are good approximatations of physically feasible polycrystalline microstructures. We use several heuristics because defining a feasible polycrystal is not a trivial task. First, because ROGSH values must exist between \cite{bunge} $[-1, 1]$, we can eliminate any parameter value which leads to generated samples from the MOGRF outside of this range. Second, we restrict the parameters such that the generated covariances are periodic in order to be compatible with the MOGRF sampling algorithm \cite{andreas_NGRF}. Using this algorithm is necessary for computational feasability due to the high dimensionality of the microstructures. We enforce periodicity by imposing that the covariances gradually decay before the end of the domain, i.e., in the language of materials science: restricting the coherence length, the longest distance two local materials states influence each-other. The coherence length is controlled by the MOSM Kernel's ${A}_{\beta \gamma}^{(q)}$ parameter. Finally, we select the number of mixtures empirically using an approximation test. We select the minimum $Q$ such that we can optimize a sufficient approximation of an experimentally collected polycrystal's multi-output covariance map, App. \ref{app:num_mixture_elements}. With these heuristics, we empirically derive parameter bounds through experimentation. A summary of these experiments and the selected bounds can be found in App. \ref{app:MOSM}. With the bounds specified, 2000 sets of kernel parameters (for structures sized 128 voxels cubed) were sampled using Latin Hyper Cube Sampling to construct a diverse covariance dataset. 

\subsection{Local Neighborhoods}
\label{sec:localrefinement}

We train five diffusion models to approximate the local neighborhoods from the five experimentally obtained data points. An exemplar neighborhood is shown for each of the five neighborhood classes in Fig.  \ref{fig:misc:framework}, and additional examples for each class can be seen in App. \ref{app:hoods}. We chose to utilize an independent diffusion model for each neighborhood, rather than a singular diffusion model encompassing them all, to fight mode collapse caused by potential imbalances in the difficulty of learning each neighborhood \cite{andreas_LGD}. This strategy ensures that the generated dataset represents each local morphology equally. We use a similar training strategy to that described by Buzzy \textit{et al} \cite{buzzy2024statistically}; each experimental microstructure was converted into a dataset of local neighborhoods by taking patches of $32$ voxels cubed. The patches were taken with overlap (stride of $8$), resulting in approximately $10,000$ patches per structure. Combining the neighborhoods with the generated covariances, we sampled a dataset of $30,000$ synthetic polycrystalline microstructures each 128 voxels cubed, see Fig. \ref{fig:misc:framework} and App. \ref{app:dataset}.

\section{PolyMicros}
\label{sec:modular_conditioning}

We utilize the constructed dataset to train PolyMicros, a foundation model approximating the unconditional distribution of polycrystalline microstructures. In the language of Bayesian statistics, PolyMicros is a generalized prior detailing the characteristics that define a generalized polycrystalline microstructure. PolyMicros is a UNet-based unconditional diffusion model implemented and trained using the EDM infrastructure \cite{karras2022elucidating}. It operates directly on the ambient representation of polycrystals without any dimensionality reduction. In order to simplify the memory and computational cost associated with training, PolyMicros is pre-trained on the union of the collected $32^3$ neighborhood datasets from \autoref{sec:localrefinement}. Finally, it is trained on the full curated $128^3$ dataset. Training details can be found in App. \ref{app:training_neighborhood_models}. We chose the diffusion-model infrastructure because many techniques have been developed to perform modular conditioning post-training \cite{bansal2023universal, sun2024provable, learnedconditionedDDPM, zeni2025generative}. As a result, PolyMicros' generalized knowledge -- the characteristics of a polycrystal extracted from multiple polycrystal classes and numerous spatial arrangements -- can be accessed and reused in many applications using Bayes' rule to adapt the diffusion process 
\begin{equation}
    \nabla_m \log P(m|\phi) = \nabla_m \log P(\phi | m) + \nabla_m \log P(m).
\end{equation}

Here, $\phi$ is an arbitrary task (the next section will provide examples). $m$ is a polycrystalline microstructure. Generally, we found that the specific optimal post-training conditioning strategy depends on the application. For statistics based conditioning, we propose and utilize an optimization based strategy closely related to conditioning using proximal methods \cite{sun2024provable}. Here, we alternate between directly numerically optimizing the field to enforce target statistics and diffusion sampling steps to enforce polycrystalline characteristics. Precise details can be found in App. \ref{app:additional_experimental_results}.

\section{Experiments}
\label{sec:exp}
\subsection{Dataset Diversity}

\FloatBarrier

\begin{figure}[!h]
\centering
\includegraphics[width=\linewidth]{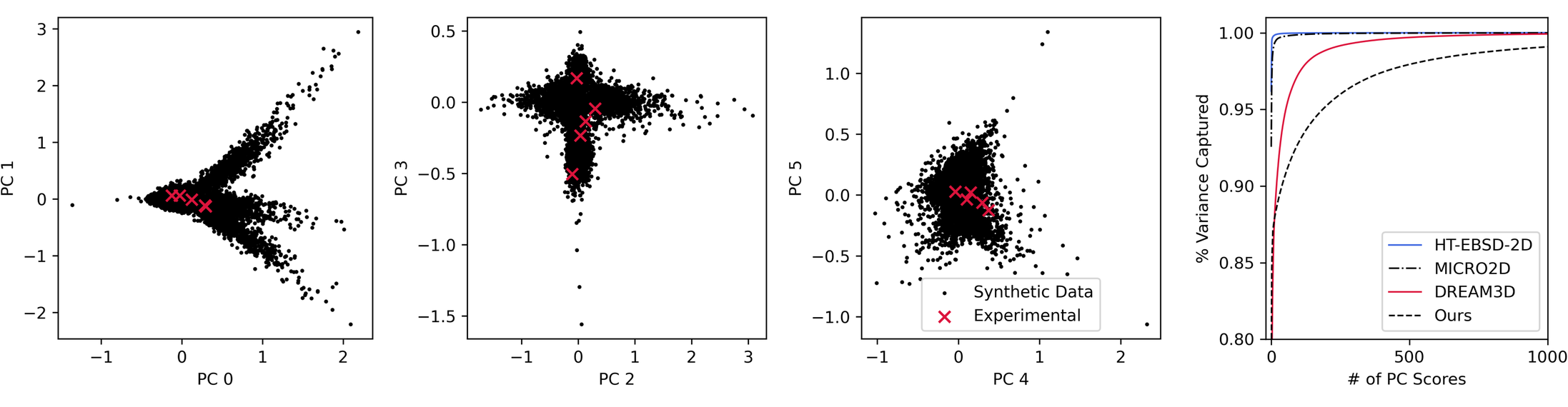}
\caption{Principal Component projections (i.e., PC basis coefficients computed from the 2-point statistics) comparing the statistical diversity of the original 5 experimental images against the augmented dataset. Additionally, comparison of explained variance saturation across various microstructure datasets.}
\label{fig:DGEN:PCA}
\end{figure}

We begin inspecting the first key claim of this work: that the data bootstrapping procedure produces a dataset of \textit{substantially greater} second order statistical diversity than the original five data points. We begin qualitatively using a well-established feature extraction procedure  \cite{polycrystal2ps}: principal component analysis (PCA) of the computed covariances of the experimental and synthetic structures. Fig. \ref{fig:DGEN:PCA} illustrates the synthetic dataset's shape and its relative diversity with respect to the original data points via a series of two-dimensional projections of the first five PC scores. First, the synthetic data successfully performs simple augmentations -- filling out the space between experimental samples. More importantly, it expands \textit{outside} of the original space, e.g., the three large tails in (a) and the horizontal band in (b). This is a direct consequence of the broad statistical exploration targeted by the global statistics DoX effort, \autoref{sec:globalstatistics}. We note that a limitation of PC analysis is that it is challenging to assign physical interpretation to absolute distances between points. Therefore, the importance of the synthetic dataset's spread is best defined relative to the original experimental data. To aid here, we emphasize that the original experimental points were themselves derived from diverse material systems (i.e., with widely different manufacturing origins and morphologies). While seemingly near each other in PC space, in a practical sense the experimental points themselves span a considerable space.

Next, we quantitatively inspect the diversity of the synthetic dataset. Since diversity means very little without reference, we compare our dataset to three other microstructural datasets \cite{fowler2024high, robertson2024micro2d, dream3d, dream3d_generation}. The first of these datasets (HT-EBSD-2D) is a 2D high-throughput experimental dataset produced by autonomous experimentation, and represents the most thorough to date experimental effort for creating large polycrystalline microstructure datasets. MICRO2D is a 2D synthetically generated dataset of binary microstructures designed with the intent of second order statistical diversity. Finally, DREAM3D is a dataset generated using the popular DREAM3D software package for creating synthetic 3D polycrystalline volumes using classical ellipsoid packing approaches. Using this software we sampled $5,000$ volumes\footnote{These were excluded from the synthetic dataset due to their lack of realism. See \cite{buzzy2024statistically}}. Using the same PCA + covariance method, we compare our dataset against these three by inspecting the convergence of the percent variance captured with increasing PC scores. Here, we use the rate of convergence as a measure of the complexity of the dataset, with slower convergence indicating greater complexity and thus requiring more PC basis vectors to span the dataset. The PolyMicros dataset converges much more slowly when compared to the reference datasets, Fig. \ref{fig:DGEN:PCA}, indicating greater complexity. We believe this is compelling evidence of the effectiveness of our bootstrapping approach, and indicates that the generated dataset is indeed of meaningful diversity.

\subsection{Reconstruction from Partial Observations}
\label{sec:cs1}

\begin{figure}
\centering
\includegraphics[width=\linewidth]{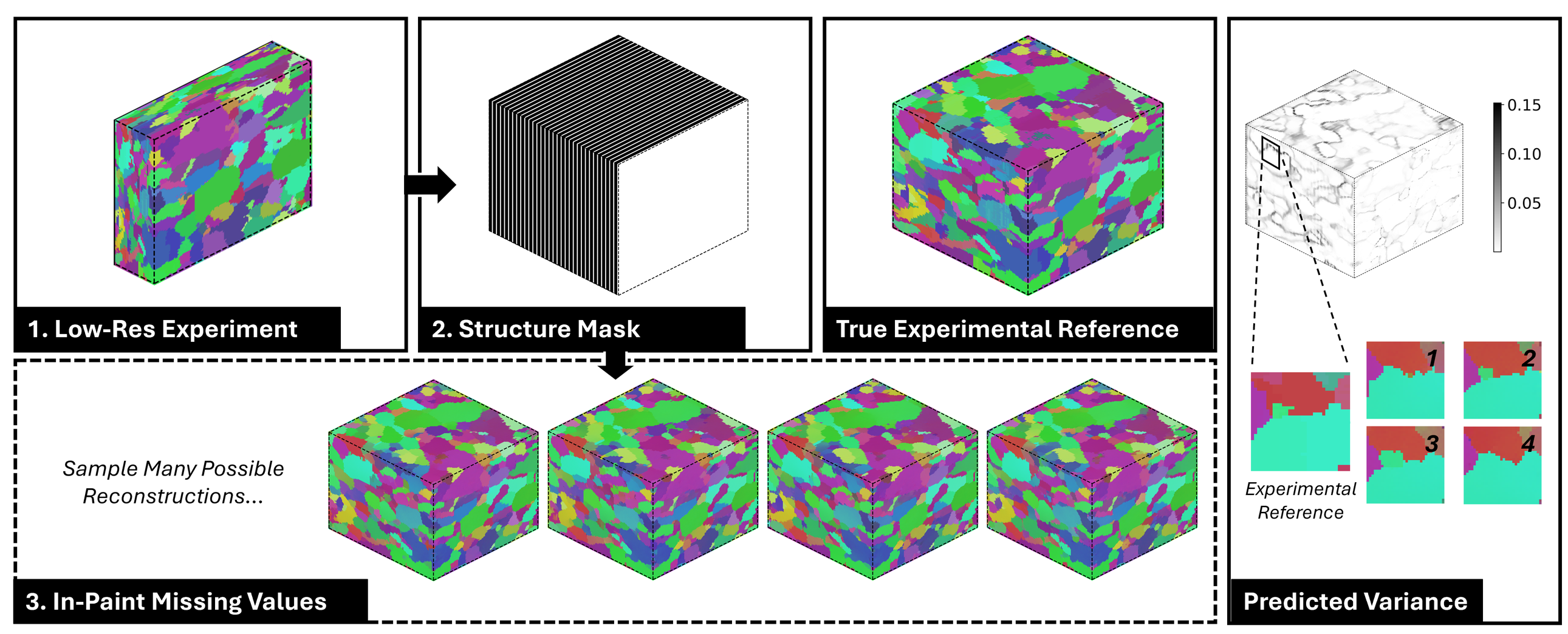}
\caption{Overview of modular post-training conditioning of the PolyMicros Foundation Model to accelerate serial-sectioning microscopy.}
\label{fig:CS1:CS1Over}
\end{figure}

We now inspect the second key claim of this work: that the PolyMicros foundation model, trained on carefully curated synthetic data, can serve as a generalist tool aiding materials scientists to solve real problems outside of its original training scope. We study two case studies beginning with microstructure super-resolution. Microstructure super-resolution aims to alleviate the cost and time burdens of serial-sectioning experiments -- where a 3D observation of a microstructure is formed through alternating between imaging of the 2D surface and erosion of the surface material. This process is extremely costly -- single samples can take up to a month \cite{chapman2021afrl,rowenhorst2010three,vaughan2024mechanistic, stinville2022multi, jangid2023titanium, BELADI20131404} -- because it requires a large number of removal steps to achieve high resolution along the erosion direction. Recent work has accelerated the process by imaging only every fourth slice and using specialized super-resolution transformers to predict the missing intermediate slices \cite{jangid2024q}. Here, we leverage PolyMicros to zero-shot the same problem. Posing microstructure super-resolution as an in-painting problem, we condition the PolyMicros model on the known observations from low-resolution experimental data, and utilize our model to impute the missing slices.

Fig. \ref{fig:CS1:CS1Over} outlines the super-resolution setup. To validate against a known 3D volume, we apply it to a held-out high resolution experimental reference \cite{BELADI20131404} which we down-sample to simulate a low-resolution experiment. Similar to \cite{jangid2024q}, we use a 4X down-sample along one axis to approximate the serial-sectioning scenario. The low resolution microstructure is shown in Fig. \ref{fig:CS1:CS1Over}. Because the absolute position (with respect to the machine's reference frame) of each voxel is known in serial sectioning experiments, we can prescribe the known low-resolution values at their appropriate positions in a 3D volume congruent with the full-resolution of the PolyMicros model. This forms a voxel structure mask of known values (Fig. \ref{fig:CS1:CS1Over}). To condition during the diffusion process, we simply set these positions to their known values after each diffusion step. This reprojection guides the diffusion model to produce samples conditioned on the known values from the low-resolution experimental observation. We found that if we perform this reprojection after every diffusion step the generated images possessed vertical artifacts where the reprojected slices contrasted with the imputed ones due to slight errors. To avoid this, we only perform the reprojection for the first $75\%$ of the diffusion process, allowing the remainder to progress normally. See App. \ref{app:superres} more details and ablation.

Inspecting Fig. \ref{fig:CS1:CS1Over}, the generated structures are nearly visually identical to the experimental reference. We compute a point-wise MAPE between the generated and reference microstructure of $4.05\%$ indicating good agreement. In addition, we can inspect the uncertainty associated with the prediction by producing many samples from the conditioned generative process. The uncertainty is primarily concentrated around the grain boundaries with a width of 1-5 voxels. The model is very confident in its prediction of the constant values within the grain, but indicates uncertainty in exact placement of the grain boundary within a narrow window. Furthermore, we can observe that along the vertical slices where known observations were enforced the variance is much lower. 

\subsection{Reconstruction from Partial Statistics}

\begin{figure}
\centering
\includegraphics[width=\linewidth]{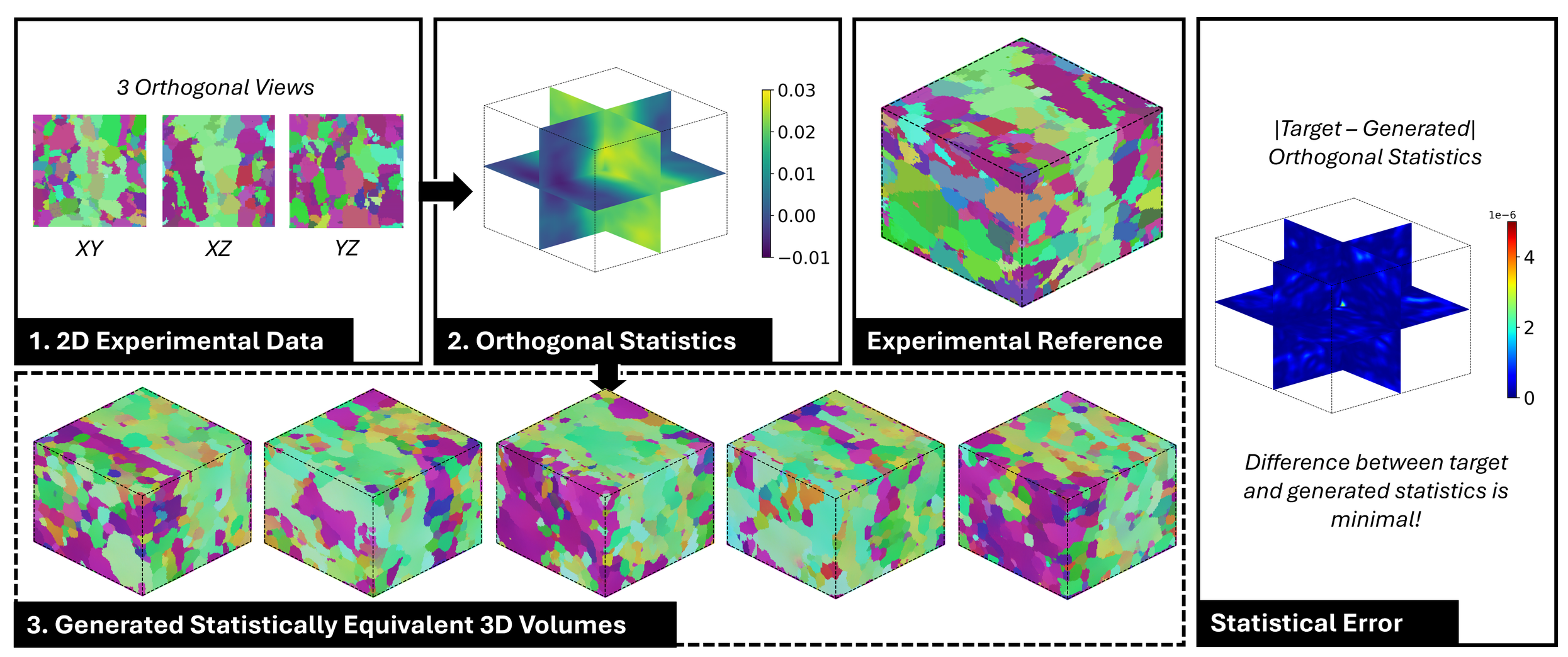}
\caption{Overview of modular post-training conditioning of the PolyMicros Foundation Model to propose potential 3D microstructures from 2D microscopy.}
\label{fig:CS2:CS2Over}
\end{figure}

The second application is more challenging: microstructure dimensionality expansion. Here, the goal is to avoid 3D characterization entirely and instead generate plausible 3D volumes based on few orthogonal 2D experimental observations, Fig. \ref{fig:CS2:CS2Over}(1). While few algorithms exist for other material types, these solutions have proven insufficient for polycrystals \cite{2Dto3DGAN}. In addition to limited conditioning information (just three slices, far less than what was available in the super-resolution task), this problem is further complicated because 2D experimentation strategies are unable to capture the relative 3D position information of voxel \textit{between} slices. Instead, we are limited to relative voxel location within a single image as well as knowing the relative orientation of each slice. For this reason, dimensionality expansion strategies aim to generate statistically equivalent 3D volumes instead of pursuing the previous in-painting approach \cite{2Dto3DGAN, seibert2023two, dream3d_generation, voronoi_laguerre, markovturner, lee2024multi, phan2024generating}. We refer the reader to App. \ref{app:RPS} for more background and motivation. Fig. \ref{fig:CS2:CS2Over} summarizes the dimensionality expansion process using PolyMicros. We target generating microstructures characterized by the same 2-point statistics along the observed orthogonal directions as the 2D experiments. We target the 2-point statistics because of their previously discussed physical importance, however the proposed procedure can be used with any differentiable statistic that can be computed without relative slice positions (e.g., \cite{threepoint_bayesiangen}). Again, we simulate performing the 2D experiments by extracting them from a 3D reference in order to have a ground truth to compare against. We compute target 2-point statistics from each of the three orthogonal slices to be used to guide the PolyMicro's diffusion process. Next, we sample PolyMicros using the conditional sampling process described in App. \ref{app:RPS}. Briefly, sampling is performed by alternating between optimizing the current noised microstructure to match the target statistics and a standard unconditional PolyMicros diffusion step.

Conditionally drawn samples display good qualitative agreement with the experimental reference, Fig. \ref{fig:CS2:CS2Over}. Both share prominent morphological features such as vertical green-blue banding spanning the full vertical domain broken up by randomly oriented regions. Notably, these patterns' exact locations vary because the conditioned statistics only enforce relative structure. The generated microstructures also contain realistic grain shapes similar to the reference. Furthermore, the variability in the sampling process displays expected trends: the conditional sampling process produces greater variety than in the previous case study. This increase in uncertainty successfully reflects the significant decrease in conditioning information provided. Comparing the 2-point statistics quantitatively confirms these initial observations. We achieve a maximum error below $1e^{-5}$ on the target statistics across the generated samples. App. \ref{app:RPS} presents comparisons of the full 2-point statistics errors. Overall, this result is incredibly precise, and is the first successful attempt at performing the microstructure dimensionality task for polycrystalline materials beyond first-order accuracy.

\section{Conclusions}

In this work we introduce PolyMicros, a Foundation Model for polycrystalline microstructure. To achieve this in materials science's ultra-low data environment, we propose a physics-driven data augmentation scheme which expands a small dataset (5 data points) into a diverse dataset of 30,000 unique materials. We show that PolyMicros -- trained to unconditionally generate this carefully curated synthetic dataset -- is capable of zero-shot solving two longstanding challenges in 3D experimental microscopy: super-resolution and dimensionality expansion. Looking forward within the materials domain, the techniques exemplified here can be readily extended to solve a wider set of structure-property inverse problems \cite{buzzy2025active}. Additionally, they can be used to increase the robustness of high-throughput microscopy techniques such as \cite{fowler2024high} and enable Agentic AI systems for accelerating materials research. Potential future improvements to unlock this are suggested in App. \ref{app:limitations}. Finally, although exemplified on microstructures, the techniques developed within can be extended to other data-sparse scientific domains provided that any spatial variables are also stationary \cite{qiu2024derivative, lim2023score, kovachki2023neural, oommen2024integrating}.

\begin{ack}

We acknowledge the generous support from Microsoft made available for this research via Georgia Tech Cloud Hub. 

Surya Kalidindi was supported by the Army Research Laboratory under Cooperative Agreement Number W911NF-22-2-0106.

A.E. Robertson would like to acknowledge the Jack Kent Cooke Foundation for their extended support. 

This article has been authored by an employee of National Technology \& Engineering Solutions of Sandia, LLC under Contract No. DE-NA0003525 with the U.S. Department of Energy (DOE). The employee owns all right, title, and interest in and to the article and is solely responsible for its contents. The United States Government retains and the publisher, by accepting the article for publication, acknowledges that the United States Government retains a non-exclusive, paid-up, irrevocable, world-wide license to publish or reproduce the published form of this article or allow others to do so, for United States Government purposes. The DOE will provide public access to these results of federally sponsored research in accordance with the DOE Public Access Plan www.energy. gov/downloads/doe-public-access-plan.

\end{ack}

\newpage

\bibliographystyle{plainnat}
\bibliography{references}

\begin{thebibliography}{97}
\providecommand{\natexlab}[1]{#1}
\providecommand{\url}[1]{\texttt{#1}}
\expandafter\ifx\csname urlstyle\endcsname\relax
  \providecommand{\doi}[1]{doi: #1}\else
  \providecommand{\doi}{doi: \begingroup \urlstyle{rm}\Url}\fi

\bibitem[Abdin et~al.(2024{\natexlab{a}})Abdin, Aneja, Awadalla, Awadallah, Awan, Bach, Bahree, Bakhtiari, Bao, Behl, et~al.]{abdin2024phi3}
Marah Abdin, Jyoti Aneja, Hany Awadalla, Ahmed Awadallah, Ammar~Ahmad Awan, Nguyen Bach, Amit Bahree, Arash Bakhtiari, Jianmin Bao, Harkirat Behl, et~al.
\newblock Phi-3 technical report: A highly capable language model locally on your phone.
\newblock \emph{arXiv preprint arXiv:2404.14219}, 2024{\natexlab{a}}.

\bibitem[Abdin et~al.(2024{\natexlab{b}})Abdin, Aneja, Behl, Bubeck, Eldan, Gunasekar, Harrison, Hewett, Javaheripi, Kauffmann, et~al.]{abdin2024phi4}
Marah Abdin, Jyoti Aneja, Harkirat Behl, S{\'e}bastien Bubeck, Ronen Eldan, Suriya Gunasekar, Michael Harrison, Russell~J Hewett, Mojan Javaheripi, Piero Kauffmann, et~al.
\newblock Phi-4 technical report.
\newblock \emph{arXiv preprint arXiv:2412.08905}, 2024{\natexlab{b}}.

\bibitem[Adams et~al.(2013)Adams, Kalidindi, and Fullwood]{kalidbook1}
B.L. Adams, S.R. Kalidindi, and D.T. Fullwood.
\newblock \emph{Microstructure Sensitive Design for Performance Optimization}.
\newblock Butterworth-Heinemann, Waltham, MA, 2013.

\bibitem[Attari et~al.(2020)Attari, Honarmandi, Duong, Sauceda, Allaire, and Arroyave]{attari2020uncertainty}
Vahid Attari, Pejman Honarmandi, Thien Duong, Daniel~J Sauceda, Douglas Allaire, and Raymundo Arroyave.
\newblock Uncertainty propagation in a multiscale calphad-reinforced elastochemical phase-field model.
\newblock \emph{Acta Materialia}, 183:\penalty0 452--470, 2020.

\bibitem[Bansal et~al.(2023)Bansal, Chu, Schwarzschild, Sengupta, Goldblum, Geiping, and Goldstein]{bansal2023universal}
Arpit Bansal, Hong-Min Chu, Avi Schwarzschild, Soumyadip Sengupta, Micah Goldblum, Jonas Geiping, and Tom Goldstein.
\newblock Universal guidance for diffusion models.
\newblock In \emph{Proceedings of the IEEE/CVF Conference on Computer Vision and Pattern Recognition}, pages 843--852, 2023.

\bibitem[Barber et~al.()Barber, Leake, and Clyne]{doitpoms_micrograph_database}
Z.H. Barber, J.A. Leake, and T.W. Clyne.
\newblock The doitpoms project: Micrograph library.
\newblock URL \url{https://www.doitpoms.ac.uk/miclib/index.php}.

\bibitem[Batatia et~al.(2024)Batatia, Benner, Chiang, Elena, Kovács, Riebesell, Advincula, Asta, Avaylon, Baldwin, Berger, Bernstein, Bhowmik, Blau, Cărare, Darby, De, Pia, Deringer, Elijošius, El-Machachi, Falcioni, Fako, Ferrari, Genreith-Schriever, George, Goodall, Grey, Grigorev, Han, Handley, Heenen, Hermansson, Holm, Jaafar, Hofmann, Jakob, Jung, Kapil, Kaplan, Karimitari, Kermode, Kroupa, Kullgren, Kuner, Kuryla, Liepuoniute, Margraf, Magdău, Michaelides, Moore, Naik, Niblett, Norwood, O'Neill, Ortner, Persson, Reuter, Rosen, Schaaf, Schran, Shi, Sivonxay, Stenczel, Svahn, Sutton, Swinburne, Tilly, van~der Oord, Varga-Umbrich, Vegge, Vondrák, Wang, Witt, Zills, and Csányi]{batatia2024foundationmodelatomisticmaterials}
Ilyes Batatia, Philipp Benner, Yuan Chiang, Alin~M. Elena, Dávid~P. Kovács, Janosh Riebesell, Xavier~R. Advincula, Mark Asta, Matthew Avaylon, William~J. Baldwin, Fabian Berger, Noam Bernstein, Arghya Bhowmik, Samuel~M. Blau, Vlad Cărare, James~P. Darby, Sandip De, Flaviano~Della Pia, Volker~L. Deringer, Rokas Elijošius, Zakariya El-Machachi, Fabio Falcioni, Edvin Fako, Andrea~C. Ferrari, Annalena Genreith-Schriever, Janine George, Rhys E.~A. Goodall, Clare~P. Grey, Petr Grigorev, Shuang Han, Will Handley, Hendrik~H. Heenen, Kersti Hermansson, Christian Holm, Jad Jaafar, Stephan Hofmann, Konstantin~S. Jakob, Hyunwook Jung, Venkat Kapil, Aaron~D. Kaplan, Nima Karimitari, James~R. Kermode, Namu Kroupa, Jolla Kullgren, Matthew~C. Kuner, Domantas Kuryla, Guoda Liepuoniute, Johannes~T. Margraf, Ioan-Bogdan Magdău, Angelos Michaelides, J.~Harry Moore, Aakash~A. Naik, Samuel~P. Niblett, Sam~Walton Norwood, Niamh O'Neill, Christoph Ortner, Kristin~A. Persson, Karsten Reuter, Andrew~S. Rosen, Lars~L. Schaaf,
  Christoph Schran, Benjamin~X. Shi, Eric Sivonxay, Tamás~K. Stenczel, Viktor Svahn, Christopher Sutton, Thomas~D. Swinburne, Jules Tilly, Cas van~der Oord, Eszter Varga-Umbrich, Tejs Vegge, Martin Vondrák, Yangshuai Wang, William~C. Witt, Fabian Zills, and Gábor Csányi.
\newblock A foundation model for atomistic materials chemistry, 2024.
\newblock URL \url{https://arxiv.org/abs/2401.00096}.

\bibitem[Beladi and Rohrer(2013)]{BELADI20131404}
Hossein Beladi and Gregory~S. Rohrer.
\newblock The relative grain boundary area and energy distributions in a ferritic steel determined from three-dimensional electron backscatter diffraction maps.
\newblock \emph{Acta Materialia}, 61\penalty0 (4):\penalty0 1404--1412, 2013.
\newblock ISSN 1359-6454.
\newblock \doi{https://doi.org/10.1016/j.actamat.2012.11.017}.
\newblock URL \url{https://www.sciencedirect.com/science/article/pii/S135964541200821X}.

\bibitem[Bodnar et~al.(2024)Bodnar, Bruinsma, Lucic, Stanley, Vaughan, Brandstetter, Garvan, Riechert, Weyn, Dong, Gupta, Thambiratnam, Archibald, Wu, Heider, Welling, Turner, and Perdikaris]{bodnar2024foundationmodelearth}
Cristian Bodnar, Wessel~P. Bruinsma, Ana Lucic, Megan Stanley, Anna Vaughan, Johannes Brandstetter, Patrick Garvan, Maik Riechert, Jonathan~A. Weyn, Haiyu Dong, Jayesh~K. Gupta, Kit Thambiratnam, Alexander~T. Archibald, Chun-Chieh Wu, Elizabeth Heider, Max Welling, Richard~E. Turner, and Paris Perdikaris.
\newblock A foundation model for the earth system, 2024.
\newblock URL \url{https://arxiv.org/abs/2405.13063}.

\bibitem[Bommasani et~al.(2022)Bommasani, Hudson, Adeli, Altman, Arora, von Arx, Bernstein, Bohg, Bosselut, Brunskill, Brynjolfsson, Buch, Card, Castellon, Chatterji, Chen, Creel, Davis, Demszky, Donahue, Doumbouya, Durmus, Ermon, Etchemendy, Ethayarajh, Fei-Fei, Finn, Gale, Gillespie, Goel, Goodman, Grossman, Guha, Hashimoto, Henderson, Hewitt, Ho, Hong, Hsu, Huang, Icard, Jain, Jurafsky, Kalluri, Karamcheti, Keeling, Khani, Khattab, Koh, Krass, Krishna, Kuditipudi, Kumar, Ladhak, Lee, Lee, Leskovec, Levent, Li, Li, Ma, Malik, Manning, Mirchandani, Mitchell, Munyikwa, Nair, Narayan, Narayanan, Newman, Nie, Niebles, Nilforoshan, Nyarko, Ogut, Orr, Papadimitriou, Park, Piech, Portelance, Potts, Raghunathan, Reich, Ren, Rong, Roohani, Ruiz, Ryan, Ré, Sadigh, Sagawa, Santhanam, Shih, Srinivasan, Tamkin, Taori, Thomas, Tramèr, Wang, Wang, Wu, Wu, Wu, Xie, Yasunaga, You, Zaharia, Zhang, Zhang, Zhang, Zhang, Zheng, Zhou, and Liang]{bommasani2022opportunitiesrisksfoundationmodels}
Rishi Bommasani, Drew~A. Hudson, Ehsan Adeli, Russ Altman, Simran Arora, Sydney von Arx, Michael~S. Bernstein, Jeannette Bohg, Antoine Bosselut, Emma Brunskill, Erik Brynjolfsson, Shyamal Buch, Dallas Card, Rodrigo Castellon, Niladri Chatterji, Annie Chen, Kathleen Creel, Jared~Quincy Davis, Dora Demszky, Chris Donahue, Moussa Doumbouya, Esin Durmus, Stefano Ermon, John Etchemendy, Kawin Ethayarajh, Li~Fei-Fei, Chelsea Finn, Trevor Gale, Lauren Gillespie, Karan Goel, Noah Goodman, Shelby Grossman, Neel Guha, Tatsunori Hashimoto, Peter Henderson, John Hewitt, Daniel~E. Ho, Jenny Hong, Kyle Hsu, Jing Huang, Thomas Icard, Saahil Jain, Dan Jurafsky, Pratyusha Kalluri, Siddharth Karamcheti, Geoff Keeling, Fereshte Khani, Omar Khattab, Pang~Wei Koh, Mark Krass, Ranjay Krishna, Rohith Kuditipudi, Ananya Kumar, Faisal Ladhak, Mina Lee, Tony Lee, Jure Leskovec, Isabelle Levent, Xiang~Lisa Li, Xuechen Li, Tengyu Ma, Ali Malik, Christopher~D. Manning, Suvir Mirchandani, Eric Mitchell, Zanele Munyikwa, Suraj Nair,
  Avanika Narayan, Deepak Narayanan, Ben Newman, Allen Nie, Juan~Carlos Niebles, Hamed Nilforoshan, Julian Nyarko, Giray Ogut, Laurel Orr, Isabel Papadimitriou, Joon~Sung Park, Chris Piech, Eva Portelance, Christopher Potts, Aditi Raghunathan, Rob Reich, Hongyu Ren, Frieda Rong, Yusuf Roohani, Camilo Ruiz, Jack Ryan, Christopher Ré, Dorsa Sadigh, Shiori Sagawa, Keshav Santhanam, Andy Shih, Krishnan Srinivasan, Alex Tamkin, Rohan Taori, Armin~W. Thomas, Florian Tramèr, Rose~E. Wang, William Wang, Bohan Wu, Jiajun Wu, Yuhuai Wu, Sang~Michael Xie, Michihiro Yasunaga, Jiaxuan You, Matei Zaharia, Michael Zhang, Tianyi Zhang, Xikun Zhang, Yuhui Zhang, Lucia Zheng, Kaitlyn Zhou, and Percy Liang.
\newblock On the opportunities and risks of foundation models, 2022.
\newblock URL \url{https://arxiv.org/abs/2108.07258}.

\bibitem[Bonilla et~al.(2007)Bonilla, Chai, and Williams]{bonilla2007linearcoregionalization}
Edwin~V Bonilla, Kian Chai, and Christopher Williams.
\newblock Multi-task gaussian process prediction.
\newblock \emph{Advances in neural information processing systems}, 20, 2007.

\bibitem[Brodnik et~al.(2023)Brodnik, Muir, Tulshibagwale, Rossin, Echlin, Hamel, Kramer, Pollock, Kiser, Smith, and Daly]{brodniksurvey}
N.R. Brodnik, C.~Muir, N.~Tulshibagwale, J.~Rossin, M.P. Echlin, C.M. Hamel, S.L.B. Kramer, T.M. Pollock, J.D. Kiser, C.~Smith, and S.H. Daly.
\newblock Perspective: Machine learning in experimental solid mechanics.
\newblock \emph{Journal of the Mechanics and Physics of Solids}, 173:\penalty0 105231, 2023.
\newblock \doi{10.1016/j.jmps.2023.105231}.

\bibitem[Brown et~al.(2020)Brown, Mann, Ryder, Subbiah, Kaplan, Dhariwal, Neelakantan, Shyam, Sastry, Askell, et~al.]{brown2020language}
Tom Brown, Benjamin Mann, Nick Ryder, Melanie Subbiah, Jared~D Kaplan, Prafulla Dhariwal, Arvind Neelakantan, Pranav Shyam, Girish Sastry, Amanda Askell, et~al.
\newblock Language models are few-shot learners.
\newblock \emph{Advances in neural information processing systems}, 33:\penalty0 1877--1901, 2020.

\bibitem[Bunge(1982)]{bunge}
H-J Bunge.
\newblock \emph{Texture Analysis in Materials Science: Mathematic Methods}.
\newblock Butterworth \& Co., Berlin, 1982.

\bibitem[Buzzy et~al.(2024)Buzzy, Robertson, and Kalidindi]{buzzy2024statistically}
Michael~O Buzzy, Andreas~E Robertson, and Surya~R Kalidindi.
\newblock Statistically conditioned polycrystal generation using denoising diffusion models.
\newblock \emph{Acta Materialia}, page 119746, 2024.

\bibitem[Buzzy et~al.(2025)Buzzy, de~Oca~Zapiain, Generale, Kalidindi, and Lim]{buzzy2025active}
Michael~O Buzzy, David~Montes de~Oca~Zapiain, Adam~P Generale, Surya~R Kalidindi, and Hojun Lim.
\newblock Active learning for the design of polycrystalline textures using conditional normalizing flows.
\newblock \emph{Acta Materialia}, 284:\penalty0 120537, 2025.

\bibitem[Chapman et~al.(2021)Chapman, Shah, Donegan, Scott, Shade, Menasche, and Uchic]{chapman2021afrl}
Michael~G Chapman, Megna~N Shah, Sean~P Donegan, J~Michael Scott, Paul~A Shade, David Menasche, and Michael~D Uchic.
\newblock Afrl additive manufacturing modeling series: challenge 4, 3d reconstruction of an in625 high-energy diffraction microscopy sample using multi-modal serial sectioning.
\newblock \emph{Integrating Materials and Manufacturing Innovation}, 10:\penalty0 129--141, 2021.

\bibitem[Cheng et~al.(2022)Cheng, Jiao, and Ren]{threepoint_bayesiangen}
S.~Cheng, Y.~Jiao, and Y.~Ren.
\newblock Data-driven learning of 3-point correlation functions as microstructure representations.
\newblock \emph{Acta Materialia}, 229:\penalty0 117800, 2022.
\newblock \doi{10.1016/j.actamat.2022.117800}.

\bibitem[Dalla-Torre et~al.(2024)Dalla-Torre, Gonzalez, Mendoza-Revilla, Lopez~Carranza, Grzywaczewski, Oteri, Dallago, Trop, de~Almeida, Sirelkhatim, et~al.]{dalla2024nucleotide}
Hugo Dalla-Torre, Liam Gonzalez, Javier Mendoza-Revilla, Nicolas Lopez~Carranza, Adam~Henryk Grzywaczewski, Francesco Oteri, Christian Dallago, Evan Trop, Bernardo~P de~Almeida, Hassan Sirelkhatim, et~al.
\newblock Nucleotide transformer: building and evaluating robust foundation models for human genomics.
\newblock \emph{Nature Methods}, pages 1--11, 2024.

\bibitem[DeCost and Holm(2016)]{holm_powderdatabase}
Brain~L. DeCost and Elizabeth~A. Holm.
\newblock A large dataset of synthetic sem images of powder materials and their ground truth 3d structures.
\newblock \emph{Data in Brief}, 9:\penalty0 727--731, 2016.
\newblock \doi{10.1016/j.dib.2016.10.011}.

\bibitem[DeCost and Holm(2015)]{holm_decost_classification}
Brian~L. DeCost and Elizabeth~A. Holm.
\newblock A computer vision approach for automated analysis and classification of microstructural images.
\newblock \emph{Computational Materials Science}, 110:\penalty0 126--133, 2015.
\newblock \doi{10.1016/j.commatsci.2015.08.011}.

\bibitem[DeCost et~al.(2017)DeCost, Hecht, Francis, Webler, Picard, and Holm]{carbonsteel_database}
Brian~L. DeCost, Matthew Hecht, Toby Francis, Bryan~A. Webler, Yoosuf~N. Picard, and Elizabeth Holm.
\newblock Uhcsdb: Ultra high carbon steel micrograph database.
\newblock \emph{Integrating Materials and Manufacturing Innovation}, 6:\penalty0 197--205, 2017.
\newblock \doi{10.1007/s40192-017-0097-0}.

\bibitem[Devlin et~al.(2019)Devlin, Chang, Lee, and Toutanova]{devlin2019bertpretrainingdeepbidirectional}
Jacob Devlin, Ming-Wei Chang, Kenton Lee, and Kristina Toutanova.
\newblock Bert: Pre-training of deep bidirectional transformers for language understanding, 2019.
\newblock URL \url{https://arxiv.org/abs/1810.04805}.

\bibitem[Dimiduk et~al.(2018)Dimiduk, Holm, and Niezgoda]{neizgodasurvey}
D.M. Dimiduk, E.A. Holm, and S.R. Niezgoda.
\newblock Perspective on the impact of machine learning, deep learning, and artificial intelligence on materials, processes, and structures engineering.
\newblock \emph{Integrating Materials and Manufacturing Innovation}, 7:\penalty0 157--172, 2018.
\newblock \doi{10.1007/s40192-018-0117-8}.

\bibitem[Dureth et~al.(2022)Dureth, Seibert, Rucker, Handford, Kastner, and Gude]{seibertDiffusion}
C.~Dureth, P.~Seibert, D.~Rucker, S.~Handford, M.~Kastner, and M.~Gude.
\newblock Conditional diffusion-based microstructure reconstruction.
\newblock \emph{ArXiV}, 2022.

\bibitem[Falco et~al.(2017)Falco, Jiang, Cola, and Petrinic]{voronoi_laguerre}
S.~Falco, J.~Jiang, F.~De Cola, and N.~Petrinic.
\newblock Generation of 3d polycrystalline microstructures with a conditioned laguerre-voronoi tessellation technique.
\newblock \emph{Computational Materials Science}, 136:\penalty0 20--28, 2017.
\newblock \doi{10.1016/j.commatsci.2017.04.018}.

\bibitem[Fowler et~al.(2024)Fowler, Ruggles, Cillessen, Johnson, Jauregui, Craig, Bianco, Henriksen, and Boyce]{fowler2024high}
J~Elliott Fowler, Timothy~J Ruggles, Dale~E Cillessen, Kyle~L Johnson, Luis~J Jauregui, Robert~L Craig, Nathan~R Bianco, Amelia~A Henriksen, and Brad~L Boyce.
\newblock High-throughput microstructural characterization and process correlation using automated electron backscatter diffraction.
\newblock \emph{Integrating Materials and Manufacturing Innovation}, 13\penalty0 (3):\penalty0 641--655, 2024.

\bibitem[Fullwood et~al.(2008)Fullwood, Adams, and Kalidindi]{fullwoodpolycrystalSCE}
D.T. Fullwood, B.L. Adams, and S.R. Kalidindi.
\newblock A strong contrast homogenization formulation for multi-phase anistropic materials.
\newblock \emph{Journal of the Mechanics and Physics of Solids}, 56:\penalty0 2287--2297, 2008.

\bibitem[Gao and Liu(2021)]{gao_stochasticdesign}
Y.~Gao and Y.~Liu.
\newblock Relibaility-based topology optimization with stochastic heterogeneous microstructure properties.
\newblock \emph{Materials and Design}, 2021.
\newblock \doi{10.1016/j.matdes.2021.109713}.

\bibitem[Generale et~al.(2023{\natexlab{a}})Generale, Kelly, Harrington, Robertson, Buzzy, and Kalidindi]{generale2023a}
Adam Generale, Conlain Kelly, Grayson Harrington, Andreas Robertson, Michael Buzzy, and Surya Kalidindi.
\newblock A bayesian approach to designing microstructures and processing pathways for tailored material properties.
\newblock In \emph{AI for Accelerated Materials Design - NeurIPS 2023 Workshop}, 2023{\natexlab{a}}.
\newblock URL \url{https://openreview.net/forum?id=zZPICTs5gB}.

\bibitem[Generale and Kalidindi(2021)]{adam}
A.P. Generale and S.R. Kalidindi.
\newblock Reduced-order models for microstructure-sensitive effective thermal conductivity of woven ceramic matrix composites with residual porosity.
\newblock \emph{Composite Structures}, 274:\penalty0 114399, 2021.
\newblock \doi{10.1016/j.compstruct.2021.114399}.

\bibitem[Generale et~al.(2023{\natexlab{b}})Generale, Robertson, Kelly, and Kalidindi]{adam_ISMD}
A.P. Generale, A.E. Robertson, C.~Kelly, and S.R. Kalidindi.
\newblock Inverse stochastic microstructure design.
\newblock \emph{SSRN: Preprint}, 2023{\natexlab{b}}.
\newblock \doi{10.2139/ssrn.4590691}.

\bibitem[Ghosh et~al.(2021)Ghosh, Shen, Kotha, and Chakraborty]{ghosh2021watmus}
Somnath Ghosh, Jinlei Shen, Shravan Kotha, and Pritam Chakraborty.
\newblock Watmus: Wavelet transformation-induced multi-time scaling for accelerating fatigue simulations at multiple spatial scales.
\newblock \emph{Integrating Materials and Manufacturing Innovation}, 10:\penalty0 568--587, 2021.

\bibitem[Groeber and Jackson(2014)]{dream3d}
M.A. Groeber and M.A. Jackson.
\newblock Dream.3d: A digital representation environment for the analysis of microstructure in 3d.
\newblock \emph{Integrating Materials and Manufacturing Innovation}, 3:\penalty0 56--72, 2014.
\newblock \doi{10.1186/2193-9772-3-5}.

\bibitem[Groeber et~al.(2008)Groeber, Ghosh, Uchic, and Dimiduk]{dream3d_generation}
M.A. Groeber, S.~Ghosh, M.D. Uchic, and D.M. Dimiduk.
\newblock A framework for automated analysis and simulation of 3d polycrystalline microstructures. part 2: Synthetic microstructure generation.
\newblock \emph{Acta Materialia}, 56:\penalty0 1274--1287, 2008.
\newblock \doi{10.1016/j.actamat.2007.11.040}.

\bibitem[Hashemi and Kalidindi(2021)]{polycrysEvolution}
S.~Hashemi and S.R. Kalidindi.
\newblock A machine learning framework for the temporal evolution of microstructure during static recrystallization of polycrystalline materials simulated by cellular automaton.
\newblock \emph{Computational Materials Science}, 188:\penalty0 110132, 2021.
\newblock \doi{10.1016/j.commatsci.2020.110132}.

\bibitem[Hashemi and Kalidindi(2023)]{sepidifferential}
Sepideh Hashemi and Surya~R. Kalidindi.
\newblock Gaussian process autoregression models for the evolution of polycrystalline microstructures subjected to arbitrary stretching tensors.
\newblock \emph{International Journal of Plasticity}, 162:\penalty0 103532, 2023.
\newblock \doi{10.1016/j.ijplas.2023.103532}.

\bibitem[Ho and Salimans(2022)]{learnedconditionedDDPM}
J.~Ho and T.~Salimans.
\newblock Classifier-free diffusion generation.
\newblock \emph{ArXiV}, 2022.
\newblock \doi{10.48550/arXiv.2207.12598}.

\bibitem[Ho et~al.(2020)Ho, Jain, and Abbeel]{ho}
J.~Ho, A.~Jain, and P.~Abbeel.
\newblock Denoising diffusion probabilistic models.
\newblock \emph{NeurIPS}, 2020.

\bibitem[Jakubik et~al.(2025)Jakubik, Yang, Blumenstiel, Scheurer, Sedona, Maurogiovanni, Bosmans, Dionelis, Marsocci, Kopp, et~al.]{jakubik2025terramind}
Johannes Jakubik, Felix Yang, Benedikt Blumenstiel, Erik Scheurer, Rocco Sedona, Stefano Maurogiovanni, Jente Bosmans, Nikolaos Dionelis, Valerio Marsocci, Niklas Kopp, et~al.
\newblock Terramind: Large-scale generative multimodality for earth observation.
\newblock \emph{arXiv preprint arXiv:2504.11171}, 2025.

\bibitem[Jangid et~al.(2024)Jangid, Brodnik, Echlin, Gudavalli, Levenson, Pollock, Daly, and Manjunath]{jangid2024q}
Devendra~K Jangid, Neal~R Brodnik, McLean~P Echlin, Chandrakanth Gudavalli, Connor Levenson, Tresa~M Pollock, Samantha~H Daly, and BS~Manjunath.
\newblock Q-rbsa: high-resolution 3d ebsd map generation using an efficient quaternion transformer network.
\newblock \emph{npj Computational Materials}, 10\penalty0 (1):\penalty0 27, 2024.

\bibitem[Jangid et~al.(2023)Jangid, Brodnik, Echlin, Daly, Polloc, and Manjunath]{jangid2023titanium}
Devendra~Kumar Jangid, Neal~R Brodnik, McLean~P Echlin, Samantha Daly, Tresa Polloc, and BS~Manjunath.
\newblock Titanium 3d microstructure for physics-based generative models: a dataset and primer.
\newblock In \emph{1st Workshop on the Synergy of Scientific and Machine Learning Modeling@ ICML 2023}. https://openreview. net/, 2023.

\bibitem[Jr.(1955)]{brownSCE}
W.F.~Brown Jr.
\newblock Solid mixture permittivities.
\newblock \emph{Journal of Chemical Physics}, 23:\penalty0 1514--1517, 1955.

\bibitem[Jumper et~al.(2021)Jumper, Evans, Pritzel, Green, Figurnov, Ronneberger, Tunyasuvunakool, Bates, {\v{Z}}{\'\i}dek, Potapenko, et~al.]{jumper2021highly}
John Jumper, Richard Evans, Alexander Pritzel, Tim Green, Michael Figurnov, Olaf Ronneberger, Kathryn Tunyasuvunakool, Russ Bates, Augustin {\v{Z}}{\'\i}dek, Anna Potapenko, et~al.
\newblock Highly accurate protein structure prediction with alphafold.
\newblock \emph{nature}, 596\penalty0 (7873):\penalty0 583--589, 2021.

\bibitem[Jung et~al.(2020)Jung, Yoon, Park, Jo, and Kim]{jong_design_latentspaces}
Jaimyun Jung, Jae~Ik Yoon, Hyung~Keun Park, Hyeontae Jo, and Hyoung~Seop Kim.
\newblock Microstructure design using machine learning generated low dimensional and continuous design space.
\newblock \emph{Materialia}, 11:\penalty0 100690, 2020.
\newblock \doi{https://doi.org/10.1016/j.mtla.2020.100690}.

\bibitem[Karras et~al.(2022)Karras, Aittala, Aila, and Laine]{karras2022elucidating}
Tero Karras, Miika Aittala, Timo Aila, and Samuli Laine.
\newblock Elucidating the design space of diffusion-based generative models, 2022.

\bibitem[Kench and Cooper(2021)]{2Dto3DGAN}
S.~Kench and S.J. Cooper.
\newblock Generating three-dimensional structures from a two-dimensional slice with generative adversarial network-based dimensionality expansion.
\newblock \emph{Nature Machine Intelligence}, 3:\penalty0 299--305, 2021.
\newblock \doi{10.1038/s42256-021-00322-1}.

\bibitem[Kovachki et~al.(2023)Kovachki, Li, Liu, Azizzadenesheli, Bhattacharya, Stuart, and Anandkumar]{kovachki2023neural}
Nikola Kovachki, Zongyi Li, Burigede Liu, Kamyar Azizzadenesheli, Kaushik Bhattacharya, Andrew Stuart, and Anima Anandkumar.
\newblock Neural operator: Learning maps between function spaces with applications to pdes.
\newblock \emph{Journal of Machine Learning Research}, 24\penalty0 (89):\penalty0 1--97, 2023.

\bibitem[Krishnamoorthi et~al.(2023)Krishnamoorthi, Bandyopadhyay, and Sangid]{titanium_generator}
Sidharth Krishnamoorthi, Ritwik Bandyopadhyay, and Michael~D. Sangid.
\newblock A microstructure-based fatigue model for additively manufactured ti-6al-4v, including the role of prior $\beta$ boundaries.
\newblock \emph{International Journal of Plasticity}, 163:\penalty0 103569, 2023.
\newblock \doi{https://doi.org/10.1016/j.ijplas.2023.103569}.

\bibitem[Lee and Yun(2024)]{lee2024multi}
Kang-Hyun Lee and Gun~Jin Yun.
\newblock Multi-plane denoising diffusion-based dimensionality expansion for 2d-to-3d reconstruction of microstructures with harmonized sampling.
\newblock \emph{npj Computational Materials}, 10\penalty0 (1):\penalty0 99, 2024.

\bibitem[Leung and Bovy(2023)]{Leung_2023}
Henry~W Leung and Jo~Bovy.
\newblock Towards an astronomical foundation model for stars with a transformer-based model.
\newblock \emph{Monthly Notices of the Royal Astronomical Society}, 527\penalty0 (1):\penalty0 1494–1520, October 2023.
\newblock ISSN 1365-2966.
\newblock \doi{10.1093/mnras/stad3015}.
\newblock URL \url{http://dx.doi.org/10.1093/mnras/stad3015}.

\bibitem[Lim et~al.(2023)Lim, Kovachki, Baptista, Beckham, Azizzadenesheli, Kossaifi, Voleti, Song, Kreis, Kautz, et~al.]{lim2023score}
Jae~Hyun Lim, Nikola~B Kovachki, Ricardo Baptista, Christopher Beckham, Kamyar Azizzadenesheli, Jean Kossaifi, Vikram Voleti, Jiaming Song, Karsten Kreis, Jan Kautz, et~al.
\newblock Score-based diffusion models in function space.
\newblock \emph{arXiv preprint arXiv:2302.07400}, 2023.

\bibitem[Mahbub et~al.(2017)Mahbub, Hsu, Epting, Nuhfer, Hackett, Abernathy, Rollett, Graef, Litster, and Salvador]{sofc_experimental_database}
Rubayyat Mahbub, Tim Hsu, William~K. Epting, Noel~T. Nuhfer, Gregory~A. Hackett, Harry Abernathy, Anthony~D. Rollett, Marc~De Graef, Shawn Litster, and Paul~A. Salvador.
\newblock A method for quantitative 3d mesoscale analysis of solid oxide fuel cell microstructures using xe-plasma focused ion beam coupled with sem.
\newblock \emph{ECS Transactions}, 78:\penalty0 2159--2170, 2017.
\newblock \doi{10.1149/07801.2159ecst}.

\bibitem[Marshall and Kalidindi(2021)]{marshallmeanfield}
A.~Marshall and S.R. Kalidindi.
\newblock Autonomous development of a machine-learning model for the plastic response of two-phase composites from micromechanical finite element models.
\newblock \emph{JOM}, 73:\penalty0 2085--2095, 2021.
\newblock \doi{10.1007/s11837-021-04696-w}.

\bibitem[McDowell(2008)]{mcdowellviscoplast}
David~L. McDowell.
\newblock Viscoplasticity of heterogeneous metallic materials.
\newblock \emph{Materials Science and Engineering: R: Reports}, 62\penalty0 (3):\penalty0 67--123, 2008.
\newblock \doi{https://doi.org/10.1016/j.mser.2008.04.003}.

\bibitem[Meng et~al.(2022)Meng, He, Song, Song, Wu, Zhu, and Ermon]{sdedit}
C.~Meng, Y.~He, Y.~Song, J.~Song, J.~Wu, J.~Zhu, and S.~Ermon.
\newblock Sdedit: Guided image synthesis and editing with stochastic differential equations.
\newblock \emph{ArXiv}, 2022.

\bibitem[Mizrahi et~al.(2023)Mizrahi, Bachmann, Kar, Yeo, Gao, Dehghan, and Zamir]{mizrahi20234m}
David Mizrahi, Roman Bachmann, Oguzhan Kar, Teresa Yeo, Mingfei Gao, Afshin Dehghan, and Amir Zamir.
\newblock 4m: Massively multimodal masked modeling.
\newblock \emph{Advances in Neural Information Processing Systems}, 36:\penalty0 58363--58408, 2023.

\bibitem[Ohana et~al.(2024)Ohana, McCabe, Meyer, Morel, Agocs, Beneitez, Berger, Burkhart, Dalziel, Fielding, et~al.]{ohana2024well}
Ruben Ohana, Michael McCabe, Lucas Meyer, Rudy Morel, Fruzsina Agocs, Miguel Beneitez, Marsha Berger, Blakesly Burkhart, Stuart Dalziel, Drummond Fielding, et~al.
\newblock The well: a large-scale collection of diverse physics simulations for machine learning.
\newblock \emph{Advances in Neural Information Processing Systems}, 37:\penalty0 44989--45037, 2024.

\bibitem[Oommen et~al.(2024)Oommen, Bora, Zhang, and Karniadakis]{oommen2024integrating}
Vivek Oommen, Aniruddha Bora, Zhen Zhang, and George~Em Karniadakis.
\newblock Integrating neural operators with diffusion models improves spectral representation in turbulence modeling.
\newblock \emph{arXiv preprint arXiv:2409.08477}, 2024.

\bibitem[Parker et~al.(2024)Parker, Lanusse, Golkar, Sarra, Cranmer, Bietti, Eickenberg, Krawezik, McCabe, Morel, et~al.]{parker2024astroclip}
Liam Parker, Francois Lanusse, Siavash Golkar, Leopoldo Sarra, Miles Cranmer, Alberto Bietti, Michael Eickenberg, Geraud Krawezik, Michael McCabe, Rudy Morel, et~al.
\newblock Astroclip: a cross-modal foundation model for galaxies.
\newblock \emph{Monthly Notices of the Royal Astronomical Society}, 531\penalty0 (4):\penalty0 4990--5011, 2024.

\bibitem[Parra and Tobar(2017)]{parra2017spectralmixturekernelsmultioutput}
Gabriel Parra and Felipe Tobar.
\newblock Spectral mixture kernels for multi-output gaussian processes, 2017.
\newblock URL \url{https://arxiv.org/abs/1709.01298}.

\bibitem[Paulson et~al.(2017)Paulson, Priddy, McDowell, and Kalidindi]{polycrystal2ps}
N.H. Paulson, M.W. Priddy, D.L. McDowell, and S.R. Kalidindi.
\newblock Reduced-order structure-property linkages for polycrystalline microstructures based on 2-point statistics.
\newblock \emph{Acta Mater.}, 129:\penalty0 428, 2017.
\newblock \doi{10.1016/j.actamat.2017.03.009}.

\bibitem[Phan et~al.(2024)Phan, Sarmad, Ruspini, Kiss, and Lindseth]{phan2024generating}
Johan Phan, Muhammad Sarmad, Leonardo Ruspini, Gabriel Kiss, and Frank Lindseth.
\newblock Generating 3d images of material microstructures from a single 2d image: a denoising diffusion approach.
\newblock \emph{Scientific Reports}, 14\penalty0 (1):\penalty0 6498, 2024.

\bibitem[Pilchak et~al.(2016)Pilchak, Shank, Tucker, Srivatsa, Fagin, and Semiatin]{titanium_database}
Adam~L. Pilchak, Jared Shank, Joseph~C. Tucker, Shesh Srivatsa, Patrick~N. Fagin, and S.~Lee Semiatin.
\newblock A dataset for the development, verification, and validation of microstructure-sensitive process models for near-alpha titanium alloys.
\newblock \emph{Integrating Materials and Manufacturing Innovation}, pages 1--18, 2016.
\newblock \doi{10.1186/s40192-016-0056-1}.

\bibitem[Prasad et~al.(2019)Prasad, Vajragupta, and Hartmaier]{kanapy}
M.R.G. Prasad, N.~Vajragupta, and A.~Hartmaier.
\newblock Kanapy: A python package for generating complex synthetic polycrystalline microstructures.
\newblock \emph{Journal of Open Source Software}, 4:\penalty0 1732, 2019.
\newblock \doi{10.21105/joss.01732}.

\bibitem[Qiu et~al.(2024)Qiu, Bridges, and Chen]{qiu2024derivative}
Yuan Qiu, Nolan Bridges, and Peng Chen.
\newblock Derivative-enhanced deep operator network.
\newblock \emph{Advances in Neural Information Processing Systems}, 37:\penalty0 20945--20981, 2024.

\bibitem[Riaz et~al.(2025)Riaz, Bhabesh, Arannil, Ballesteros, and Horwood]{riaz2025metasynth}
Haris Riaz, Sourav Bhabesh, Vinayak Arannil, Miguel Ballesteros, and Graham Horwood.
\newblock Metasynth: Meta-prompting-driven agentic scaffolds for diverse synthetic data generation.
\newblock \emph{arXiv preprint arXiv:2504.12563}, 2025.

\bibitem[Robertson and Kalidindi(2022)]{andreas_NGRF}
A.E. Robertson and S.R. Kalidindi.
\newblock Efficient generation of n-field microstructures from 2-point statistics using multi-output gaussian random fields.
\newblock \emph{Acta Materialia}, 232:\penalty0 117927, 2022.
\newblock \doi{10.1016/j.actamat.2022.117927}.

\bibitem[Robertson et~al.(2023)Robertson, Kelly, Buzzy, and Kalidindi]{andreas_LGD}
Andreas~E. Robertson, Conlain Kelly, Michael Buzzy, and Surya~R. Kalidindi.
\newblock Local–global decompositions for conditional microstructure generation.
\newblock \emph{Acta Materialia}, 253:\penalty0 118966, 2023.
\newblock ISSN 1359-6454.
\newblock \doi{https://doi.org/10.1016/j.actamat.2023.118966}.

\bibitem[Robertson et~al.(2024)Robertson, Generale, Kelly, Buzzy, and Kalidindi]{robertson2024micro2d}
Andreas~E Robertson, Adam~P Generale, Conlain Kelly, Michael~O Buzzy, and Surya~R Kalidindi.
\newblock Micro2d: A large, statistically diverse, heterogeneous microstructure dataset.
\newblock \emph{Integrating Materials and Manufacturing Innovation}, pages 1--35, 2024.

\bibitem[Rodgers et~al.(2017)Rodgers, Mitchell, and Tikare]{rodgers2017monte}
Theron~M Rodgers, John~A Mitchell, and Veena Tikare.
\newblock A monte carlo model for 3d grain evolution during welding.
\newblock \emph{Modelling and Simulation in Materials Science and Engineering}, 25\penalty0 (6):\penalty0 064006, 2017.

\bibitem[Rossin et~al.(2022)Rossin, Leser, Pusch, Frey, Vogel, Saville, Torbet, Clarke, Daly, and Pollock]{rossin_nondestructive}
J.~Rossin, P.~Leser, K.~Pusch, C.~Frey, S.C. Vogel, A.I. Saville, C.~Torbet, A.J. Clarke, S.~Daly, and T.M. Pollock.
\newblock Single crystal elastic constants of additively manufactured components determined by resonant ultrasound spectroscopy.
\newblock \emph{Materials Characterization}, 192:\penalty0 112244, 2022.
\newblock \doi{10.1016/j.matchar.2022.112244}.

\bibitem[Roters et~al.(2010)Roters, Eisenlohr, Hantcherli, Tjahjanto, Bieler, and Raabe]{cpfem_raabe}
F.~Roters, P.~Eisenlohr, L.~Hantcherli, D.D. Tjahjanto, T.R. Bieler, and D.~Raabe.
\newblock Overview of constitutive laws, kinematics, homogenization and multiscale methods in crystal plasticity finite element modeling: Theory, experiments, applications.
\newblock \emph{Acta Materialia}, 58:\penalty0 1152, 2010.

\bibitem[Rowenhorst et~al.(2010)Rowenhorst, Lewis, and Spanos]{rowenhorst2010three}
DJ~Rowenhorst, AC~Lewis, and G~Spanos.
\newblock Three-dimensional analysis of grain topology and interface curvature in a $\beta$-titanium alloy.
\newblock \emph{Acta Materialia}, 58\penalty0 (16):\penalty0 5511--5519, 2010.

\bibitem[Safdari et~al.(2012)Safdari, Baniassadi, Garmestani, and Al-Haik]{garmestaniSCE}
M.~Safdari, M.~Baniassadi, H.~Garmestani, and M.S. Al-Haik.
\newblock A modified strong-constrast expansion for estimating the effective thermal conductivity of multiphase heterogeneous materials.
\newblock \emph{Journal of Applied Physics}, 112:\penalty0 114318, 2012.

\bibitem[Samo and Roberts(2015)]{samo2015generalizedspectralkernels}
Yves-Laurent~Kom Samo and Stephen Roberts.
\newblock Generalized spectral kernels, 2015.
\newblock URL \url{https://arxiv.org/abs/1506.02236}.

\bibitem[Seibert et~al.(2021)Seibert, Ambati, Rabloff, and Kastner]{marreddy_2D}
P.~Seibert, M.~Ambati, A.~Rabloff, and M.~Kastner.
\newblock Reconstructing random heterogeneous media through differentiable optimization.
\newblock \emph{Computational Materials Science}, 196:\penalty0 110455, 2021.
\newblock \doi{10.1016/j.commatsci.2021.110455}.

\bibitem[Seibert et~al.(2022)Seibert, Rabloff, Ambati, and Kastner]{marreddy_3D}
P.~Seibert, A.~Rabloff, M.~Ambati, and M.~Kastner.
\newblock Descriptor-based reconstruction of three-dimensional microstructures through gradient-based optimization.
\newblock \emph{Acta Materialia}, 227:\penalty0 117667, 2022.
\newblock \doi{10.1016/j.actamat.2022.117667}.

\bibitem[Seibert et~al.(2023{\natexlab{a}})Seibert, Husert, Wollner, Kalina, and Kastner]{seibert_localoptimization}
P.~Seibert, M.~Husert, M.P. Wollner, K.A. Kalina, and M.~Kastner.
\newblock Fast reconstruction of microstructures with ellipsoidal inclusions using analytic descriptors.
\newblock \emph{ArXiv}, 2023{\natexlab{a}}.
\newblock \doi{10.48550/arXiv.2306.08316}.

\bibitem[Seibert et~al.(2023{\natexlab{b}})Seibert, Ra{\ss}loff, Kalina, Gussone, Bugelnig, Diehl, and K{\"a}stner]{seibert2023two}
Paul Seibert, Alexander Ra{\ss}loff, Karl~A Kalina, Joachim Gussone, Katrin Bugelnig, Martin Diehl, and Markus K{\"a}stner.
\newblock Two-stage 2d-to-3d reconstruction of realistic microstructures: Implementation and numerical validation by effective properties.
\newblock \emph{Computer Methods in Applied Mechanics and Engineering}, 412:\penalty0 116098, 2023{\natexlab{b}}.

\bibitem[Song and Yan(2013)]{neu_steel_defect_database}
Kechen Song and Yunhui Yan.
\newblock A noise robust method based on completed local binary patterns for hot-rolled steel strip surface defects.
\newblock \emph{Applied Surface Science}, 285P:\penalty0 858--864, 2013.
\newblock \doi{10.1016/j.apsusc.2013.09.002}.

\bibitem[Song et~al.(2023)Song, Dhariwal, Chen, and Sutskever]{song2023consistency}
Yang Song, Prafulla Dhariwal, Mark Chen, and Ilya Sutskever.
\newblock Consistency models.
\newblock 2023.

\bibitem[Stinville et~al.(2022)Stinville, Hestroffer, Charpagne, Polonsky, Echlin, Torbet, Valle, Nygren, Miller, Klaas, et~al.]{stinville2022multi}
Jean-Charles Stinville, JM~Hestroffer, Marie-Agathe Charpagne, AT~Polonsky, MP~Echlin, CJ~Torbet, Val{\'e}ry Valle, KE~Nygren, MP~Miller, Ottmar Klaas, et~al.
\newblock Multi-modal dataset of a polycrystalline metallic material: 3d microstructure and deformation fields.
\newblock \emph{Scientific Data}, 9\penalty0 (1):\penalty0 460, 2022.

\bibitem[Sun et~al.(2024)Sun, Wu, Chen, Feng, and Bouman]{sun2024provable}
Yu~Sun, Zihui Wu, Yifan Chen, Berthy~T Feng, and Katherine~L Bouman.
\newblock Provable probabilistic imaging using score-based generative priors.
\newblock \emph{IEEE Transactions on Computational Imaging}, 2024.

\bibitem[Torquato(1997)]{torquatov1SCE}
S.~Torquato.
\newblock Effective stiffness tensor of composite media: 1. exact series expansions.
\newblock \emph{Journal of the Mechanics and Physics of Solids}, 45:\penalty0 1421--1448, 1997.

\bibitem[Torquato(2002)]{torquato}
S.~Torquato.
\newblock \emph{Random Heterogeneous Materials}.
\newblock Springer, New York, NY, 2002.

\bibitem[Turner and Kalidindi(2016)]{markovturner}
D.~Turner and S.~R. Kalidindi.
\newblock Statistical construction of 3d microstructures from 2-d examplars collected on oblique sections.
\newblock \emph{Acta Materialia}, 102:\penalty0 136--148, 2016.
\newblock \doi{10.1016/j.actamat.2015.09.011}.

\bibitem[Vaughan et~al.(2024)Vaughan, Lim, Pham, Seede, Polonsky, Johnson, and Noell]{vaughan2024mechanistic}
MW~Vaughan, H~Lim, B~Pham, R~Seede, AT~Polonsky, KL~Johnson, and PJ~Noell.
\newblock The mechanistic origins of heterogeneous void growth during ductile failure.
\newblock \emph{Acta Materialia}, 274:\penalty0 119977, 2024.

\bibitem[Vlassis and Sun(2023)]{vlassis_diffusion_design}
Nikolaos~N. Vlassis and WaiChing Sun.
\newblock Denoising diffusion algorithm for inverse design of microstructures with fine-tuned nonlinear material properties.
\newblock \emph{Computer Methods in Applied Mechanics and Engineering}, 413:\penalty0 116126, 2023.
\newblock \doi{https://doi.org/10.1016/j.cma.2023.116126}.

\bibitem[Wilson and Adams(2013)]{wilson_2013_GaussianSpectralMixture}
A.G. Wilson and R.P. Adams.
\newblock Gaussian process kernels for pattern discovery and extrapolation.
\newblock In \emph{Proceedings of the 30th International Conference on Machine Learning}, volume~28 of \emph{Proceedings of Machine Learning Research}, pages 1067--1075. PMLR, 2013.

\bibitem[Xia et~al.(2024)Xia, Wu, Deng, Liu, Zhang, Guo, Cui, Pei, Wu, Xie, Chen, Lu, Hu, Wu, Chan, Chen, Zhou, Yu, Liu, Guo, Qin, and Liu]{xia2024target-aware}
Yingce Xia, Kehan Wu, Pan Deng, Renhe Liu, Yuan Zhang, Han Guo, Yumeng Cui, Qizhi Pei, Lijun Wu, Shufang Xie, Si~Chen, Xi~Lu, Song Hu, Jinzhi Wu, Chi-Kin Chan, Shuo Chen, Liangliang Zhou, Nenghai Yu, Haiguang Liu, Jinjiang Guo, Tao Qin, and Tie-Yan Liu.
\newblock Target-aware molecule generation for drug design using a chemical language model.
\newblock January 2024.

\bibitem[Yang et~al.(2024)Yang, Yadav, Melville, Harley, Krause, and Tonks]{yang2024triple}
Lin Yang, Vishal Yadav, Joseph Melville, Joel~B Harley, Amanda~R Krause, and Michael~R Tonks.
\newblock A triple junction energy study using an inclination-dependent anisotropic monte carlo potts grain growth model.
\newblock \emph{Materials \& Design}, 239:\penalty0 112763, 2024.

\bibitem[Ye et~al.(2025)Ye, Huang, Chen, Liu, Wang, and Dong]{ye2025pdeformerfoundationmodelonedimensional}
Zhanhong Ye, Xiang Huang, Leheng Chen, Hongsheng Liu, Zidong Wang, and Bin Dong.
\newblock Pdeformer: Towards a foundation model for one-dimensional partial differential equations, 2025.
\newblock URL \url{https://arxiv.org/abs/2402.12652}.

\bibitem[Yenduri et~al.(2023)Yenduri, M, G, Y, Srivastava, Maddikunta, G, Jhaveri, B, Wang, Vasilakos, and Gadekallu]{yenduri2023generativepretrainedtransformercomprehensive}
Gokul Yenduri, Ramalingam M, Chemmalar~Selvi G, Supriya Y, Gautam Srivastava, Praveen Kumar~Reddy Maddikunta, Deepti~Raj G, Rutvij~H Jhaveri, Prabadevi B, Weizheng Wang, Athanasios~V. Vasilakos, and Thippa~Reddy Gadekallu.
\newblock Generative pre-trained transformer: A comprehensive review on enabling technologies, potential applications, emerging challenges, and future directions, 2023.
\newblock URL \url{https://arxiv.org/abs/2305.10435}.

\bibitem[Zelaia et~al.(2023)Zelaia, Cheng, Mayeur, Ziabari, and Kirka]{paxti_polycrystal_diffusion}
P.F. Zelaia, J.~Cheng, J.R. Mayeur, A.K. Ziabari, and M.M. Kirka.
\newblock Digital polycrystalline microstructure generation using diffusion probabilistic models.
\newblock \emph{SSRN}, 2023.
\newblock \doi{10.2139/ssrn.4419461}.

\bibitem[Zeni et~al.(2025)Zeni, Pinsler, Z{\"u}gner, Fowler, Horton, Fu, Wang, Shysheya, Crabb{\'e}, Ueda, et~al.]{zeni2025generative}
Claudio Zeni, Robert Pinsler, Daniel Z{\"u}gner, Andrew Fowler, Matthew Horton, Xiang Fu, Zilong Wang, Aliaksandra Shysheya, Jonathan Crabb{\'e}, Shoko Ueda, et~al.
\newblock A generative model for inorganic materials design.
\newblock \emph{Nature}, pages 1--3, 2025.

\bibitem[Zhang et~al.(2023)Zhang, Rao, and Agrawala]{zhang2023adding}
Lvmin Zhang, Anyi Rao, and Maneesh Agrawala.
\newblock Adding conditional control to text-to-image diffusion models.
\newblock In \emph{Proceedings of the IEEE/CVF international conference on computer vision}, pages 3836--3847, 2023.

\end{thebibliography}

\newpage
\appendix
\restoreTOC
\renewcommand{\contentsname}{Appendices}
\tableofcontents

\newpage

\section{Background}
\subsection{Previous Efforts in Big Dataset Design for Materials Science}

Materials datasets -- mirroring many other scientific fields that study spatial fields -- have historically been disparate and specialized \cite{holm_powderdatabase, holm_decost_classification, neu_steel_defect_database, titanium_database, doitpoms_micrograph_database, carbonsteel_database, sofc_experimental_database}. Although such datasets are useful for targeted tasks, they lack the diversity and scale to support generalist foundation models. 

Recently, several pioneering efforts demonstrate potential strategies for generalizing. The Polymathic AI Collaboration's `The Well' is a 15 terabyte dataset comprised of numerical simulations from 16 disparate scientific domains \cite{ohana2024well}. The unifying characteristic is that data in The Well (i.e., the specific instantiations of various field variables) is derived by solving well understood partial differential equations under varying parameterizations and boundary conditions. We call this strategy "process-driven" -- since the governing PDE, which defines a process, is used to explore and sample the data space -- dataset design. It is also popular in materials science; e.g., TAMU's spinodal decomposition dataset \cite{attari2020uncertainty}. Similarly, most experimental datasets can be considered "process-driven" (e.g., \cite{sofc_experimental_database}). This big dataset generation strategy is tremendously valuable because it (1) provides time series data representing important scientific systems and (2) contains data which is strictly adherent to known governing physics. However, it also has its limitations. There are many scientific systems in which mathematical process modeling remains either (1) underdeveloped or (2) too expensive to run as a big data source. Many materials science applications -- particularly polycrystalline materials applications -- fall into this category. For example, the microstructure evolution process in metal processing (e.g., in additive manufacturing) remains poorly understood and difficult to model (category 1) \cite{yang2024triple, rodgers2017monte}. Even when the process is well modeled, crystalline process models are extremely computational expensive. For example, fatigued microstructures require tens of thousands of compute hours \textit{per simulation} to perform (category 2) \cite{ghosh2021watmus}. Furthermore, individual processes have limited exploration capacity. Many real materials used every day are only realizable due to the chaining of several processes. This type of chaining adds significant cost to an already expensive data generation strategy.

`Statistics-driven' dataset design is an alternative paradigm to process-driven methods. Here, the space of interest (e.g., the microstructure space) is explored directly and elements are added to a dataset based on a salient diversity measure (e.g., microstructure statistics) instead of by directive of the process. Robertson \textit{et al.}'s MICRO2D dataset demonstrates this paradigm to curate statistically diverse 2-phase microstructures \cite{robertson2024micro2d}. Although such microstructures could arise as a result of a process, the precise process can be unknown or not well understood. Instead, MICRO2D is designed by curating (i.e., generating) microstructures displaying diverse auto-correlations. In this sense, it is guided by concepts in statistical physics instead of specific governing equations. As demonstrated in the MICRO2D paper, this strategy is able to generate far greater statistical diversity than process-driven methods, but lacks the temporal coupling that is present in `process-driven' datasets. 

Briefly, although these two paradigms have been separated in practice, we emphasize that the two strategies are not necessarily separate. For example, `statistics-driven' methods can be used to define more complex and realistic initial conditions to seed further `process-driven' diversification. Notably, these two paradigms -- and their combination -- are not just restricted to scientific applications, they are also observable in traditional machine learning fields. For example, distillation and other methods using text generated by Large Language Models are process-driven \cite{abdin2024phi3}. In contrast, scene augmentation for autonomous driving is statistics-driven (i.e., specific concepts or settings which are deemed important and underrepresented are systematically added to datasets). Finally, concept seeding methods (e.g., \cite{abdin2024phi4, riaz2025metasynth}) initialize synthetic LLM data with heuristically selected important concepts -- a combination of the two approaches.

In PolyMicros, we expand significantly upon MICRO2D's `statistics-driven' dataset design paradigm, overcoming limitations of the MICRO2D approach (e.g., MICRO2D is limited to single variable fields). Although this technique is developed for microstructures, we emphasize that it is not conceptually restricted to this domain. The only requirement of the proposed augmentation algorithms is that the targeted system is statistically stationary. This is a loose requirement that is commonly adopted in many scientific applications -- as exemplified by the ubiquitous use of Gaussian Random Fields as seeding algorithms (e.g., \cite{qiu2024derivative, lim2023score, kovachki2023neural}). 

\subsection{Reduced Order Generalized Spherical Harmonics}
\label{app:rogsh_appendix}

Many strategies exist for representing orientation fields pointwise \cite{bunge, jangid2024q}. Basic representations such as Euler angles ($\bm{g}=\{\varphi_1, \Psi, \varphi_2\}\in\mathbb{R}^3$) provide compact representation but do not account for material symmetries \cite{bunge}. Consequently, learning problems involving these representations require the network to account for these symmetries internally, leading to learning schemes that are extremely data and parameter hungry \cite{buzzy2024statistically, jangid2024q}. Generalized Spherical Harmonics (GSH) are a powerful alternative; they are a function basis representation scheme specialized for functions defined on SO(3) (i.e., the space of orientations) \cite{bunge}. A carefully selected subset of the GSH basis functions will naturally represent functions displaying any desired material symmetry (i.e., different subsets are utilized for cubic, hexagonal, etc.) 
\begin{equation}
    h(\bm{g}) = \sum_{l=0}^{\infty} \sum_{\mu=1}^{M(l)}\sum_{\nu=1}^{N(l)} C_l^{\mu \nu} \dot{\ddot{T}}_{l}^{\mu \nu}(\bm{g}).
\end{equation}
\noindent Here, $\dot{\ddot{T}}^{\mu \nu}_{l}(g)$ are the symmetrized generalized spherical harmonic functions. We refer the interested reader to the following reference \cite{bunge} for an exhaustive list of definitions for these functions for common material symmetries. $\bm{g}$ is defined as the set of Euler angles. Here, we follow Bunge's convention \cite{bunge} and utilize intrinsic ZXZ Euler angles
\begin{equation}
    \bm{g} = \{ \varphi_1, \Phi, \varphi_2 \}.
\end{equation}
To represent a crystal orientation using the GSH method, we define the orientation distribution function (odf), $f(\bm{g})$. The odf is a probability density, defining the probability of a specific differential orientation: $f(\bm{g})d\bm{g}$. To represent a full polycrystalline material, the odf is defined pointwise in space: $f(\bm{x}, \bm{g})d\bm{g}$. In most cases, we assume that a polycrystal is defined by a single orientation at each spatial location (or, if discretized, within each voxel). In this setting, the odf reduces to a dirac measure, $f(\bm{x}, \bm{g}) = \delta(\bm{g} - \hat{\bm{g}}(\bm{x}))$. The odf -- and, by extension, the polycrystal -- is represented discretely as the GSH basis coefficients, $C_l^{\mu \nu}$ by projection
\begin{equation}
    C_l^{\mu \nu}(\bm{x}) = (2l+1) \int_{SO(3)} f(\bm{x}, \bm{g}) \dot{\ddot{T}}^{\mu \nu}_l (\bm{g})^* d\bm{g}
\end{equation}
where $*$ is the conjugate operation. For a polycrystal represented using the dirac measure, this expression analytically reduces as follows. 

\begin{equation}
    C_{l}^{uv} = (2l+1)\dot{\ddot{T}}_{l}^{*uv}(\hat{\bm{g}}(\bm{x}))
    \label{eq:delta_coef_gsh}
\end{equation}

Downstream learning is performed using a finite set of the basis coefficients as representation. In addition to naturally accounting for symmetries, the GSH representation naturally accepts standard distance measures further simplifying the application of standard machine learning techniques. For these benefits, the limitation of GSH coefficients as a representation scheme is that it is an infinite series expansion. It is standard to take large basis sets \cite{polycrystal2ps}.

The Reduced Order Generalized Spherical Harmonics (ROGSH) are a further compacted subset of the symmetrized GSH series. ROGSH coefficients are utilized for representing orientation fields which only contain a single orientation at each spatial location. For these systems, it has been shown that it is sufficient to retain only 3 carefully selected GSH coefficients \cite{buzzy2024statistically}. Thus, the ROGSH account for symmetry and distance while also providing a compact representation amenable for learning. The ROGSH functions for cubic symmetric systems are recreated below \cite{buzzy2024statistically}
\begin{multline}
Re(\dot{\ddot{T}}_{4}^{-4,1}) = \frac{\sqrt{30}}{192} \left(14 \left(\cos{\left(\Phi \right)} - 1\right)^{2} \cos{\left(4.0 \varphi_{1} \right)} + \left(\cos{\left(\Phi \right)} + 1\right)^{2} \cos{\left(4.0 \varphi_{1} + 4.0 \varphi_{2} \right)}\right) \\
\left(\cos{\left(\Phi \right)} + 1\right)^{2} + \left(\cos{\left(\Phi \right)} - 1\right)^{4} \cos{\left(4.0 \varphi_{1} - 4.0 \varphi_{2} \right)},
\end{multline}

\begin{equation}
\dot{\ddot{T}}_{4}^{0,1}=\frac{\sqrt{21}}{48} \left(5 \sin^{4}{(\Phi )} \cos{\left(4 \varphi_{2} \right)} + 35 \cos^{4}{\left(\Phi \right)} - 30 \cos^{2}{\left(\Phi \right)} + 3.0\right),
\end{equation}
and 
\begin{multline}
\dot{\ddot{T}}_{12}^{0,2} = \frac{\sqrt{166305594}}{40304640} \Bigl(1025 \left(\cos{\left(\Phi \right)} - 1\right)^{6} \left(\cos{\left(\Phi \right)} + 1\right)^{6} \cos{\left(12 \varphi_{2} \right)} \\ + 66 \left(\cos{\left(\Phi \right)} - 1\right)^{4} \left(\cos{\left(\Phi \right)} + 1\right)^{4} \cdot \bigl(161.0 \sin^{4}{\left(\Phi \right)} \\ - 280.0 \sin^{2}{\left(\Phi \right)} + 120.0\bigr) \cos{\left(8 \varphi_{2} \right)} + 99 \left(\cos{\left(\Phi \right)} - 1\right)^{2} \left(\cos{\left(\Phi \right)} + 1\right)^{2} \\ \left(7429 \cos^{8}{\left(\Phi \right)} - 9044 \cos^{6}{\left(\Phi \right)} + 3230 \cos^{4}{\left(\Phi \right)} - 340 \cos^{2}{\left(\Phi \right)} + 5\right) \cos{\left(4 \varphi_{2} \right)} \\ + 1352078 \cos^{12}{\left(\Phi \right)} - 3879876 \cos^{10}{\left(\Phi \right)} + 4157010 \cos^{8}{\left(\Phi \right)} - 2042040 \cos^{6}{\left(\Phi \right)} \\ + 450450 \cos^{4}{\left(\Phi \right)} - 36036 \cos^{2}{\left(\Phi \right)} + 462\Bigr).
\end{multline}

\subsection{Local-Global Decompositions}
\label{app:lgd_appendix}

In this section, we provide a more extended overview of the Local-Global Decomposition (LGD) framework for building statistically conditioned microstructure generative models. We refer the interested reader to \cite{andreas_LGD, buzzy2024statistically} for complete discussions. The LGD framework is a data efficient and stable mathematical framework for building conditional generative models for material microstructures. LGD models are conditioned on 1- and 2-point microstructure statistics, allowing the models to be used to systematically sample statistically diverse microstructures. Their major benefit is that, while they can be sampled conditioned on many different values of these statistics, they are trained on as a little as one example microstructure representing just one set of 1- and 2-point statistics. These dual characteristics makes them particularly well suited for the data augmentation task we pursued in this paper. 

The LGD framework utilizes the following conceptual decomposition of the statistically conditioned microstructure generating process to construct a tractable, numerically efficient, and stable generative model. To make the estimation of the statistical moments and sampling tractable, the 1- and 2-point statistics conditioned microstructure generating process (called a stochastic microstructure function) is assumed to be periodic, stationary, and ergodic, i.e., 
\begin{equation}
     p(\boldsymbol{m}_1, ..., \boldsymbol{m}_S; \boldsymbol{\mu}, \boldsymbol{f}_r) = \mathcal{N}(\hat{\boldsymbol{m}}_1, ..., \hat{\boldsymbol{m}}_S ; \boldsymbol{\mu}, \boldsymbol{f}_r) \prod_{i=1}^K p^{cond}\left(\bm{N}_i  | \hat{\bm{N}}_i, \bm{N}_i^c ; \Phi^{(3, ...)}\right).
     \label{eq:LGD}
\end{equation}

\noindent Here, $\bm{m}_s \in \mathbb{R}^H$ is the vector-valued microstructure state at the $s^{\mathrm{th}}$ voxel. $\bm{\mu} \in \mathbb{R}^H$ is the vector of mean values of each phase. $\bm{f}_r \in \mathbb{R}^{1 \times H}$ is the set of 2-point statistics (i.e., auto- and cross-correlations) at voxel separation $r$. $\hat{\bm{m}}_s$ is the $2^{\mathrm{nd}}$ order approximation of the microstructure at the $s^{\mathrm{th}}$ voxel. The global component, $\mathcal{N}(\cdot ; \bm{\mu}, \bm{f}_r)$, is a Multi-Output Gaussian Random Field evaluated at the discrete voxel locations. $K$ is the number of neighborhood locations. $p^{cond}(\cdot | \cdot, \cdot)$ is the conditional neighborhood distribution which is constructed \textit{post-training} from an unconditional diffusion model-based approximation of the neighborhood distribution. $\bm{N}_i$ is the value of the microstructures in the voxels contained in the $i^{\mathrm{th}}$ neighborhood. $\hat{\bm{N}}_i$ are the value of the $2^\mathrm{nd}$ order approximations of the microstructures in the $i^{\mathrm{th}}$ neighborhood voxels. And $\bm{N}_i^c$ is the value of the microstructures in the voxels not in the $i^{\mathrm{th}}$ neighborhood. 

\begin{figure}
\begin{center}
    
\includegraphics[width=.75\linewidth]{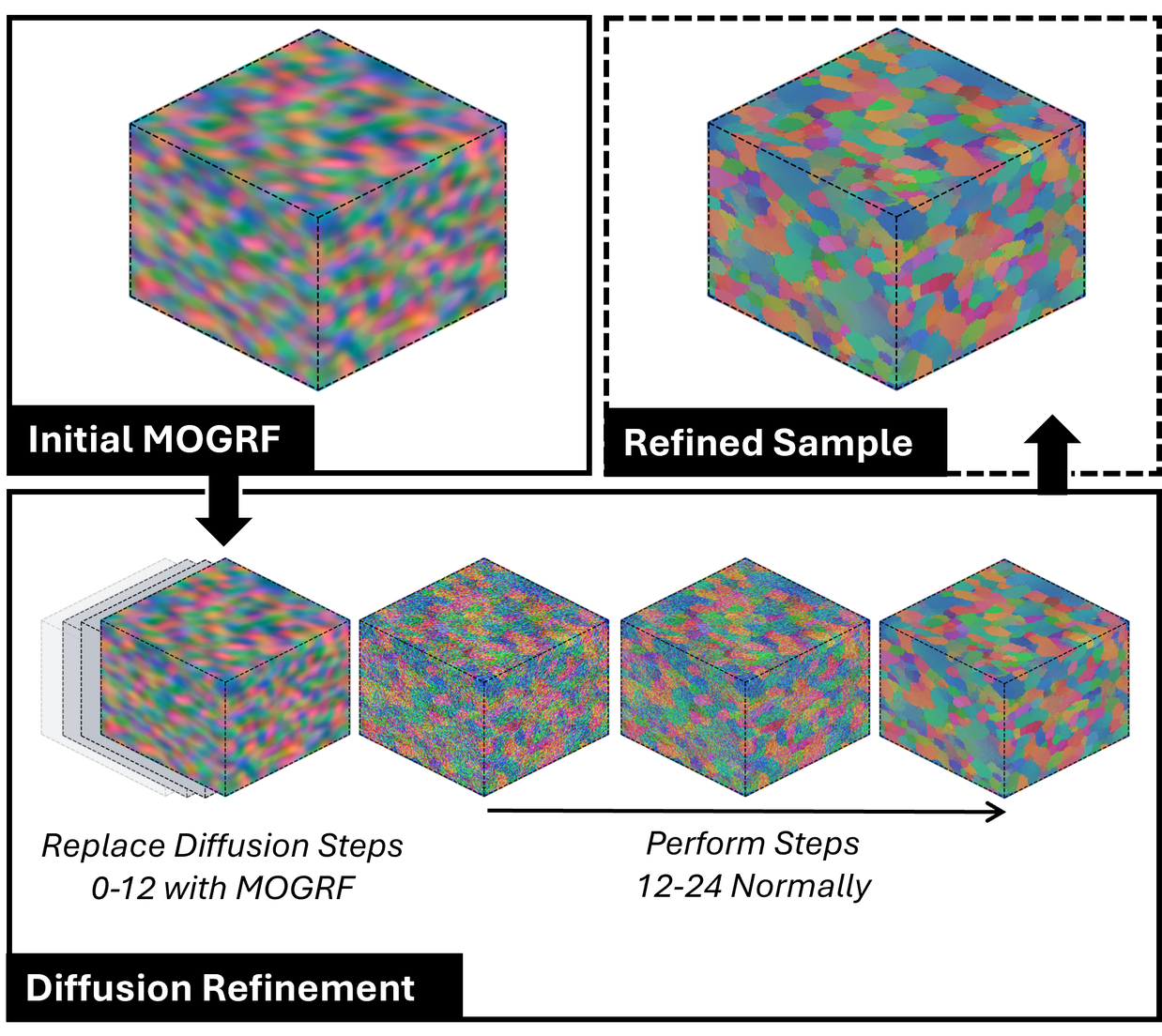}

\end{center}
\caption{Visualization of the LGD framework's two stage sampling process. A sample from the Multi-Output Gaussian Random Field (MOGRF) is used to seed a truncated diffusion process. The diffusion process locally refines microstructure neighborhoods to introduce local realism. The MOGRF imposes important statistically conditioned long range patterns.}
\label{fig:LocalRef:Denoise}
\end{figure}

Constructing an LGD model involves two components. The MOGRF, $\mathcal{N}(\cdot ; \bm{\mu}, \bm{f}_r)$, is constructed directly from a provided mean vector, $\bm{\mu}$, and 2-point statistics, $\bm{f}_r$, see App. \ref{app:mogrf_appendix}. The conditional model, $p^{cond}(\cdot | \cdot, \cdot)$, is adapted \textit{post-training} from an unconditional local neighborhood diffusion model. The unconditional local neighborhood diffusion model is trained first to generate potential patches (neighborhoods of voxels whose neighborhood size is on the order of the smallest salient feature in the microstructure: e.g., for polycrystals, a few grains). A large dataset of these patches is cropped from a single source image. Subsequently, an unconditional diffusion model is trained on the source patch dataset to approximate the neighborhood distribution, see App. \ref{app:training_neighborhood_models} for details on neighborhood training strategies used in this work. Subsequently, post-training, the backward sampling process is updated in two ways to transform the unconditional diffusion model into a conditional diffusion model. This adaptation process is utilized over a standard conditional training process because conditional paired data is unavailable. First, instead of running the entire diffusion process seeded on the original source noise distribution (i.e., scaled white noise), the diffusion process is seeded at an intermediate time from the sampled output of the MOGRF (using a technique similar to SDEdit \cite{sdedit}). The precise intermediate time is empirically selected to balance adherence with the MOGRF's spatial statistics and the realism of the generated samples (e.g., see App. C.3 in \cite{andreas_LGD}). For an EDM based sampling process \cite{karras2022elucidating} (for an overview, see App. \ref{app:ddm_appendix}), roughly half the sampling steps are used. Second, after each update step, the mean value of each field is manually corrected to adhere to the target 1-point statistics\footnote{This is a deviation from the original LGD framework \cite{andreas_LGD}, which utilizes a posterior form of the score function. This adaptation has demonstrated better performance for polycrystalline systems \cite{buzzy2024statistically}.}. Altogether, sampling a microstructure is achieved by first identifying a target 1-point statistics, 2-point statistics, and neighborhood distribution (represented by the unconditional diffusion model). The constructed MOGRF is sampled and used to seed the diffusion process. The adapted diffusion process is subsequently performed, Fig. \ref{fig:LocalRef:Denoise}. For polycrystalline systems, a final step of inverting the generated ROGSH coefficients back into Euler angles is optionally performed if Euler angles are the desired representation \cite{buzzy2024statistically}. Notably, the input statistics (i.e., the 1- and 2-point statistics) can be adjusted to generate microstructures displaying dissimilar spatial statistics from the original training source image. For example, for simple 2-phase microstructures, Robertson \textit{et al.} constructed a dataset of varying autocorrelations and used this property to expand a small set of source images into a large 2-phase microstructure dataset, MICRO2D. A similar concept motivates this work. However, this extrapolation ability has its limits; empirically, 2-phase systems appear to display better extrapolation than polycrystalline systems using traditional LGD sampling. We hypothesize that this is due to weaker coupling between the 1- and 2-point statistics and the neighborhood distribution in N-phase systems. This empirical stability defines the maximum deviation from the source image's spatial statistics that can still stably seed the diffusion process. In App. \ref{app:lgd_sampling_adjustments}, we introduce new small adjustments to the LGD sampling process for polycrystalline microstructures that we found to empirically improve extrapolation stability during the course of this work. In that section, we also include the LGD algorithm, Alg. \ref{alg:lgd_datagen}. In App. \ref{app:limitations}, we identify potential methods for improving extrapolation stability whose exploration we leave to future work.

\subsection{Terminology Clarification: N-point Statistics Versus Covariances}

\label{app:covVstats}

We would like to briefly clarify the language barrier between the materials and computational communities. N-point statistics (often used interchangeably with spatial correlations) are terms used to indicate the set of auto- and cross- correlations computed from spatially resolved voxelized material observations. Importantly these statistics contain the mean. In contrast the statistics community will often discuss the zero mean covariance structure of the random field instead. These two representations are trivially related to one another via:

\begin{equation}
     \Sigma^{\alpha \beta} = f^{\alpha \beta} - \mu^{\alpha} \mu^{\beta}.
\end{equation}

In materials, it is standard to assume statistical stationarity. Therefore, the means are spatially constant while the covariance/correlation between the field at two locations depends only on the vector separation between the locations. Throughout this work we have been very intentionally in using the term n-point statistic when the mean is added, and the term covariance when the mean is not added. However, the above equation makes it clear that the correlation structure of these two entities is identical, and generating one allows the other to trivially be computed. Kernels generate covariances not the n-point statistic typically computed from experimental micrographs, however we will often think of a kernel as generating n-point statistics moving forward and vice versa since the proposal of a mean is trivial.

\subsection{Multi-Output Gaussian Random Field}
\label{app:mogrf_appendix}

The Multi-Output Gaussian Random Field model is a second order approximation of the joint distribution between every dimension of the vector-valued microstructure in every voxel describing the spatial domain. Specifically, the joint distribution is approximated as a multivariate Gaussian over vectors in $\mathbb{R}^{S\times H}$ (for a microstructure with $H$ states and $S$ voxels). Notably, this joint distribution is never constructed or sampled directly using standard sampling methods due to extreme memory and computational costs. For example, the MOGRF covariance is size $(SH)^2$ elements: that would be over $40$ trillion elements for the application described in this paper. Instead, it is observed that because the system is assumed to be periodic and statistically stationary the covariance matrix is near block circulant (precisely, it is composed of blocks -- indexed by the state $h\in \{0, ..., H-1\}$ -- which are themselves block circulant). This property is leveraged to efficiently sample in $\mathcal{O}(HS\ln S)$ computational cost and $\mathcal{O}(HS)$ memory cost. The algorithm for sampling the MOGRF is recreated from \cite{andreas_NGRF} in Alg. \ref{alg:mogrf}.

\begin{algorithm}
\caption{For efficiently drawing samples from a stationary, periodic Multi-Output Gaussian Random Field. Algorithm recreated from \cite{andreas_NGRF}.}\label{alg:mogrf}
\begin{algorithmic}[1]

\Require Target 2-point statistics (autocorrelation of reference state and cross-correlations with remaining states). Reference state should be selected to have non-constant value: $\bm{f} \in \mathbb{R}^{S\times 1 \times H}$. Target 1-point statistics: $\bm{\mu} \in \mathbb{R}^H$.

\State $H :=$ number of microstructure states at a single spatial location. E.g., for ROGSH, $H=3$.
\State $S :=$ number of voxels in discrete grid.
\State $\mathcal{F}[\cdot] := \mathrm{Discrete \ Fourier \ Transform \ using \ FFT}$
\State $\mathcal{F}[\cdot]^{-1} := \mathrm{Inverse \ DFT}$
\State $G[\cdot]:=$ a function returning either the imaginary or real component of a complex number.
\State $\delta_{ab}:=$ the Kronecker delta
\Procedure{MOGRFSampler}{$\bm{\mu}$, $\bm{f}$, $\epsilon=1\times 10^{-12}$}
\State $\epsilon_1, \epsilon_2 \sim \mathcal{N}(0, 1)$
\State $x^0_s=\mu_0 + \mathcal{G} \left[ \mathcal{F} \left[ \left( \frac{1}{S} (1 - \delta_{0t}) \mathcal{F} \left[ f_r^{00} \right]_t \right)^{1/2} (\epsilon_1+i \epsilon_2)_t \right]_s \right]$
\State $F_t^{00} = \mathcal{F}[f^{00}_r]_t$
\State $X_t^{0} = \mathcal{F}[x^{0}_s]_t$
\For{$\gamma \in \{ 1, ..., H-1\}$}
    \State $F_t^{0\gamma} = \mathcal{F}[f^{0\gamma}_r]_t$
    \State $x^{\gamma}=\mathcal{F}^{-1}\left[ \frac{F^{0\gamma}_t}{F^{00}_t + \epsilon} X^0_t \right]$
\EndFor

\State \textbf{return} $\bm{x}=\{ \bm{x}^0, ..., \bm{x}^{H-1} \} \in \mathbb{R}^{S \times H}$
\EndProcedure

\end{algorithmic}
\end{algorithm}


\subsection{Multi-Output Spectral Mixture Kernel}
\label{app:mosmbackground}

The Multi-Output Spectral Mixture Kernel is a recently developed kernel for Gaussian Process Regression modeling capable of modeling complex correlation structures in multi-task learning environments \cite{parra2017spectralmixturekernelsmultioutput}. In this work we identify this kernel as having several key attributes useful for modeling the spatial correlation structures observed in polycrystalline materials. Inspecting the kernel function in Eq. \ref{eq:MOSM} we can understand the necessary properties. Firstly, the kernel is capable of producing valid sets of auto- and cross- correlations (i.e., correlations which are compatible with one another) through the spectral mixture of Gaussians. This characteristic is critical as the multi-channel nature of polycrystalline materials (practically consisting of a three channel image) necessitates the specification of both kinds of statistics for subsequent GRF sampling steps. While the MOSM kernel is not unique in this capacity it additionally meets a second requirement, namely that it can model auto- and cross- correlations which are out of phase with one another. Practically, when observing polycrystalline spatial statistics, we find that very rarely do the cross-correlations posses centered peaks co-located with the auto-correlation's center peak. Instead the cross-correlations have their peaks offset from the center. The MOSM kernel readily models this behavior as a phase shift via the added cosine term (which, given the constraints shown, below only applies to the cross correlations). Other methods for generating cross- correlations do not offer this ability to offset their peaks. The Linear Method of Coregionalization \cite{bonilla2007linearcoregionalization}, for example, produces valid cross correlations through linear combinations of the auto correlations, thus necessitating that the cross-correlations share the same centered structure as the auto-correlations. Finally, the MOSM kernel is dense in the space of all kernels given that the number of mixture elements is infinite \cite{parra2017spectralmixturekernelsmultioutput, wilson_2013_GaussianSpectralMixture}. This last feature likely indicates that the spatial correlations produced by the MOSM kernel are a superset of the space of correlations realizable by experimentally obtained polycrystalline materials. This is important from a data generation standpoint as it means we theoretically have the potential to actualize any set of polycrystalline 2-point statistics synthetically without needing to utilize costly experiments for direct observations.

The MOSM kernel as shown in the main body of the paper posses many parameters. It is important that such parameters are properly specified such that the kernel produces valid auto- and cross- correlations. The following equations (copied from \cite{parra2017spectralmixturekernelsmultioutput}) decomposed the parameters into sub-parameters such that the generated correlations are valid: 

\begin{equation}
    {A}_{\beta \gamma} = 2 {A}_\beta  \left( {A}_\beta  + {A}_\gamma \right)^{-1} {A}_\gamma, 
\end{equation}

\begin{equation}
    \varsigma_{\beta \gamma} = \left( {A}_\beta  + {A}_\gamma \right)^{-1} \left( {A}_\beta  \varsigma_\beta  + {A}_\gamma \varsigma_\gamma \right),
\end{equation}

\begin{equation}
    w_{\beta \gamma} = w_\beta  w_\gamma \exp{\left(  - \frac{1}{4} (\varsigma_\beta  - \varsigma_\gamma)^\intercal ({A}_\beta  + {A}_\gamma)^{-1} (\varsigma_\beta  - \varsigma_\gamma)\right)},
\end{equation}

\begin{equation}
    \theta_{\beta \gamma} = \theta_\beta  - \theta_\gamma,
\end{equation}

\begin{equation}
    \phi_{\beta \gamma} = \phi_\beta  - \phi_\gamma,
\end{equation}

\begin{equation}
\alpha_{\beta \gamma}^{(q)} = w_{\beta \gamma}^{(q)} (2 \pi)^{\frac{n}{2}}|{A}_{\beta \gamma}^{(q)}|^{1/2},
\end{equation}
where $n=3$ for a 3D dimensional vector $r$.

\subsection{Denoising Diffusion Models}
\label{app:ddm_appendix}

In this work we utilize the EDM formulation of diffusion from \cite{karras2022elucidating}, due to its enhanced training stability and sampling efficiency. EDM aims to learn a network $F_{\theta}(\cdot)$ which acts as a denoiser of the form:

\begin{equation}
    D_{\theta}(\bm{x};\sigma) = c_{skip}(\sigma) + c_{out}(\sigma)F_{\theta}(c_{in}(\sigma)\bm{x};c_{noise}(\sigma)).
\end{equation}

\noindent Here, $c_{skip}(\sigma)$, $c_{out}(\sigma)$, and $c_{noise}(\sigma)$ are modulation constants constructed to make the inputs and outputs of $F_{\theta}(\cdot)$ unit variance. $\sigma$ is the variance of the noise schedule used in diffusion modeling. During training the denoiser aims to minimize the expected $L_2$ denoising error for samples drawn from $p_{data}$ for every noise level $\sigma$:

\begin{equation}
    \mathbb{E}_{\bm{y} \sim p_{data}} \mathbb{E}_{n \sim \mathcal{N}(\bm{0},\sigma^2 \bm{\text{I}})} ||D(y+n;\sigma) - y||^2_2.
\end{equation}

\noindent The denoiser then samples from a data distribution through its relationship to the score via integration of the probability flow ODE:

\begin{equation}
    d\bm{x} = \dot{\sigma}(t)\sigma(t)\nabla_{\bm{x}} \log p(\bm{x};\sigma(t)) dt, 
\end{equation}
where 
\begin{equation}
    \nabla_{\bm{x}} \log p(\bm{x};\sigma) = (D(\bm{x};\sigma) - x)/  \sigma^2.
\end{equation}
Importantly the EDM formulation identifies key values for various hyper parameters needed both during training and sampling which we adopt without modification \cite{karras2022elucidating}. Additionally they propose a stochastic sampling procedure which combines a second order deterministic ODE integrator with the addition of a stochastic churn step. We adopt this sampler with a few minor modifications, namely, the ability to start the diffusion process from an arbitrary timestep, and the ability to call an arbitrary condition function after each diffusion step. The sampling code is reproduced in Alg. \ref{alg:stochastic_sampler} with our changes to make clear the specifics of how and, more importantly, in what order these modifications occur. 

\begin{algorithm}
\caption{Modified Second-Order Stochastic Sampler}\label{alg:stochastic_sampler}
\begin{algorithmic}[1]

\Require Noise Schedule: $t_{i\in\{0,...,N\}}$, Parameters $S_{noise}, S_{churn}, S_{tmin}, S_{tmax}$

\Procedure{ModifiedSampler}{$\bm{x}, D_{\theta}(\bm{x};\sigma), skip, F_{cond}(\bm{x},i)$}

\If{$skip = 0$ }
\State \textbf{sample} $\bm{x} \sim \mathcal{N}(0,t^2_{0} \bm{I})$
\EndIf \\

\For{$i \in \{skip, ..., N-1\}$}

\State $\bm{\epsilon}_i \sim \mathcal{N}(0,S^2_{noise})$

\State $\hat{\gamma_i} \leftarrow \texttt{min}(\frac{S_{churn}}{N}, \sqrt{2}-1) \text{ if } t_i \in [S_{tmin}, S_{tmax}] \text{, } 0 \text{ otherwise }$

\State $\hat{t}_i \leftarrow t_i + \hat{\gamma_i }t_i$ 

\State $\bm{\hat{x}}_i \leftarrow \bm{x} + \sqrt{\hat{t}^2_i - t^2_i} \bm{\epsilon}_i$

\State $\bm{d}_i \leftarrow (\bm{\hat{x}}_i - D_{\theta}(\bm{\hat{x}}_i;\hat{t}_i))/\hat{t}_i$ 

\State $\bm{x}_{i+1} \leftarrow \bm{\hat{x}}_i + (t_{i+1} - \hat{t}_i)\bm{d}_i$\\

\If{$t_{i+1} \neq 0$ }
\State $\bm{d^{\prime}}_i \leftarrow (\bm{x}_{i+1} -  D_{\theta}(\bm{x}_{i+1};t_{i+1}))/t_{i+1}$

\State $\bm{x}_{i+1} \leftarrow \bm{\hat{x}}_i + (t_{i+1}-\hat{t}_i)(\frac{1}{2}\bm{d}_i + \frac{1}{2}\bm{d^\prime}_i)$

\EndIf \\

\If{$f_{cond} \neq \varnothing$ }
\State $\bm{x}_{i+1} = F_{cond}(\bm{x}_{i+1},i)$
\EndIf\\

\EndFor

\State \textbf{return} $\bm{x}_N$

\EndProcedure

\end{algorithmic}
\end{algorithm}

\FloatBarrier

\section{Data Generation}

\subsection{Multi-Output Spectral Mixture Kernel}
\label{app:MOSM}

Having identified the MOSM kernel as a suitable parametrization of second order polycrystalline spatial statistics in App. \ref{app:mosmbackground}, we now turn our attention to the practical usage of this kernel for proposing polycrystalline spatial covariances \footnote{Please see App. \ref{app:covVstats} for the distinction between 2-Point statistics and covariances}. To reiterate, our goal here is to synthetically produce realistic long range statistics to approximate a diverse set of global polycrystals, which will later be locally refined using the LGD approach. This framing leads to several guidelines that we will utilize during the global generation procedure. 1) We wish to attempt to generate "valid" polycrystals. Although the space of polycrystals is not well understood, there are clear violations (i.e non-valid ROGSH coefficients) which would indicate we are producing non-physical covariances. 2) We wish to generate "true" polycrystals. Much of the space of polycrystalline materials is dominated by geometries which posses very few grains, i.e. mostly single crystals. Our interests are not in these systems as they do not represent most bulk engineering materials of interest. We therefore intend to generate volumes with very high numbers of grains (>500). 3) We are interested in generating low frequency spatial information, the high frequency information will be induced later in the local refinement step. 4) The covariances must be able to be sampled efficiently by the MOGRF. Ultimately this means they must correspond to spatially periodic signals.

Establishing these guidelines is critical as the MOSM kernel can easily produce statistics which would result in a spatial field that, when sampled, would not be considered a polycrystalline material \footnote{At this point, no samples are valid polycrystals due to invalid neighborhoods. Here we mean that the field has no potential to be a valid polycrystal even after local-refinement. In other words, that the local refinement will distort the long-range statistics so dramatically in order to construct local polycrystalline structure that the effect of the long range statistics will be entirely eliminated.}. In practice we aim to convert these guidelines into bounds on the parameters which govern the MOSM Kernel, such that parameter sets within these bounds lead to meaningful covariance structures. Our intent is then to produce a diverse sampling of of covariances by sampling within these bounds. In practice, we accomplish this by defining simple bounds for each parameter (i.e. a minimum and a maximum) and performing latin hyper cube sampling (LHS) within the bounding box. The choice of bounds was a complex task involving both domain expertise as well as an aspect of trial and error. Here we list our rationale for bounds on each class of MOSM Kernel parameter. All parameters are for a domain of length $- \pi$ to $\pi$ in all three dimensions with 128 voxels per side. 

\begin{figure}[!h]
\centering
\includegraphics[width=.9\linewidth]{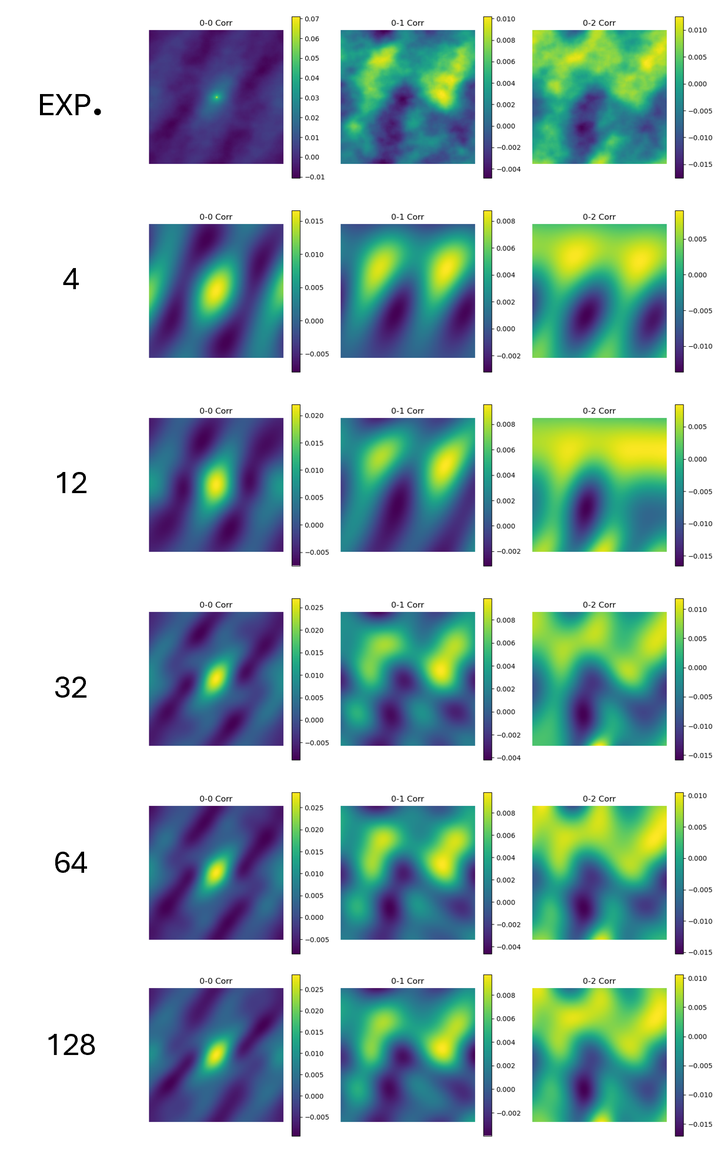}
\caption{Comparisons of an experimental microstructure's ROGSH auto- and cross-correlations (Row 1) against optimized MOSM kernel approximations using increasing numbers of mixtures. The MOSM kernel is able to faithly reconstruct the experimental microstructure statistics' complex features even with relatively few mixtures.}
\label{fig:MOSM:HMM}
\end{figure}

\begin{enumerate}
    \item \textbf{Gaussian mixture covariances} $A$: These covariances control the anisotropic decay of the spatial statistics as the spatial vector $r$ moves away from the mean $\varsigma$. A simple way to induce periodicity is to enforce that these statistics decay before the edge of the domain, such that the boundary is a homogeneous constant value. Since we want the structures to be generally low frequency at this stage we also limit the decay such that it isn't to quick, leading to a small peak. Our final choice was to limit the magnitude of the elements of this matrix to be between the values of 1.5 to 5. 

    \item \textbf{Gaussian mixture means} $\varsigma$: The mean term influences the frequency content of the correlations, here we choose low absolute values to search only the low frequency space. We limit the means to be between -5 and 5.

    \item \textbf{Weights} $\omega$: The weights control the magnitude of the statistics. Our goal is to introduce correlations such that the sampled fields do not have values outside of -1 to 1 as the un-symmetrized ROGSH cannot exceed these values. For a symmetrized system the ROGSH values will be smaller in magnitude but the symmetrized manifold is complex and only partially understood, so we enforce the boundaries loosely here. We found values of $-0.02$ to $0.02$ work well in the cubic case.

    \item \textbf{Delays} $\theta$: The delays result in a spatial offset in the cross correlations, here we limit them to $-0.5$ to $0.5$ such that the delay can be significant (with respect to a domain of -1 to 1), but will not push correlations beyond the edge of the domain and break periodicity

    \item \textbf{Phase} $\phi$: The phase shift is within a cosine term and is naturally bounded between 0 and $2 \pi$.

    \item \textbf{Number of mixtures} $Q$: The number of mixtures determines the overall complexity of the generated correlation structure. We briefly investigate the effect of this complexity qualitatively by fitting a MOSM kernel to a highly-complex experimentally obtained covariance structure. The results of this are shown in Fig. \ref{fig:MOSM:HMM}. We can see that as the number of mixture elements increases the kernel better approximates the covariance structure of the original experiment. However, with an increase in mixture elements comes a commensurate increase in the number of parameters. Since we are only interested in long range low frequency patterns we chose to keep the number of mixtures small to reduce the number of parameters. We ultimately chose 4 mixture elements which resulted in 132 kernel parameters. 
    
\end{enumerate}

We would like to note that this choice of min-max bounding and LHS sampling, while shown to be effective in this work, implicitly introduces additional limiting assumptions which may be sub-optimal, and that future work may be able to improve upon. Primarily, we are assuming the space of valid polycrystal covariances forms a box in the parameter space of the MOSM kernel. While with respect to some parameters this is likely or absolutely true, it is probably not the case for all, and a more intelligent classification of the parameters may lead to greater long range diversity (e.g., not a simple box, not convex, potentially discontinuous). In order to combat this we have chosen to be loose with the parameter bounds, meaning we have been willing to set the bounds to encompass a larger bounding box than than strictly necessary. We hope that this will allow for more diversity to be obtained by not rejecting a valid set of parameters by being too conservative. If a covariance is unreasonable, we can perform a variety of checks after the MOGRF sampling on the field itself to see if our guidelines were violated. Given the bounds above, we rejected about $8\%$ of proposed parameter sets since they clearly exited the space of feasible polycrystals (either non-periodic or non-valid ROGSH). We continuously produced more LHS samples until the total number of accepted statistics hit the target of 2000.

\label{app:num_mixture_elements}

\FloatBarrier

\subsection{Local-Global Decomposition Sampling}
\label{app:lgd_sampling_adjustments}

The aim of the LGD approach is to allow for the generation of a statistically diverse set of realistic polycrystalline volumes by modulating the underlying first- and second- order spatial statistics of the random field, while separately enforcing the higher-order local statistics necessary to achieve realism. With the target correlation structures established in the previous section, we now turn our attention to the practical considerations necessary to successful generate a synthetic datasets. The first consideration is with respect to the spatial mean (here we refer to the mean of the field not the mean of the gaussian mixture elements) of the MOGRF, since the MOSM kernel produces covariances, not n-point statistics, they therefore imply a zero mean. Inducing diversity in the mean is trivial in the ROGSH space as it only requires an addition of a constant to the field. Importantly this can be done at any step, either in the MOGRF itself, after MOGRF sampling on the global accurate field, or after the local refinement on the locally accurate field. We would like to note that the ability to induce mean diversity through simple shifts is not trivial in other orientation representations \cite{bunge, buzzy2024statistically}, and is a feature of our choice of the ROGSH. However, because mean diversity can be obtained so easily in this case we choose to perform the entire data generation process assuming zero mean. Additionally, we zero mean the experimentally obtained microstructures before breaking them into patches as well. In this way the entire generation procedure aims only to induce spatial diversity as mean diversity can be obtained at any time. 

\begin{algorithm}
\caption{Synthetic Data Generation}\label{alg:lgd_datagen}
\begin{algorithmic}[1]

\Procedure{DataGeneration}{$B_{global \in 
\{0,...,G\}}, D_{local \in \{0,...,L\}}, R, skip$}

\State $\mathcal{D} = \emptyset$
\For{$local \in \{0, ..., L\}$}
\For{$global \in \{0, ..., G\}$}
\For{$i \in \{0, ..., R\}$}

    \State $\Sigma^{\beta \gamma}(r) \leftarrow \texttt{MOSMKernel}(B_{global})$
    \State $\bm{x} \leftarrow \texttt{MOGRFSampler}(0,\Sigma^{\beta \gamma}(r))$
    \State $\bm{x} \leftarrow \texttt{ModifiedSampler}(\bm{x}, D_{local}, skip)$
    \State $\mathcal{D} \leftarrow \mathcal{D} \cup \{\bm{x}\}$

\EndFor
\EndFor
\EndFor

\State \textbf{return} $\mathcal{D}$
\EndProcedure

\end{algorithmic}
\end{algorithm}

The full data generation procedure is outlined in Alg. \ref{alg:lgd_datagen}. Here we seed the generation process with 1) $G = 2000$ global statistics $B_{global}$, 2) $L=5$ local diffusion models $D_{local}$, 3) as well as two hyperparameters, the oversampling parameter $R$ and the $skip$ parameter. The output is then the synthetically generated dataset $\mathcal{D}$. We chose $R=3$ such that three samples of the MOGRF were drawn for each set of kernel parameters, and found that a value of $skip = 12$ sufficiently refined the structures to be locally accurate without significantly perturbing the initial MOGRF field (see Fig. \ref{fig:LocalRef:Denoise}). The generation procedure was performed using a single Nvidia A100 80GB GPU. The overall generation time per structure was approximately 2 minutes with the local diffusion refinement incurring nearly the entire computational cost (the MOGRF generates a field in approx. 1.5 sec). Due to the large memory footprint of the 3D volumes, we were unable to get significant speedups from batching the local refinement steps. Overall the generation procedure incurred a substantial time cost, taking roughly 45 days to generate the full 30,000 synthetic microstructures. This could be significantly reduced in future efforts with more higher memory GPUs generating multiple batches of structures in parallel. Additionally, model distillation approaches such as consistency distillation \cite{song2023consistency} could reduce the number of model calls needed to locally refine the structure.

\FloatBarrier
\section{Model Architecture \& Training}

\label{app:training_neighborhood_models}

Throughout this work we train 6 diffusion models in total -- 5 to approximate local distributions, and 1 to serve as the PolyMicros Foundation Model. Overall the training scheme for these 6 models are all fairly similar to one another. Here, we discuss a variety of design choices made for each of these models. We adopt the UNet model architecture from DDPM \cite{ho} with a few slight modifications. Naturally, we adopt 3D convolutions rather than 2D convolutions since we are interested in 3D volumes, additionally because of the increased memory requirements of operating in 3D we replace all attention layers in the model with the less powerful, but more memory efficient, linear attention variant. We used a 3 layer UNet. The channel multiples for the local neighborhoods were 4,8, and 16 respectively for a total parameter count of $\sim$96 million parameters. For PolyMicros a larger model was used with channel multiples of 8, 16, and 24 for a model with $\sim$260 million parameters. 

Importantly the models were additionally modified to allow for switching of the padding style to change between training and inference. For the local models, the training patches are not periodic, however ideally the generated dataset would be periodic for two reasons. Firstly, The MOGRF outputs periodic fields, and secondly, it is often desirable for the generated volumes to be periodic for downstream analysis (for example to be simulated using FFT based PDE solvers). We therefore must reconcile the fact that the model is expected to learn patterns in a non-periodic context, but generate them in a periodic one. We found that training the models with replicate padding then switching the padding to periodic during evaluation time effectively solved this issue. Importantly we found that the inverse is not true. A model trained with periodic padding did not behave stably when it's padding was changed during inference. We therefore trained both the neighborhood models and PolyMicros with replicate padding, despite the fact that PolyMicros could be trained periodically given the synthetic data is periodic. This enabled stable switching of padding for PolyMicros at inference time to generate either a non-periodic or periodic volume.

With respect to other training considerations, we follow the EDM methodology with respect to all diffusion specific parameters (noise schedule, noise distributions sampling during training, optimizers, etc...) \cite{karras2022elucidating}. We train the neighborhood models for $200$ epochs with a learning rate of $0.00001$. The neighborhood models, being trained on smaller $32^3$ volumes, were trained with a batch size of 16 with 16 gradient accumulation steps for an effective batch size of 256. Each neighborhood model was trained on a single A100 80GB GPU and training took approximately 18 hours per model. Training was done a FP32 precision due to the small magnitude of ROGSH values.  

For the PolyMicros foundation model we faced additional challenges during training due to the larger model size as well as the the larger training volumes ($128^3$). This causes both a significant slowdown due to the increased number of computations, as well as from reduced batch sizes caused by limited VRAM. Because of this, we took several alternative actions to make training the PolyMicros model feasible. The primary tactic was to pre-train the PolyMicros model on the entire ensemble of 32 cubed patches aggregated from each neighborhood class. The idea here is that much of the initial training procedure is spent learning the general features of the dataset, and therefore by training on similar, but smaller volumes, we could get a larger effective batch size and speed up training. We then could go on and fine-tune the pre-trained model on the full $128^3$ synthetic dataset. This allowed us to spend far less time calibrating on the expensive high resolution data than would have taken to train from scratch. Additionally, rather than training on a single A100 80GB GPU we perform data parallel training across 8 H200 141 GB GPUS. For the pretraining step we used a batch size of 24 with 16 gradient accumulation steps for an effective batch size of 3,072 (aggregated across all 8 GPUS). We pretrained with a learning rate of $0.00001$ for $200$ epochs which in total took approximately 5 hours. For the final fine tuning we reduced the batch size down to 3 due to memory limitations and consequently increased the number of gradient accumulation steps to 64 to get an effective batch size of 1,536 across all 8 GPUs. We trained the model again with a learning rate of $.00001$ for another $50$ epochs which took just over 72 hours. All models were trained with an EMA decay of $0.9$.
\newpage
\section{Additional Experimental Results}
\label{app:additional_experimental_results}


\subsection{Reconstruction from Partial Observations}

\label{app:superres}

\begin{algorithm}
\caption{In-Painting Condition Function}\label{alg:inpaint}
\begin{algorithmic}[1]

\Require Mask of known values: $\bm{mask}$, Number of diffusion steps: $N$, Fraction of steps to condition: $\vartheta$

\Procedure{MaskedConditionFunc}{$\bm{x}, i$}

\If{$i < \vartheta N$ }
\State $\bm{x} \leftarrow \bm{mask}$ where $\bm{mask} \neq \texttt{None}$
\EndIf

\State \textbf{return} $\bm{x}$

\EndProcedure

\end{algorithmic}
\end{algorithm}

\FloatBarrier

In the first case study, we demonstrate PolyMicro's generalist capacity by using it to perform microstructure superresolution \cite{jangid2024q} without any problem -- or material system -- specific training or fine-tuning. We would like to reiterate that although other super-resolution methods exist, our goal in this work is to showcase that PolyMicros can serve to accelerate materials development efforts because it can be easily co-opted into aiding scientists with their own unique challenges. The process of co-opting PolyMicros is simple; it can be co-opted without the need for any new training or specialization because of its ability to be \textit{modularly conditioned}, Sec. \ref{sec:modular_conditioning}. Therefore, addressing new challenges simply amounts to identifying a suitable conditioning process \cite{bansal2023universal, sun2024provable, learnedconditionedDDPM, zeni2025generative, zhang2023adding}.

We perform microstructure super-resolution by utilizing PolyMicros to perform in-painting on a test structure not used during the data generation process \cite{BELADI20131404}, Fig. \ref{fig:CS1:CS1Over}. We utilize a very simple set-up. The subset of known pixel values is stored as a mask. Subsequently, during diffusion, the known pixel values are imposed (via direct replacement) the diffusion steps. Specifically, a masked conditioning function (Alg. \ref{alg:inpaint}) is used as the conditioning function in the base diffusion sampling algorithm (Alg. \ref{alg:stochastic_sampler}) to convert the unconditional diffusion process into the specific conditional diffusion processes needed to address super-resolution. We observed that the performance of this conditional diffusion process depended heavily on the order of Alg. \ref{alg:stochastic_sampler}. It is critical to perform conditioning \textit{after} diffusion.

\begin{figure}[!h]
\centering
\includegraphics[width=\linewidth]{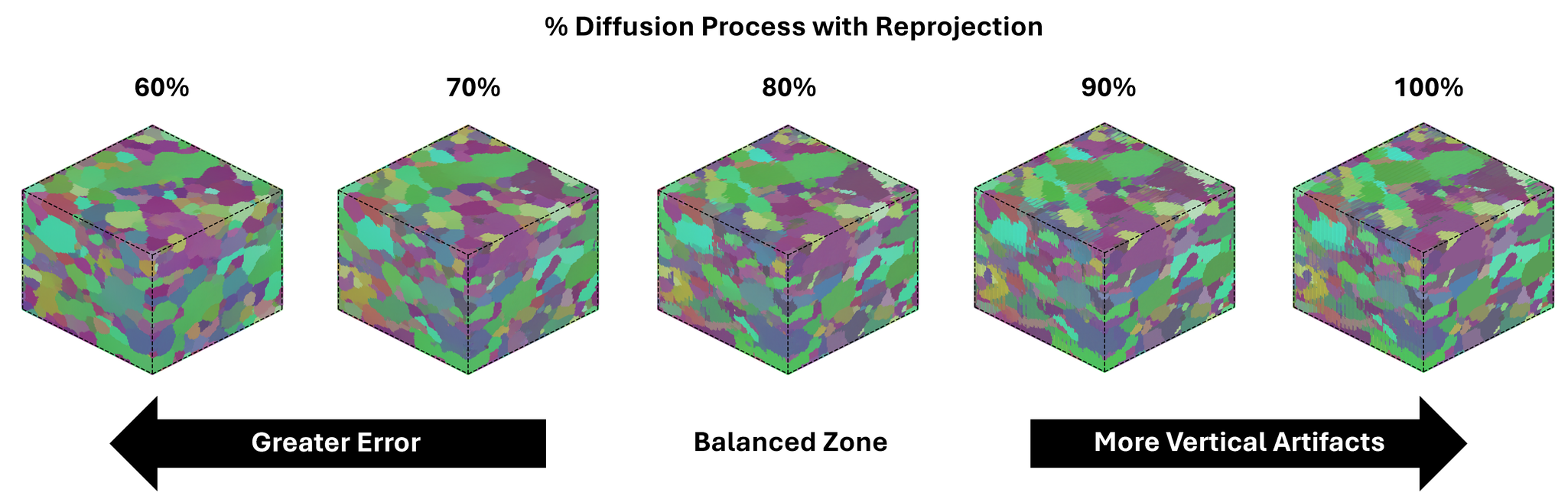}
\caption{Ablation testing the level of enforcement of the known values from the mask. After each diffusion step the known values from the low-resolution experiment are enforced through replacement. If this is done for all diffusion steps ($100\%$) vertical artifacts begin to appear due to small discrepancies between the true and predicted values. By performing the projection for only an initial portion of the diffusion process we see that these errors can be removed at the cost of reconstruction accuracy. Here we see that the vertical artifacts disappear between $70-80\%$ enforcement while still showing a strong adherence to the true structure.}
\label{fig:percAblation}
\end{figure}

We initially found that performing the projection of the known values after each diffusion step led to vertical artifacts (Visible in Fig. \ref{fig:percAblation} $100 \%$). As previously mentioned in Sec \ref{sec:cs1}, most of the error in the diffusion model is near the grain boundaries. Here the diffusion model may incorrectly predict the exact placement of the grain boundary, and therefore result in a misaligned grain boundary with respect the true boundary in the mask. This error while relatively small, results in vertical artifacts on the grain boundary where the projected values are misaligned with predicted values. To remedy this we chose only to perform the projection step only for the initial portion of the diffusion procedure, the idea being that if the true values are enforced for most of the diffusion process the generated structure will be similar to the target, but by stopping early the diffusion model will have the opportunity to remedy the artifacts and lead to higher quality generations. This naturally induces some error, as the known slices will not be represented exactly in the final structure. However, the results from Sec. \ref{sec:cs1} show that the induced error is minimal, and that this is an acceptable tradeoff to remove the non-physical artifacts. The only additional consideration introduced is for what portion of the diffusion process should the projection be performed. To determine this we performed a small ablation using 100 diffusion steps and varying the percent of the initial diffusion process where the projection was performed. The results of this are shown in Fig. \ref{fig:percAblation}. Overall we found the generation process very stable, and that there was very little sensitivity to parameter. Varying the ratio from $60-100 \%$ we can see that for lower percentages ($<60 \%$) the generated structures  begin to smooth out, and resemble the original structure to a lesser degree, while for higher percentages ($>90\%$) the vertical bands once again become apparent. We found a choice of $75\%$ worked well, and was a good choice even when the total number of diffusion steps was decreased.

\begin{figure}[!h]
\centering
\includegraphics[width=.98\linewidth]{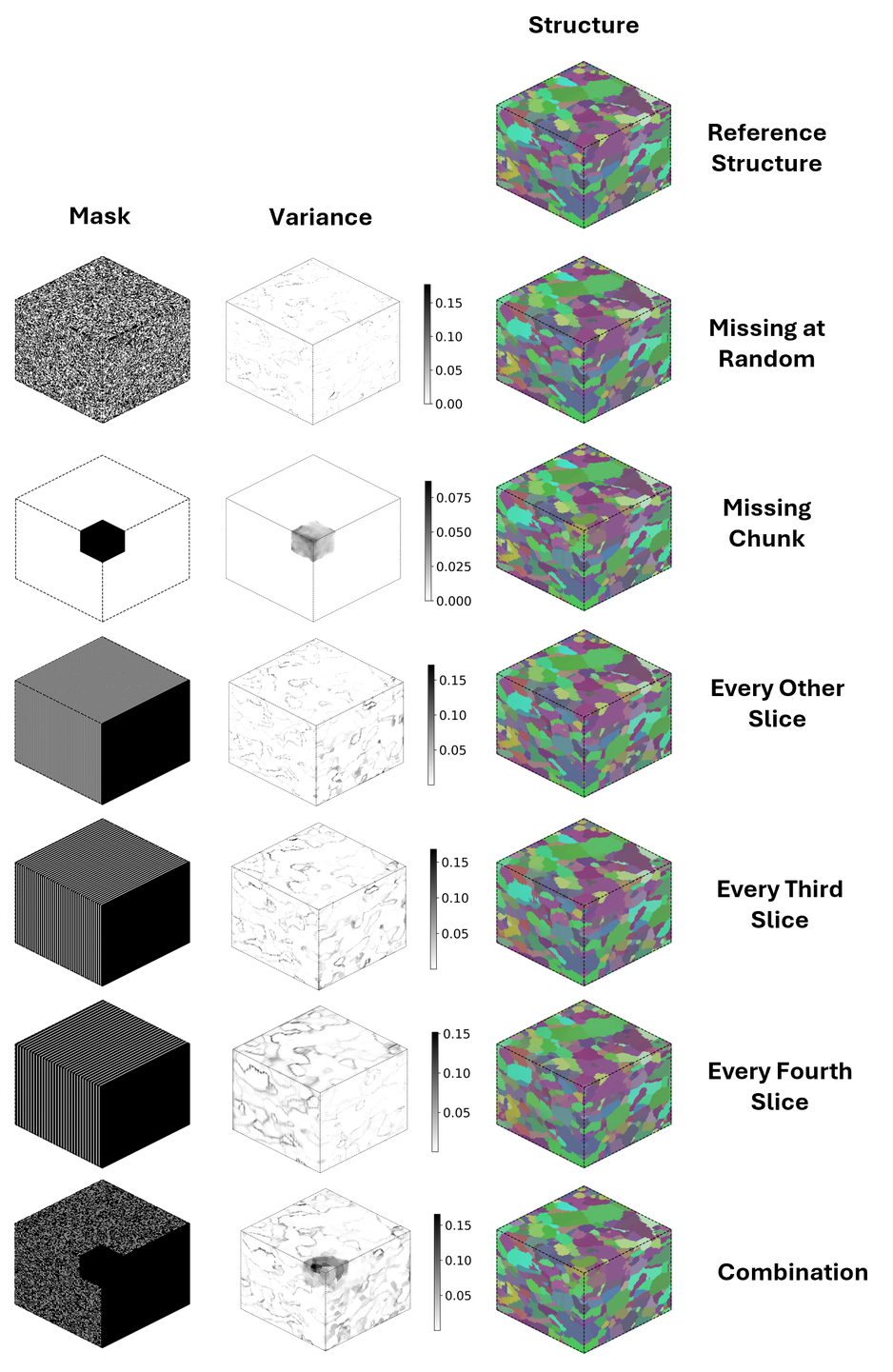}
\caption{PolyMicro's post-training conditioning performance on a variety of potential inpainting tasks incountered in experimental microscopy.}
\label{fig:masks_survey}
\end{figure}

In the main body we explore the original 4X superresolution challenge addressed by Jangid \textit{et al.} \cite{jangid2024q}. However, because the PolyMicros superresolution solution is modular, it is not specialized to this specific type of superresolution. Fig. \ref{fig:masks_survey} illustrates PolyMicros performance for a variety of different inpainting tasks that arise naturally in microscopy (e.g., removing random noise, removing corrupted regions, etc.). Again, these different problem settings are easily represented as masks and addressed using the exact same approach. We see generally good performance across these tasks, indicating that the PolyMicros Model exhibits good stability.

\subsection{Reconstruction from Partial Statistics}
\label{app:RPS}

\begin{algorithm}
\caption{Dimensionality Expansion Condition Function}\label{alg:dim_expansion}
\begin{algorithmic}[1]
\Require Target Orthogonal 2-Point Statistics: $f_{\perp}^{\beta \gamma}$, Threshold Schedule: $\varrho_{i \in \{0,...,N\}}$, Learning Rate: $\alpha_{lr}$
\Procedure{OrthoConditionFunc}{$\bm{x}, i$}

\State err $\leftarrow \infty$

\While{$\text{err} > \varrho_{i}$}

\State $\hat{f}_{\perp}^{\beta \gamma} \leftarrow \texttt{Ortho2PS}(\bm{x})$
\State $\text{err} \leftarrow \| \hat{f}_{\perp}^{\beta \gamma} - f_{\perp}^{\beta \gamma}\|^2_2$

\State \( \bm{x} \leftarrow \bm{x} - \alpha_{lr} \nabla_{\bm{x}} \text{err} \)

\EndWhile

\EndProcedure

\end{algorithmic}
\end{algorithm}

The second explored case study in experimental microscopy is microstructure dimensionality expansion. Here, the goal of microstructure dimensionality expansion is to avoid 3D characterization methods entirely. Instead, the aim is to generate synthetic 3D microstructures utilizing information collected from a select few 2D experimental observations. A common choice is 2D characterization of three orthogonal planes, Fig. \ref{fig:CS2:CS2Over}. At surface level, this problem differs from super-resolution because the amount of conditioning information is roughly an order of magnitude lower. However, there is a second, more important, difference: experimentation strategies often do not record the relative position of the collected 2D slices (e.g., they might come from different samples). This means that microstructure dimensionality expansion is not simply a highly data-scare inpainting problem. Instead, information from the collected slices must be compatibly integrated \textit{and} the missing information must be accounted for. Integration is traditionally performed by informing 3D generation using microstructure statistics extracted from the 2D experiments, instead of the direct experimental values themselves. The only constraints on the specific microstructure statistic used is that they must only rely on relative position information, and their computation must be differentiable (needed later). This task is notoriously difficult in polycrystalline materials \cite{2Dto3DGAN, lee2024multi, paxti_polycrystal_diffusion}. As of the time of writing, successful techniques are only capable of enforcing simple mean field statistics \cite{dream3d, kanapy, buzzy2024statistically}. As a result, they are unable to account for long range spatial patterns.

Here, we co-opt PolyMicros to perform 2D-to-3D dimensionality expansion. To demonstrate the procedure, we utilize 2-point spatial correlations as our statistical measure since they meet the necessary constraints, and are known to correlate strongly to the physical properties of the material. Therefore, if these statistics are matched we can expect the synthetic 3D volumes to display similar properties as the experimental reference. 

Fig. \ref{fig:CS2:CS2Over} outlines the dimensionality expansion procedure. Similarly to the super-resolution case study, we simulate performing the 2D experiments by taking them from an experimental reference, so the final generated structures can be compared to the original. For this experiment we simulate taking three 2D images along orthogonal cross sections of the material -- a common experimental procedure for gaining some sense of the 3D geometry of a material without performing a full 3D experiment. The dimensionality expansion protocol is as follows. First, we compute the 2-point spatial correlations for each of the three orthogonal views (i.e., 2D 2-point statistics). We take these statistics to estimate the subset of the full 3D 2-point correlations which lie along the three orthogonal planes. Notably, we can do this because the 2-point statistics only rely on relative positions of microstructure observations. Therefore, while we do not know the full 3D coordinates of an observation (we will be missing the x, y, or z component depending on the plane in question), we do know that the relative shift in position in the unknown component for the observations within the plane will be zero. Therefore, we can use the 2D observation to estimate the subset of the 3D 2-point statistics for which one of the three components is zero. Now, given a 3D volume (e.g., in our procedure: a microstructure generated -- or partially generated -- by the diffusion model), we can compare the subset of the generated microstructure's 3D 2-point statistics with the orthogonal experimental statistics. Since the computation of the 2-point statistics is differentiable via autograd, we can pose the generation process as an optimization, where we optimize some volume such that it matches the known statisics. Historically, this direct optimization based 2-point statistics conditioned generation procedure is known to fail to converge to reasonable results for complex material systems such as polycrystals\footnote{In fact, the optimization will fail even if all the 2-point statistics are known.} \cite{seibert_localoptimization, seibertDiffusion, marreddy_2D, marreddy_3D}. Instead, in this work we interleave these optimization steps with diffusion steps using the PolyMicros model. In this way, the PolyMicros model serves as a prior regularizing the optimization procedure. Theoretically, this is analogous to conditioning the PolyMicros model using a proximal descent type likelihood in order to approximate a conditional generative process \cite{sun2024provable}. In practice however we take several liberties to make such a conditioning process possible. The first change is that rather than mixing the updates from the diffusion process and the optimization, as would be implied by theory, we instead found that staggering the updates (doing one at a time) produced much better results. Furthermore, we found that rather than doing one diffusion step then a single optimization step, that the generative procedure converged much faster, and with greater accuracy, if we performed many more (by a factor of several thousand) optimization steps between each diffusion step. Rather, than prescribing a fixed amount to this ratio, we instead chose to perform the optimization process till the average difference between the known and generated statistics converged beneath a specified value defined via the linear threshold schedule $\varrho(t) = (N - t)*(1e^{-5}) + 1e^{-7}$ where $N$ is the number of diffusion steps, and $1e^{-5}$ and $1e^{-7}$ are the initial and final thresholds respectively  . These changes took the conditioning procedure from failing to work in any capacity to producing exceptional results. Algorithmically, an optimization based conditioning function (Alg. \ref{alg:dim_expansion}) is used as the conditioning function in the base diffusion sampling algorithm (Alg. \ref{alg:stochastic_sampler}) to convert the unconditional diffusion process into the specific conditional diffusion processes needed to address 2D-to-3D dimensionality expansion.

Several samples from this generative procedure can be seen in Fig. \ref{fig:CS2:CS2Over}. Qualitatively comparing them to the experimental reference we can see that much of the morphology is the same. Both the generated and experimental structures share a vertical greenish-blue band which alternates with a more random purplish region to either side. It is important to note that since the enforced statistics are relative the generated structures are free to have their patterning shifted in space (i.e the band will exist in the same orientation, but may be shifted left or right in the generated structures). This ability to faithfully recreate long range patterns is a significant improvement over any pre-existing approach \cite{dream3d, kanapy, buzzy2024statistically}. Furthermore, we can see that the generated structures form reasonable grain shapes similar to those found in the original images, and only suffer from a minor amount of blurring. We also see that, relative to the previous case study, the qualitative variety of the generated microstructures is much higher. For example, the location of the bands and the specific generated grains vary significantly. This is an important result and aligns with our expectations for this problem. As the amount of information provided in the conditioning is far less, we should expect a commensurate increase in the uncertainty of our reconstructions. 

\begin{figure}[!h]
\centering
\begin{minipage}[b]{0.3\textwidth}
    \includegraphics[width=\textwidth]{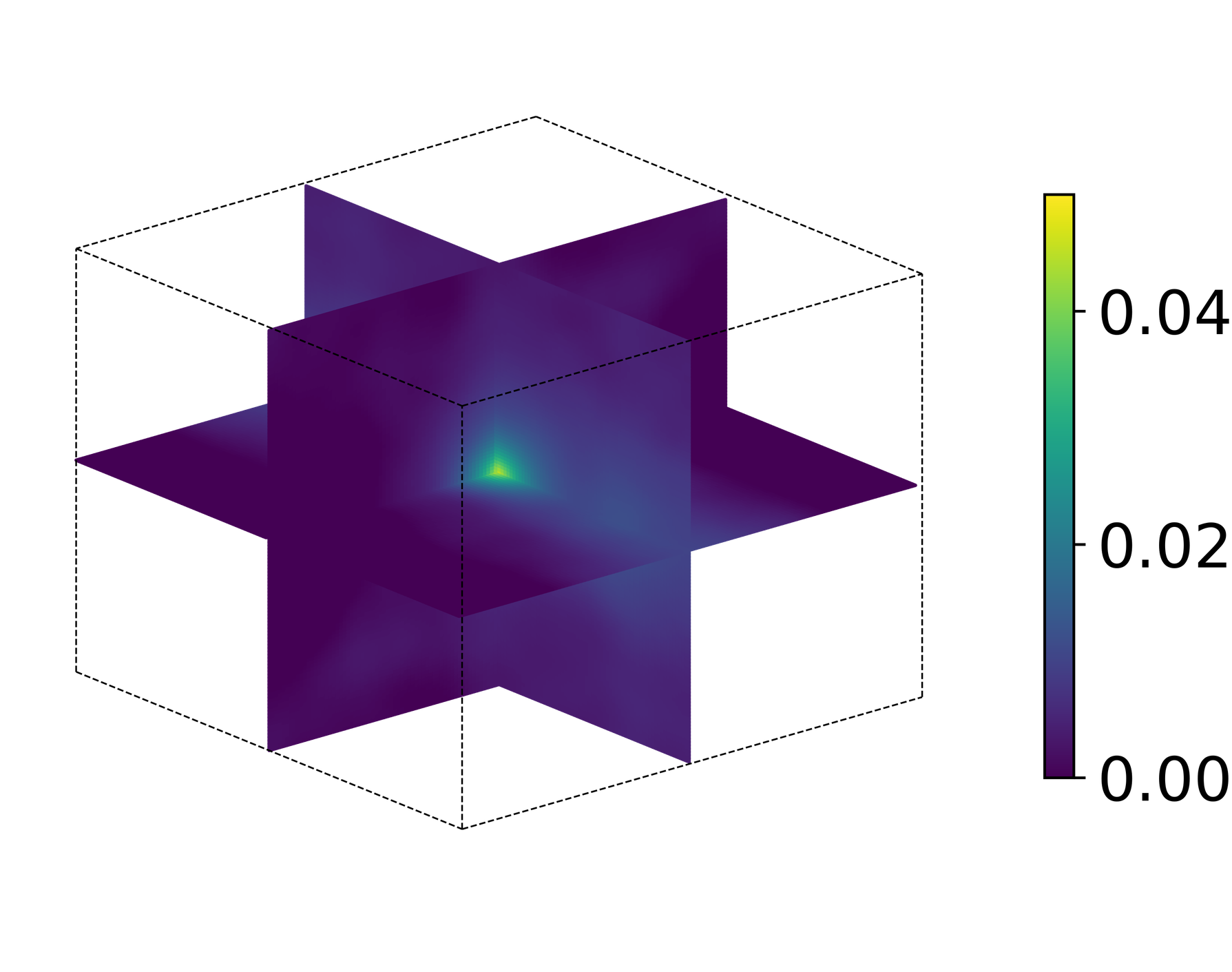}
\end{minipage}
\hfill
\begin{minipage}[b]{0.3\textwidth}
    \includegraphics[width=\textwidth]{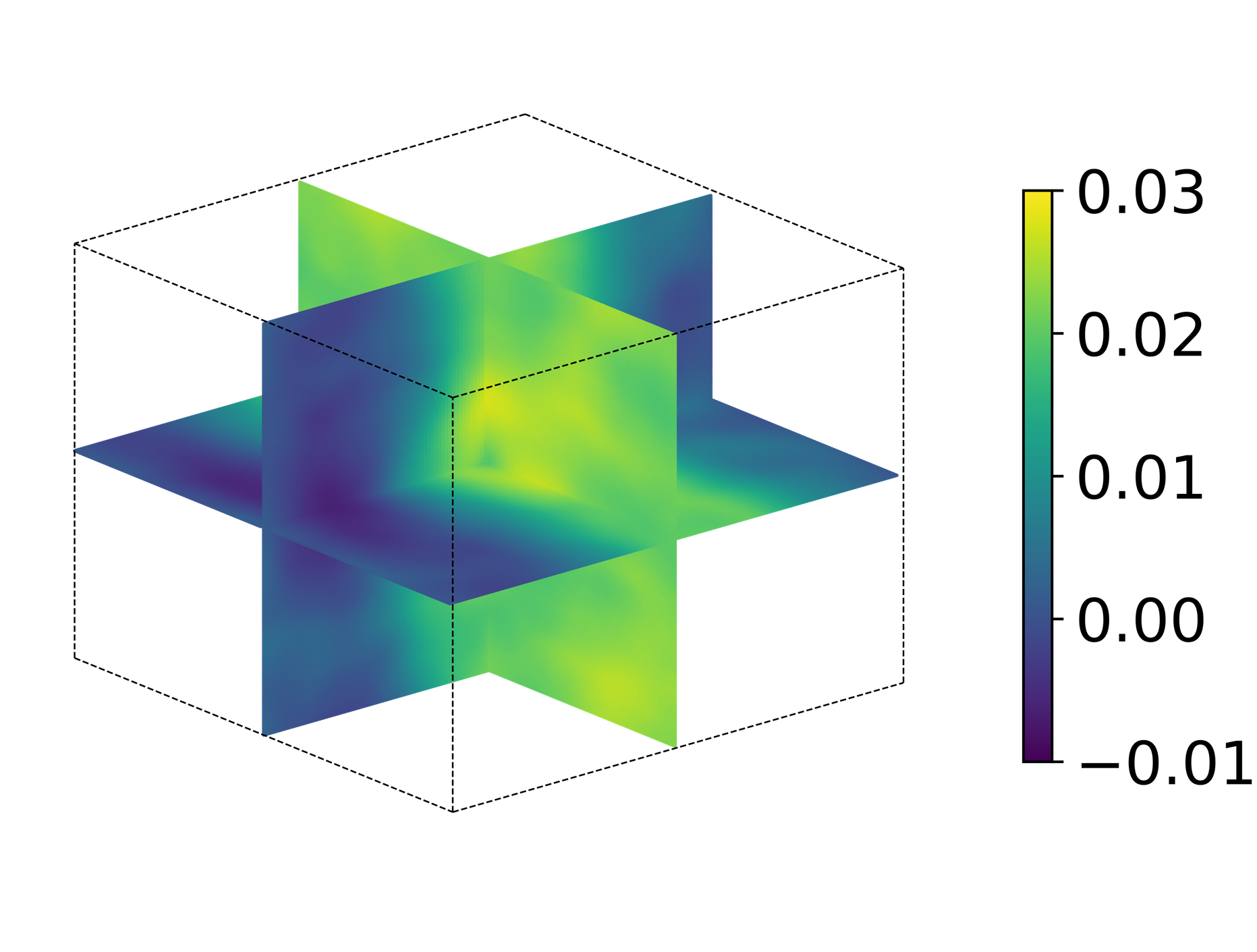}
\end{minipage}
\hfill
\begin{minipage}[b]{0.3\textwidth}
    \includegraphics[width=\textwidth]{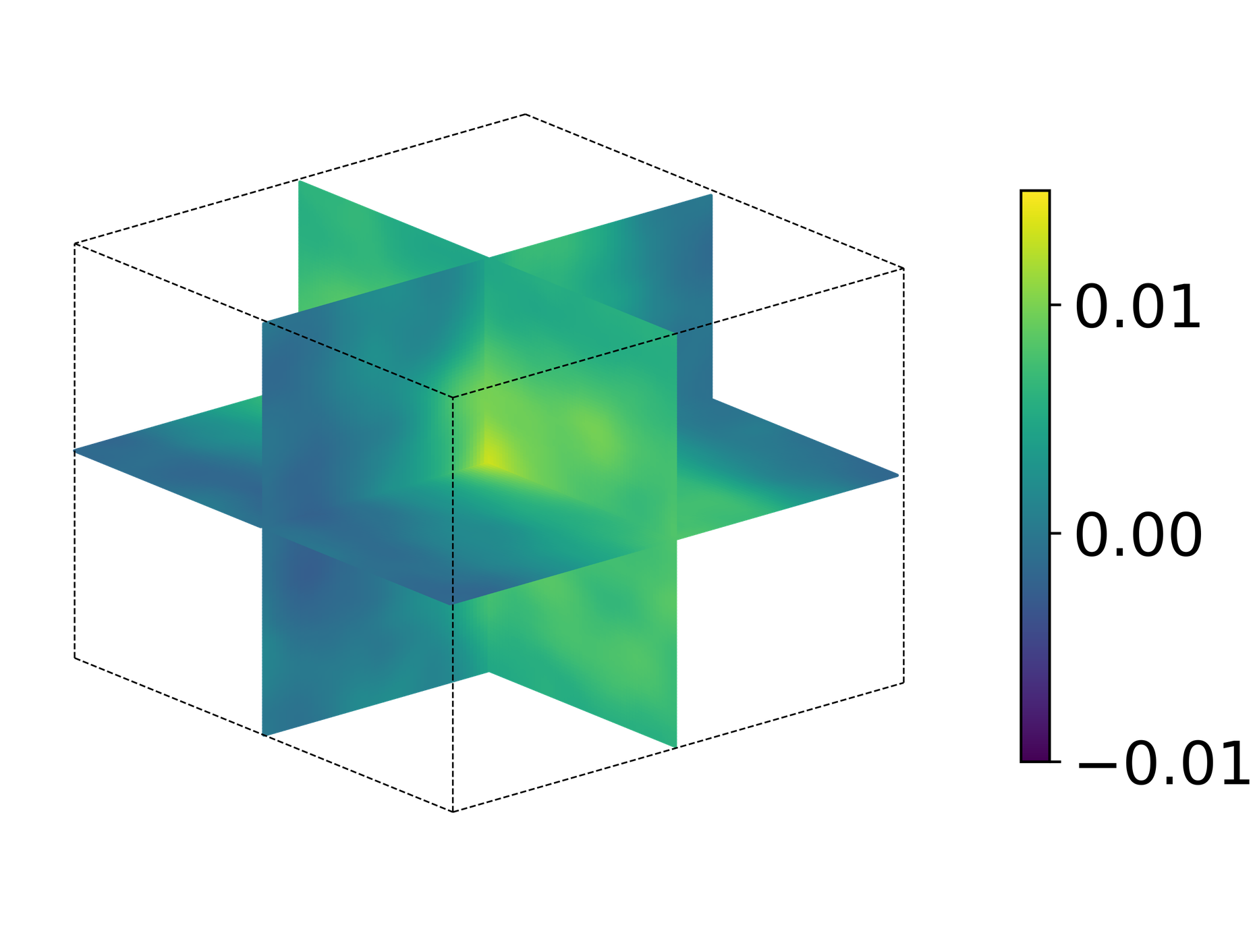}
\end{minipage}

\begin{minipage}[b]{0.3\textwidth}
    \includegraphics[width=\textwidth]{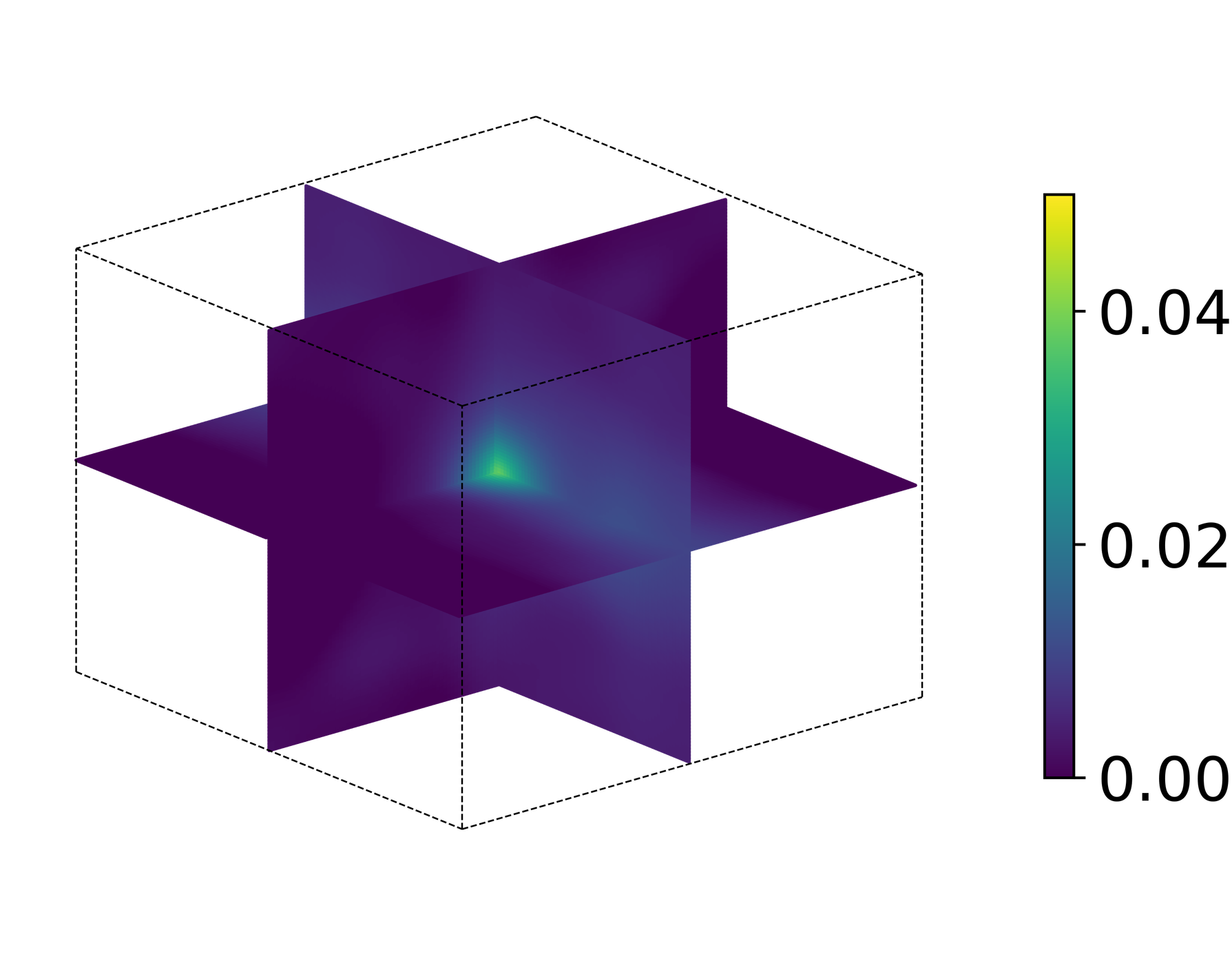}
\end{minipage}
\hfill
\begin{minipage}[b]{0.3\textwidth}
    \includegraphics[width=\textwidth]{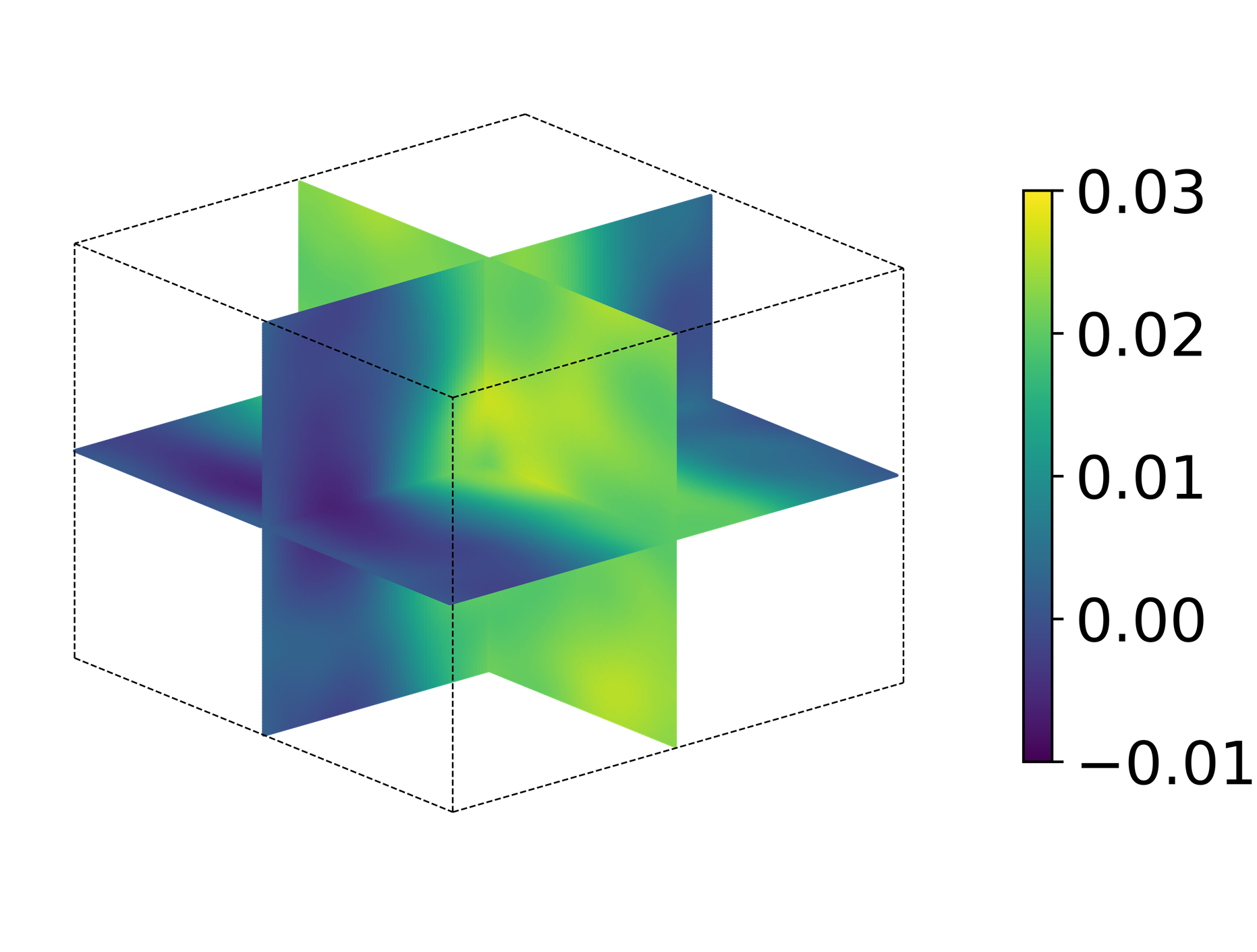}
\end{minipage}
\hfill
\begin{minipage}[b]{0.3\textwidth}
    \includegraphics[width=\textwidth]{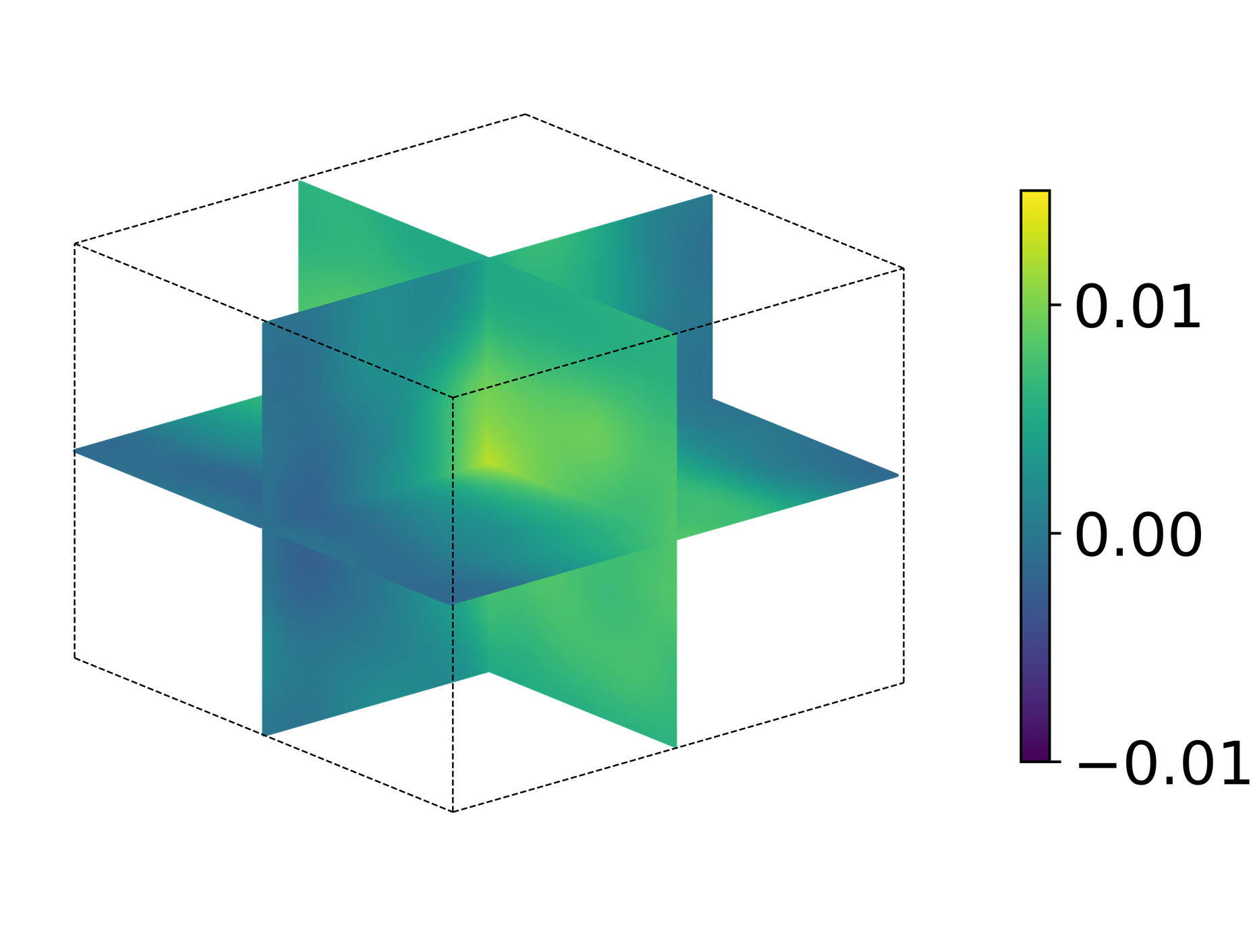}
\end{minipage}
\begin{minipage}[b]{0.3\textwidth}
    \includegraphics[width=\textwidth]{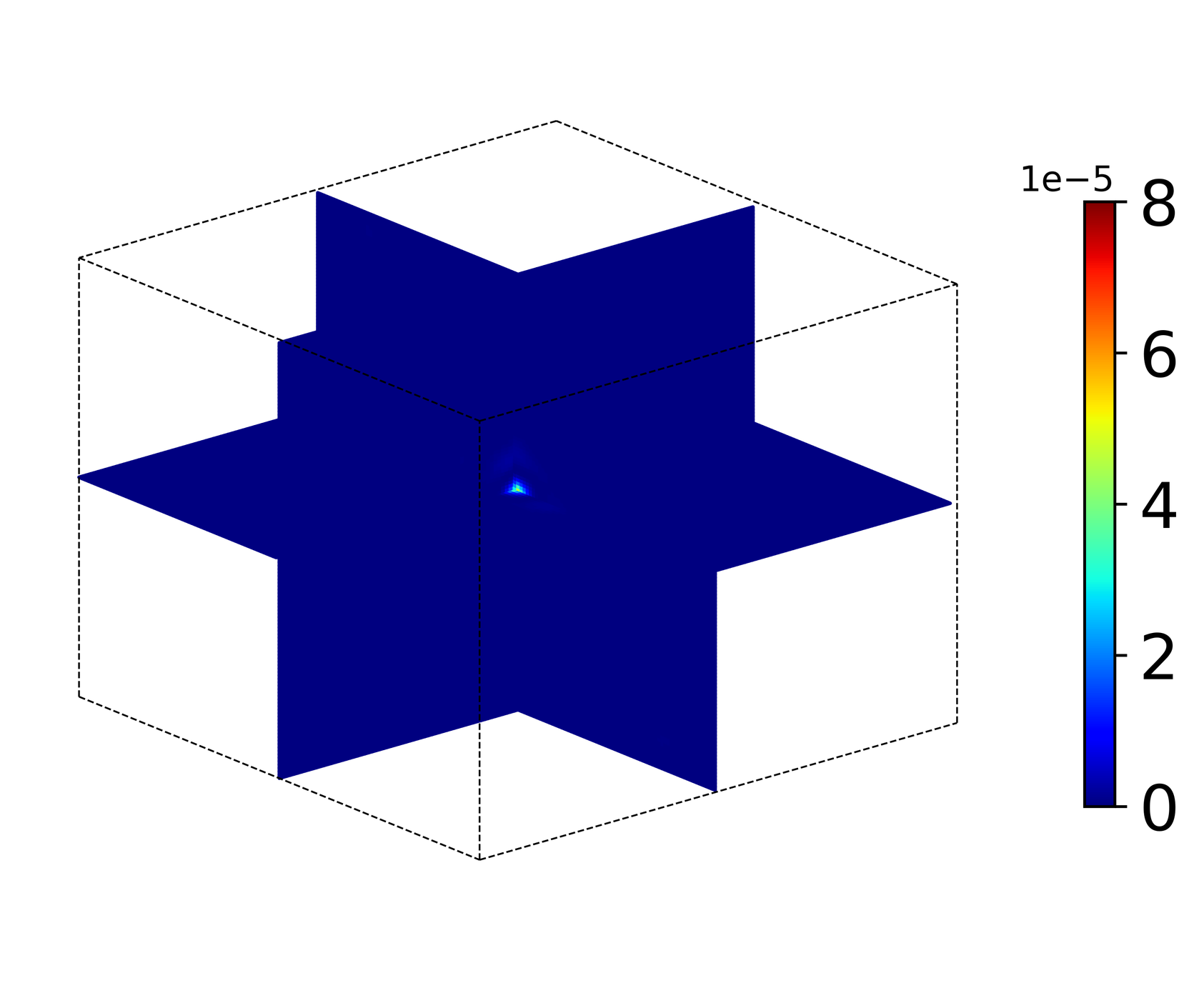}
\end{minipage}
\hfill
\begin{minipage}[b]{0.3\textwidth}
    \includegraphics[width=\textwidth]{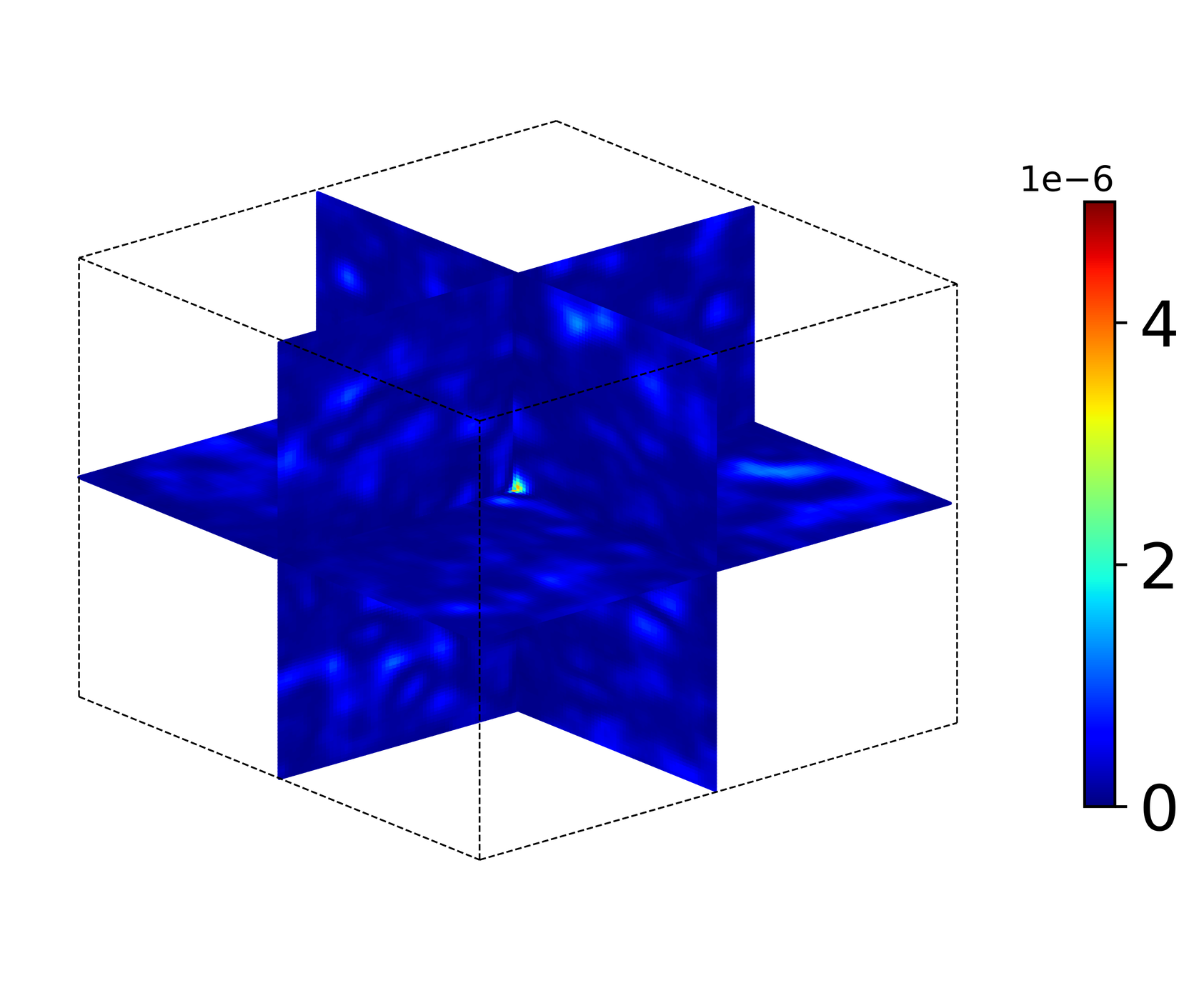}
\end{minipage}
\hfill
\begin{minipage}[b]{0.3\textwidth}
    \includegraphics[width=\textwidth]{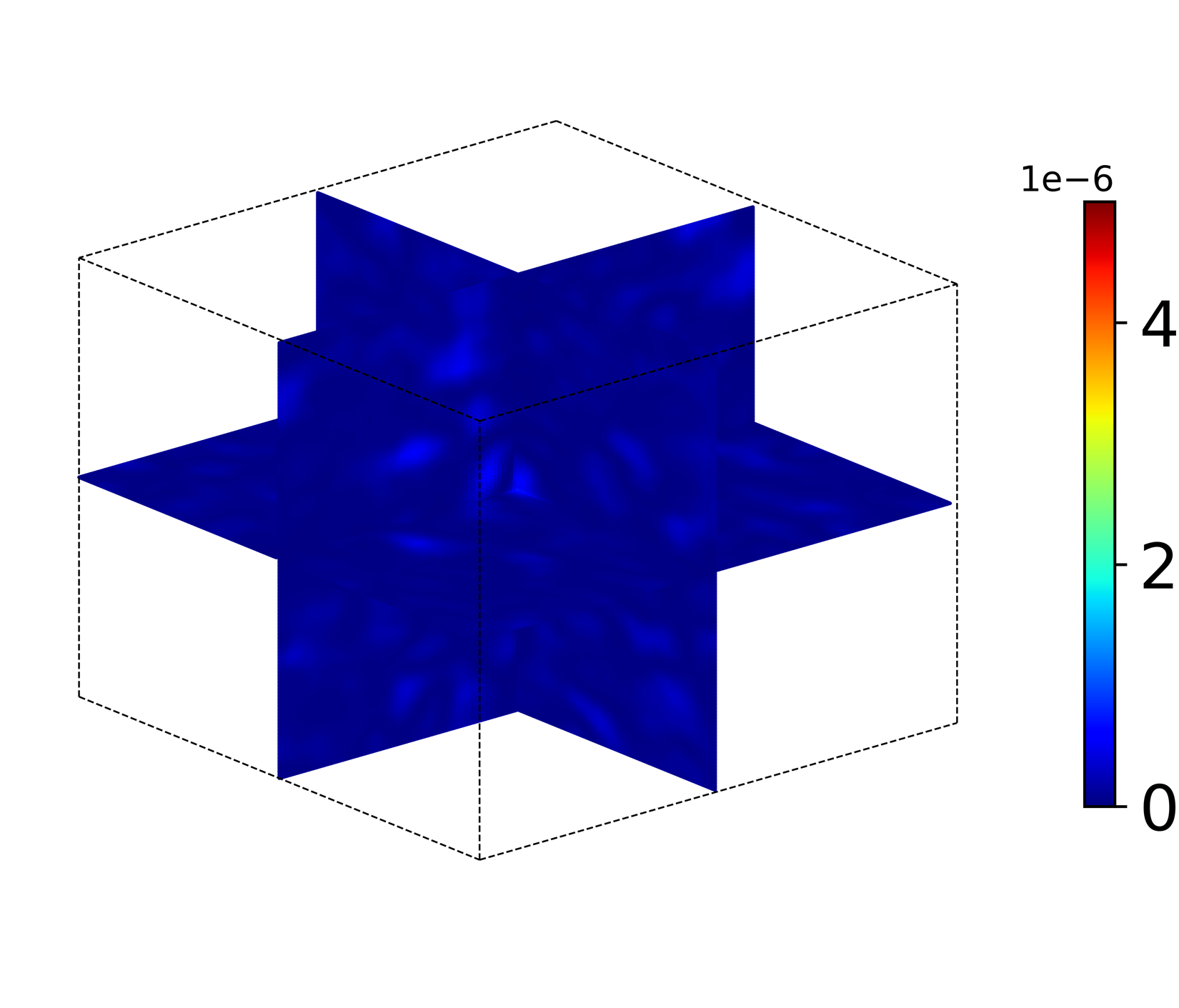}
\end{minipage}
\caption{Row One: Target Orthogonal Auto- and Cross-Correlations. Row 2: Average Orthogonal Auto- and Cross-Correlations of Generated Structures, Row 3: Difference}
\label{fig:CS2:Stats}
\end{figure}

Finally, we can also analyze performance quantitatively by comparing the generated orthogonal 2-point statistics against the original target 2-point statistics from the 2D experiments. Inspecting Fig. \ref{fig:CS2:Stats}, we compare the ensemble average of the generated microstructures' statistics relative to the target 2-Point statistics obtained from the 2D experimental data. Here, we can see that they match almost exactly with the only visual difference between them being a minor amount of smoothing in the generated samples. This is likely due to them being averaged over many samples. The errors are several orders of magnitude smaller than the magnitude of the statistics, and match almost everywhere. We see a small deviation in the first auto-correlation's peak indicating a small deviation in the averages between the structures. Overall, this result is incredibly precise, and is the first successful attempt at performing the microstructure dimensionality task for polycrystalline materials beyond first-order accuracy.

\FloatBarrier

\section{Limitations \& Next Steps}
\label{app:limitations}

The PolyMicros Foundation Model already demonstrates the ability to provide practical utility to materials scientists by modularly addressing a variety of problems in microscopy. More generally, we argue that its general modular conditioning ability -- inherited from its diffusion model base -- means that it can be further applied to solve general structure-property inverse problems. In this sense, its current practical value is the ability to regularize solutions across a wide variety of salient inverse problems. In addition, we argue that the PolyMicros Foundation Model exemplifies a foundational framework for conceptualizing the development of foundation models for the sciences. Specifically, building these models as generalized priors -- around the idea of modular conditioning -- overcomes the problem heterogeniety present in many tasks; seemingly dissimilar problems can be solved using a centralized tool through the unification of Bayesian inversion. 

In spite of these strengths, PolyMicros displays several important limitations that open important areas for further development. Primarily, we argue that these weaknesses arise in the curation of the PolyMicros dataset, unsurprising, given that data curation is the core challenge in building foundation models for data-limited scientific domains. The first outstanding limitation is the need for more experimental data. Although the framework proposes and utilizes an LGD based augmentation scheme, this augmentation is not without its limits. This is most evident in Fig. \ref{fig:DGEN:PCA}. The PolyMicros dataset extends beyond the initial experimental microstructure dataset's diversity. However, the framework cannot extend arbitrarily far. Further, it does not extend uniformly. We hypothesize that this horizon primarily arises because there remains a strong coupling between the local neighborhood -- defined by the source experimental image -- and the realizable microstructure statistics (the 2-point statistics). Specifically, the neighborhood diffusion model will ignore the spatial patterning requested by the conditioning microstructure statistics if the two are in strong disagreement. The precise definition and conditions defining `strong disagreement' remains an open question beyond the scope of this work. We will discuss methods for reduces this coupling and extending the horizon momentarily. However, even with this methods available, we argue that there is likely a limit to how far one should reasonably extrapolate. This is because microstructures augmented far from their source experimental microstructure are likely increasingly less plausible. As a result, expanded experimental datasets -- focusing on quality and statistical diversity \textit{not} scale -- are necessary to seed and improve the data augmentation process described here. 

The second limitation is with respect to number of voxels. Material microstructures -- and polycrystalline microstructures in particular -- are practically challenging because their myriad features impose several requirments on the voxel grid resolution and size. High resolution grids are necessary to resolve salient local features -- such as grain boundaries. However, large voxelized domains are necessary to resolve salient long range patterns -- the patterns dictated by the 2-point statistics -- arising at the lengthscale of many grain arrangements. Further, the physics of polycrystalline systems (i.e., slip systems and slip based deformation \cite{cpfem_raabe, mcdowellviscoplast}) means that 3D domains are needed to realistically represent these materials. Altogether, these requirements produce a memory demanding engineering challenge. In this work, we limit ourselves to $128^3$ meshes to simplify the engineering task and align ourselves with the upper end of simulation sizes in polycrystalline crystal mechanics. However, this mesh resolution leads to weak separation between the grain scale and the grain pattern scale. This produces coupling which limits the achievable diversity in long range patterns. Extending mesh sizes to larger values will allow significantly more diverse datasets to be built (e.g., MICRO2D uses $256^2$ \cite{robertson2024micro2d}). 

The third limitation is with respect to material symmetry. In this work, we focus on a single symmetry system: cubic. However, real material systems display many other symmetries. Similar dataset efforts in these domains are necessary. 

\FloatBarrier

\newpage
\section{Reference Images}

\subsection{MOGRF Fields produced by MOSM Kernel}

\begin{figure}[h!]
\centering
\includegraphics[width=\linewidth]{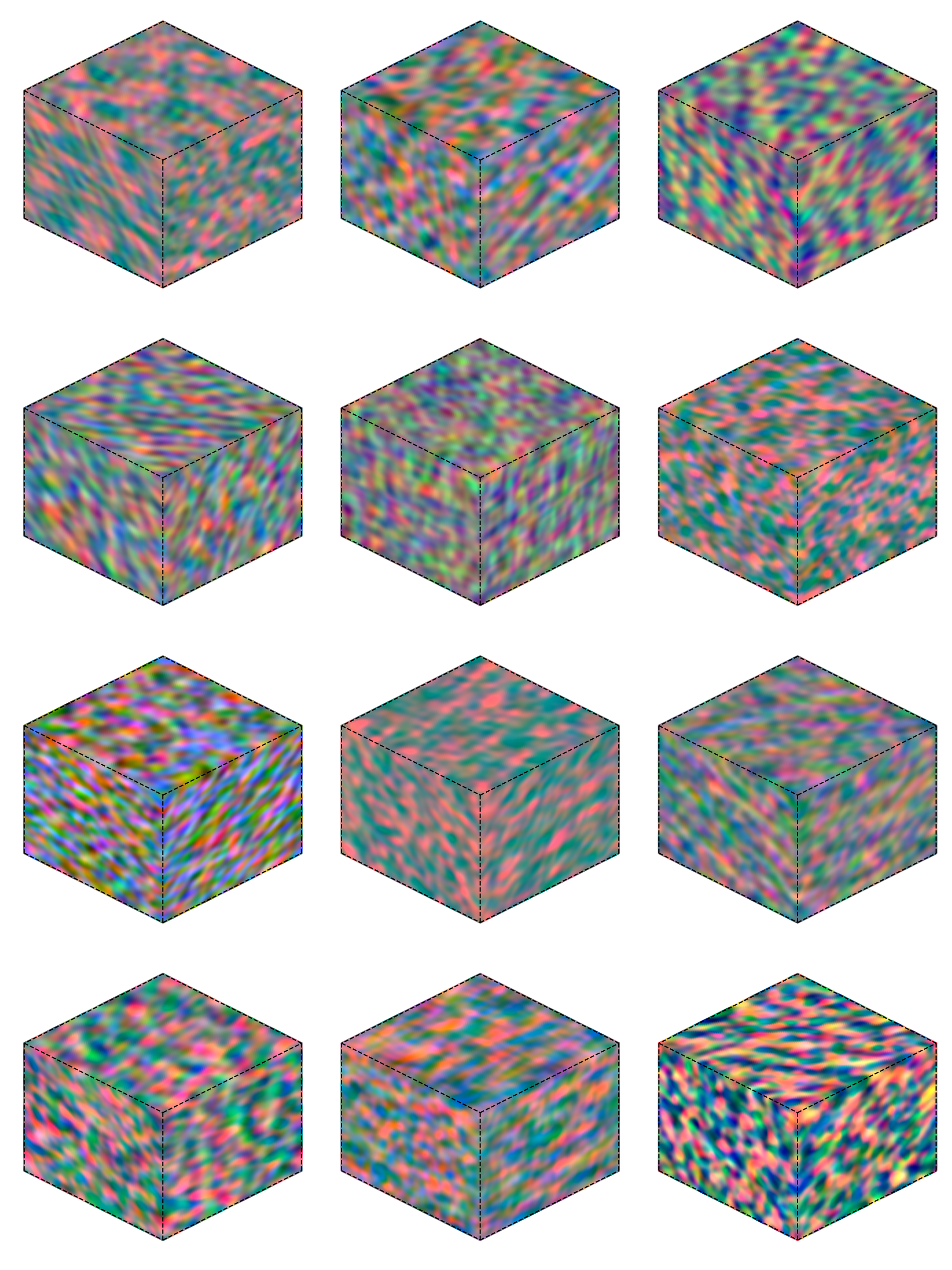}
\caption{Example Multi-Output Gaussian Random Field samples. The MOGRF covariance structure is seeded using randomly parameterized MOSM kernels.}
\label{fig:GlobalApprox:GRF}
\end{figure}

\FloatBarrier

\newpage
\subsection{Example Experimental Neighborhoods}

\label{app:hoods}

\begin{figure}[!h]
\centering
\includegraphics[width=\linewidth]
{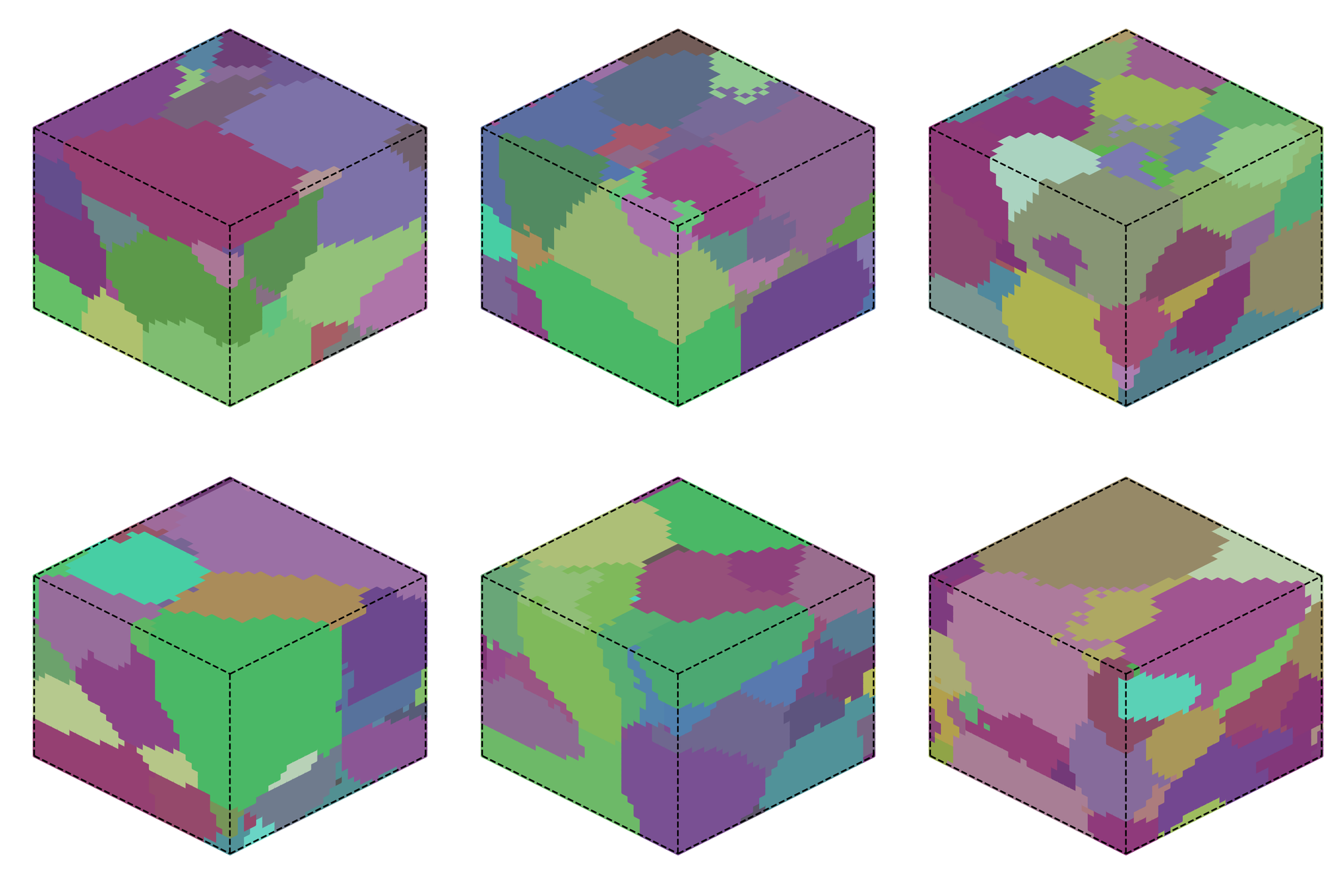}
\caption{Six Example Neighborhoods extracted from the AL2219R Experimental Microstructure \cite{vaughan2024mechanistic}. (Size of 
$32^3$)}
\label{fig:LocalRef:AL2219R}
\end{figure}

\begin{figure}[!h]
\centering
\includegraphics[width=\linewidth]{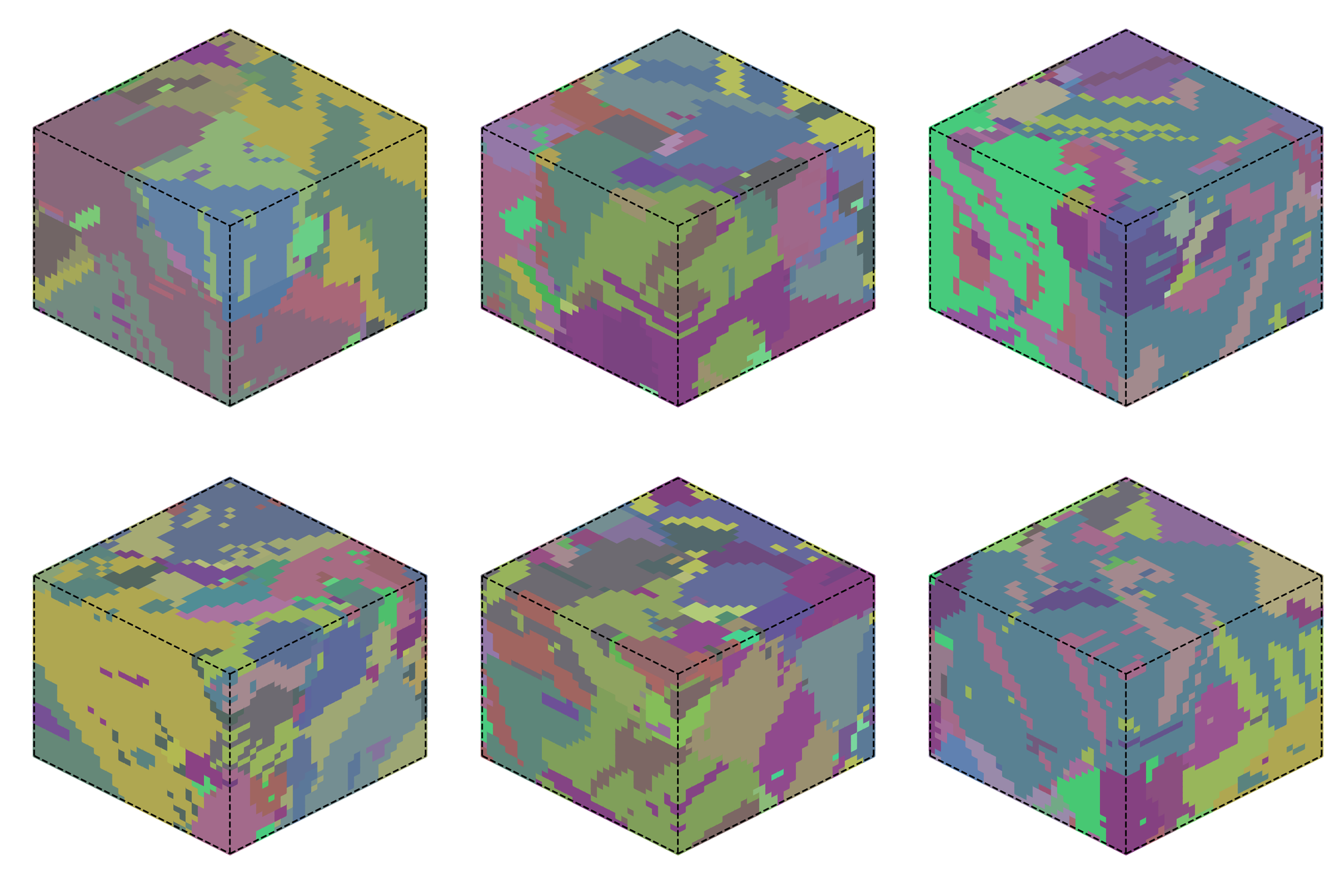}
\caption{Six Example Neighborhoods extracted from the Inconel625AM Experimental Microstructure \cite{chapman2021afrl}. (Size of $32^3$)}
\label{fig:LocalRef:Inconel625AM}
\end{figure}

\begin{figure}[!h]
\centering
\includegraphics[width=\linewidth]{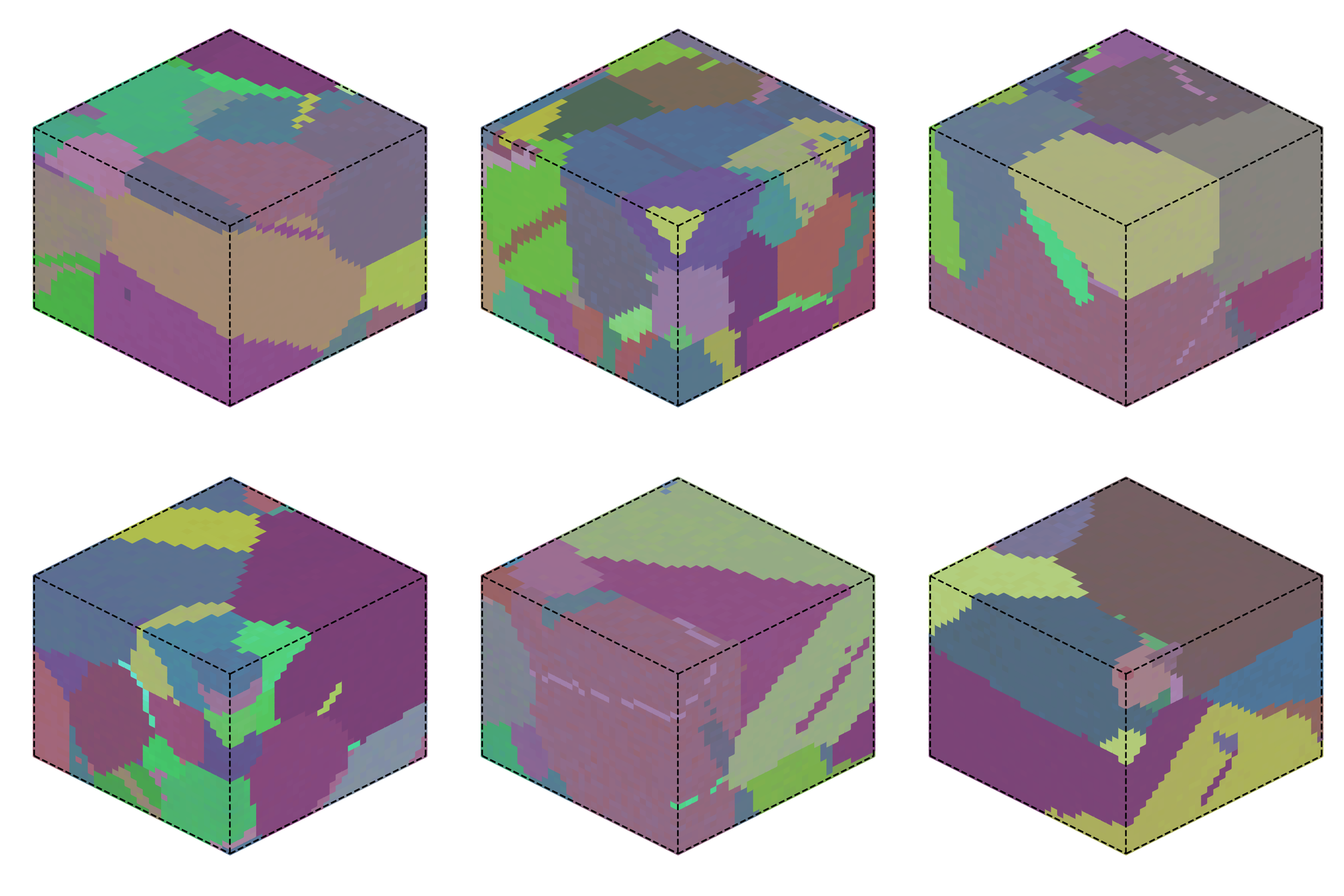}
\caption{Six Example Neighborhoods extracted from the Inconel718W Experimental Microstructure \cite{stinville2022multi}. (Size of $32^3$)}
\label{fig:LocalRef:Inconel718W}
\end{figure}

\begin{figure}[!h]
\centering
\includegraphics[width=\linewidth]{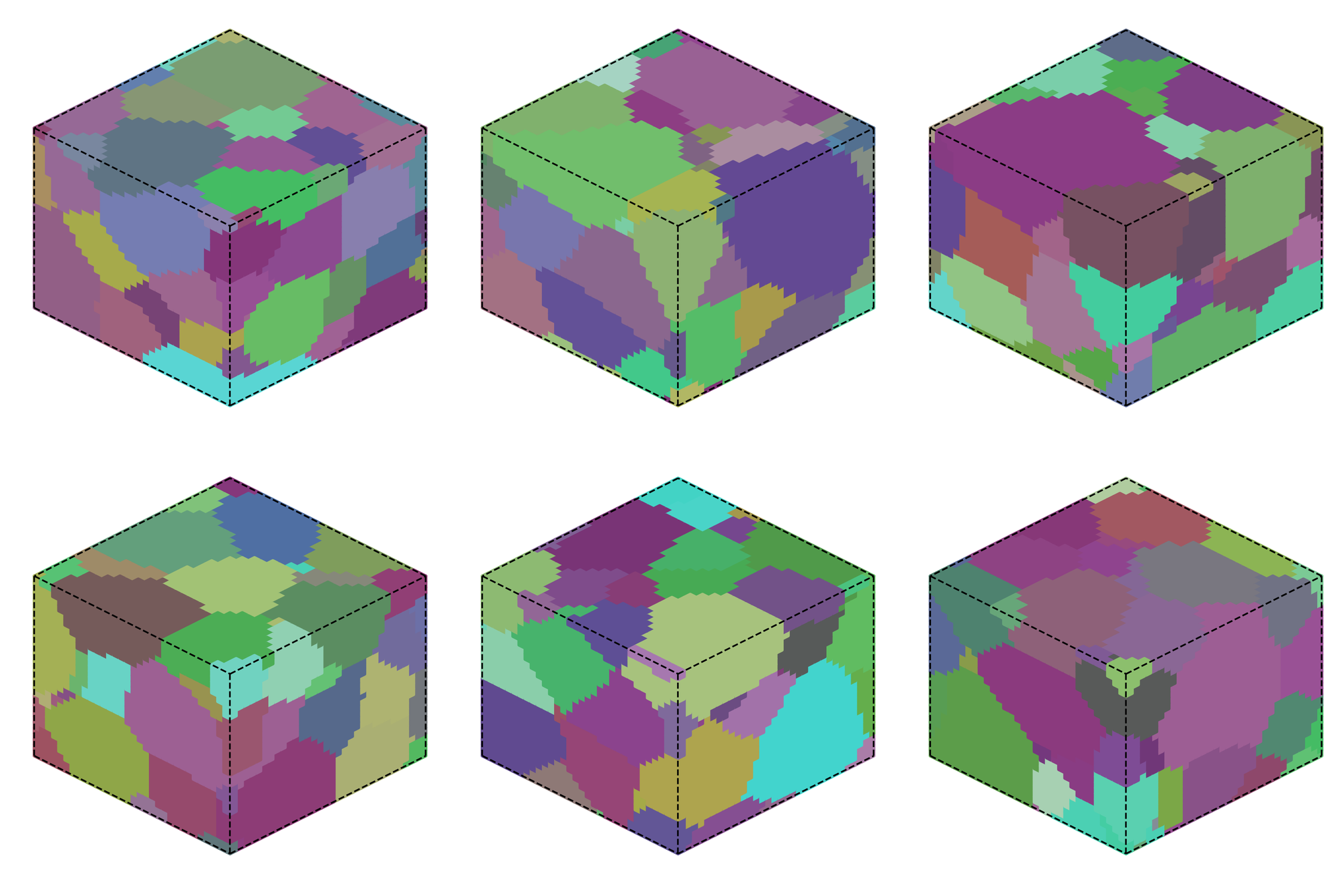}
\caption{Six Example Neighborhoods extracted from the NRL Experimental Microstructure \cite{rowenhorst2010three}. (Size of $32^3$)}
\label{fig:LocalRef:NRL}
\end{figure}

\begin{figure}[!h]
\centering
\includegraphics[width=\linewidth]{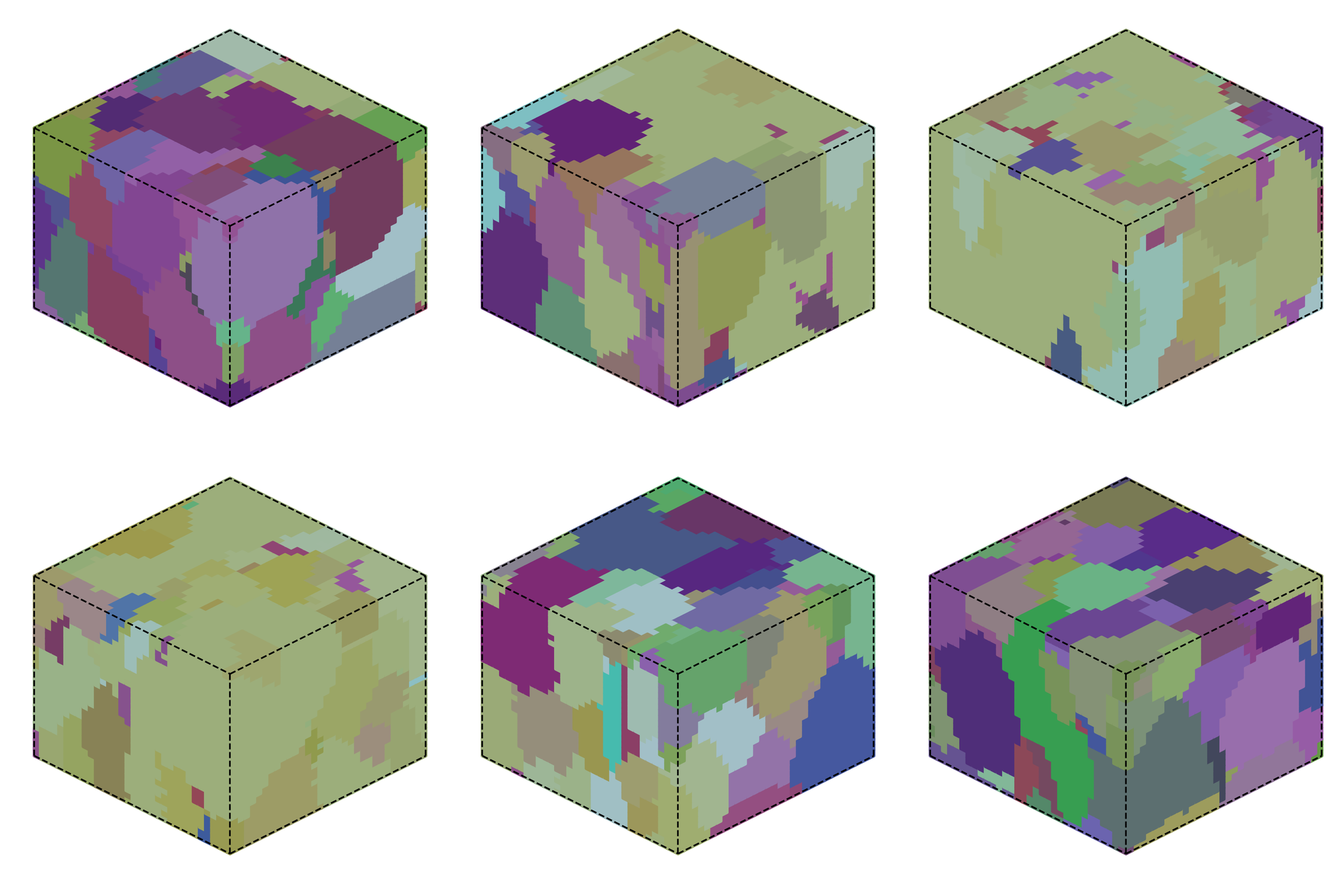}
\caption{Six Example Neighborhoods extracted from the TI64R Experimental Microstructure \cite{jangid2023titanium}. (Size of $32^3$)}
\label{fig:LocalRef:TI64R}
\end{figure}

\FloatBarrier

\newpage
\subsection{Samples From Synthetic Dataset}
\label{app:dataset}

\begin{figure}[h!]
\centering
\includegraphics[width=\linewidth]{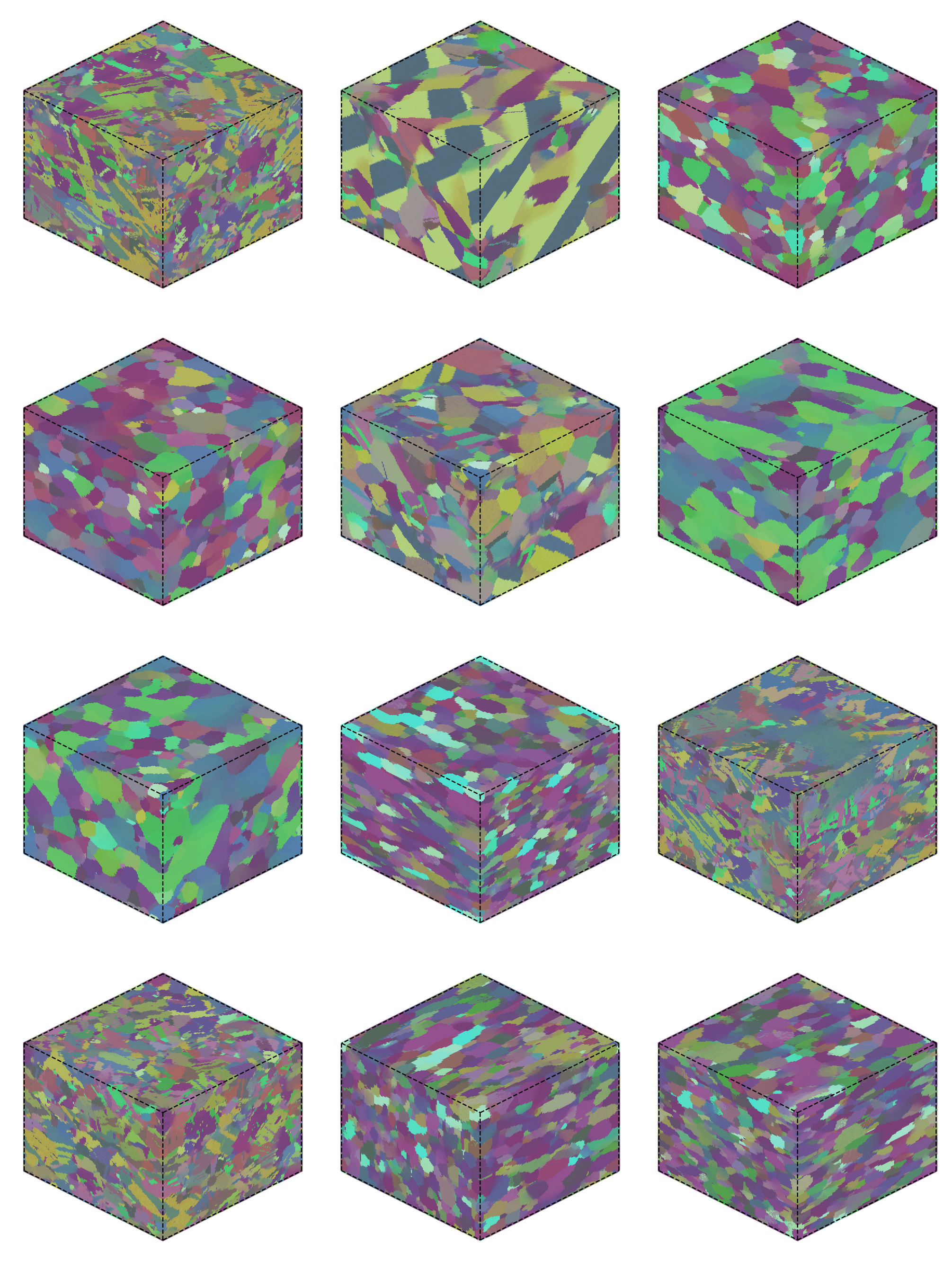}
\caption{Example synthetic microstructures (Size of $128^3$).}
\label{fig:DGEN:samples0}
\end{figure}

\begin{figure}[p]
\centering
\includegraphics[width=\linewidth]{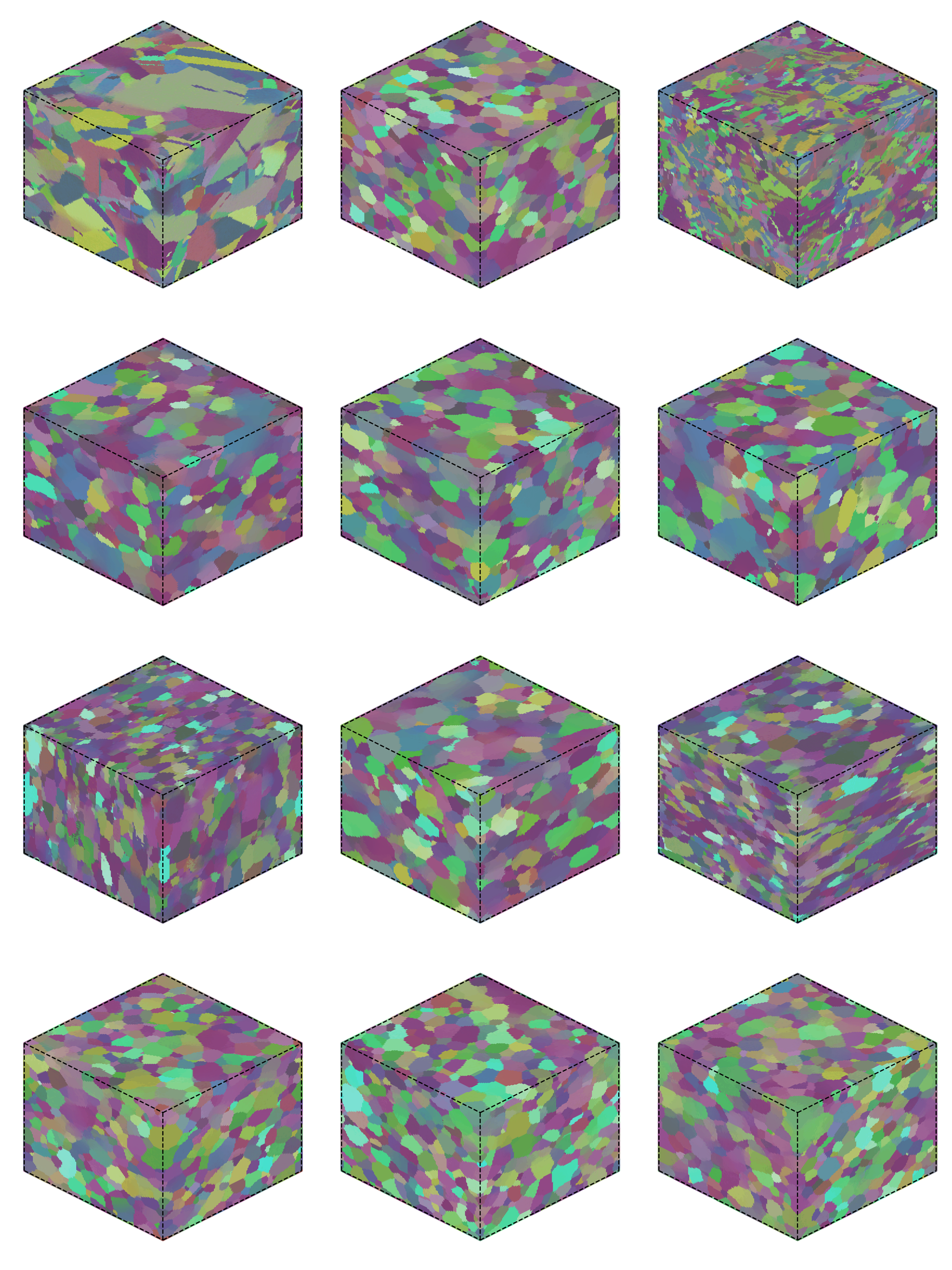}
\caption{Example synthetic microstructures (Size of $128^3$).}
\label{fig:DGEN:samples1}
\end{figure}

\begin{figure}[p]
\centering
\includegraphics[width=\linewidth]{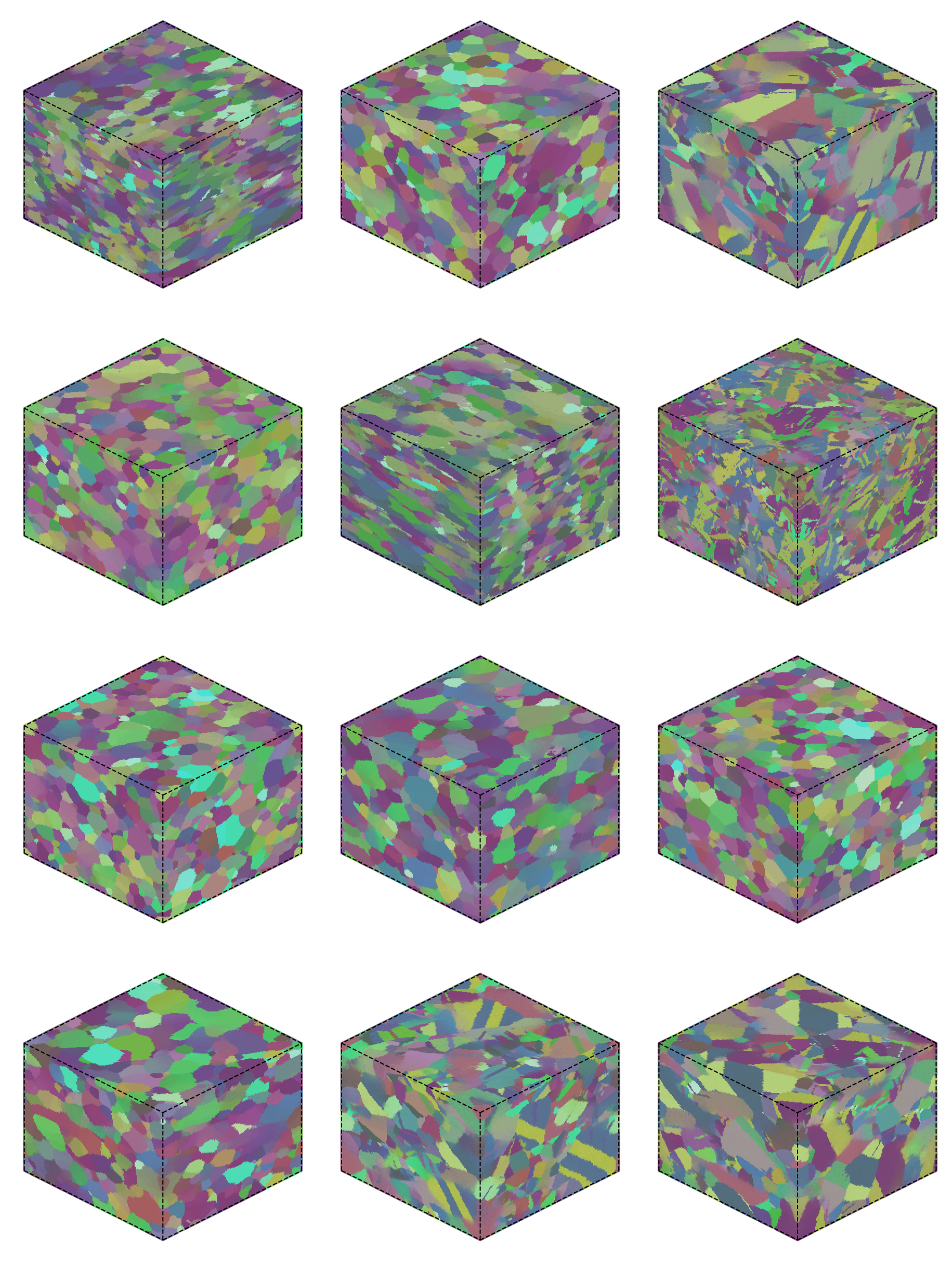}
\caption{Example synthetic microstructures (Size of $128^3$).}
\label{fig:DGEN:samples2}
\end{figure}

\FloatBarrier

\newpage

\end{document}